%% file: paper.tex
\documentclass[10pt,twocolumn]{article}

\usepackage{cite}
\usepackage[pdftex]{graphicx}
\usepackage{amsmath}
\usepackage{color}
\usepackage{algorithm,algorithmic}
\usepackage{amsmath,bm}
\usepackage{amsfonts}
\usepackage{ascmac}
\usepackage{amssymb}
\usepackage{amsthm}
\usepackage{url}

\usepackage{mcr}

\newcommand*\samethanks[1][\value{footnote}]{\footnotemark[#1]} 

\begin{document}
\title{Safe RuleFit: Learning Optimal Sparse Rule Model by Meta Safe Screening}

\author{Hiroki~Kato~\thanks{Department of Computer Science, Nagoya Institute of Technology, Nagoya, Aichi, 466-8555, Japan}~\thanks{Equally contributed.}~,
        Hiroyuki~Hanada~\thanks{Center for Advanced Intelligence Project, RIKEN, Chuo, Tokyo, 103-0027, Japan}~\samethanks[2]~,
        Ichiro~Takeuchi~\samethanks[1]~\samethanks[3]~\thanks{Center for Materials Research by Information Integration, National Institute for Materials Science, Tsukuba, Ibaraki, 305-0047, Japan}~\thanks{Corresponding to:\\{\tt takeuchi.ichiro.n6@f.mail.nagoya-u.ac.jp}}
}

\date{Last update of contents: August 17, 2021\\Last update of authors' information: March 12, 2025}

\maketitle

\begin{abstract}
\input{abst}
\end{abstract}

\input{Sec1}
\input{Sec2}
\input{Sec3}
\input{Sec4}
\input{Sec5}
\input{Sec6}

\section*{Acknowledgment}
This work was partially supported by MEXT KAKENHI (17H00758, 16H06538), JST CREST (JPMJCR1502), JST/AIP Accelerated PRISM research (JPMJCR18ZF), RIKEN Center for Advanced Intelligence Project, and JST support program for starting up innovation-hub on materials research by information integration initiative.

\bibliographystyle{unsrt}
\bibliography{bib}

\clearpage
\appendix

\input{AppA}

\input{AppB}

\input{AppC}
\input{AppD}
\end{document}

%% file: abst.tex
We consider the problem of learning a {\em sparse rule model}, a prediction model in the form of a sparse linear combination of rules, where a rule is an indicator function defined over a hyper-rectangle in the input space.
Since the number of all possible such rules is extremely large, it has been computationally intractable to select the optimal set of active rules.
In this paper, to solve this difficulty for learning the optimal sparse rule model, we propose {\em Safe RuleFit (SRF)}.
Our basic idea is to develop \emph{meta safe screening (mSS)}, which is a non-trivial extension of well-known safe screening (SS) techniques.
While SS is used for screening out one feature, mSS can be used for screening out multiple features by exploiting the inclusion-relations of hyper-rectangles in the input space.
SRF provides a general framework for fitting sparse rule models for regression and classification, and it can be extended to handle more general sparse regularizations such as group regularization.
We demonstrate the advantages of SRF through intensive numerical experiments.

%% file: Sec1.tex
\section{Introduction}\label{sec:introduction}
\begin{figure}[t]
\vskip 0.2in
\begin{center}
\centerline{\includegraphics[width=0.99\columnwidth]{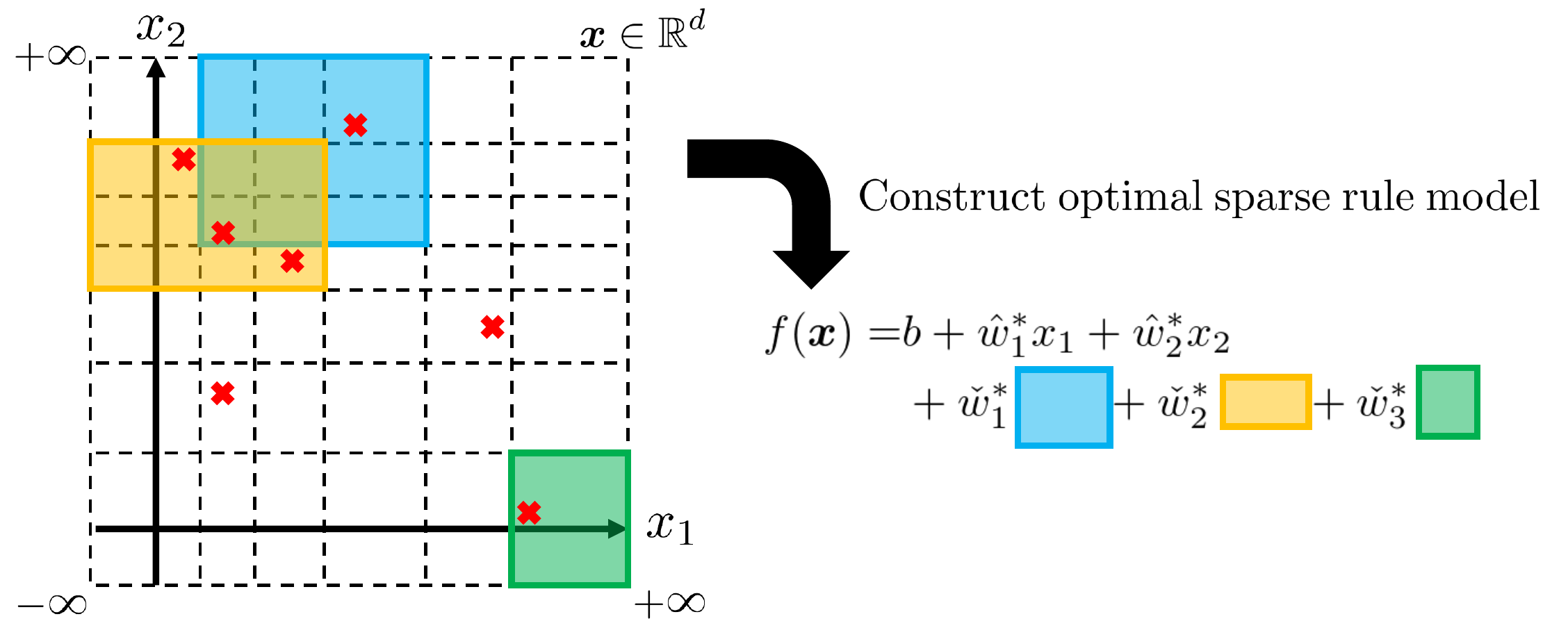}}
\caption{Illustrative example of sparse rule models.
Here, a rule corresponds to a (hyper)rectangle in the input space.
Among a large number of possible rules (hyper-rectangles),
only a small subset of them (three rules in this example) are used in the prediction model.}
\label{fig:data_matrix}
\end{center}
\vskip -0.2in
\end{figure}

We consider a \emph{sparse rule model}, a prediction model in the form of a sparse linear combination of \emph{rules}, where each rule is represented as an indicator function defined over a hyper-rectangle in the input space (e.g., \cite{friedman2008predictive}),
that is, the rule returns 1 if an instance is in the hyper-rectangle or 0 otherwise.
We regard all possible rules as features in prediction model, and study how to fit a sparse model (see \S\ref{sec:segment} and \S\ref{sec:sparse_learning} for formal definitions). Rules selected by sparse modeling can be interpreted as important for prediction.
Fig.~\ref{fig:data_matrix} shows an example of sparse rule models. 
Sparse rule models are useful since they can approximate complex nonlinear functions whilst maintaining easy interpretation.
While complex black box models such as deep neural network models can be used in some fields, interpretable models such as rule models are desirable in fields that require an explanation of decision-making mechanisms such as medical diagnosis \cite{biran2017explanation,carvalho2019machine}.
The goal of this paper is to develop a learning algorithm for finding \emph{optimal} sparse rule models. 

%
In existing studies, a sparse rule model is fitted in two steps.
First, a moderate-size dictionary of rules (a set of rules) is constructed, e.g., by recursive partitioning of the input space. 
Then, a sparse rule model is fitted
over the dictionary by using a standard sparse learning method such as LASSO~\cite{tibshirani1996regression}.
This two-step approach is suboptimal in the sense that the dictionary is not constructed by considering all possible rules, i.e., all possible hyper-rectangles in the input space. 
The difficulty lies in the fact that the number of all possible rules is extremely large.
If the dataset contains 10 original features,
and each feature has 10 distinct values,
the number of all possible rules is as large as $2.5\times 10^{17}$ (see \eqref{eq:num-all-rules}).
Therefore, existing methods restrict attention, e.g., to the rules obtained by recursive partitioning of the input space.

\begin{figure}[tp]
\begin{center}
\includegraphics[width=0.75\hsize]{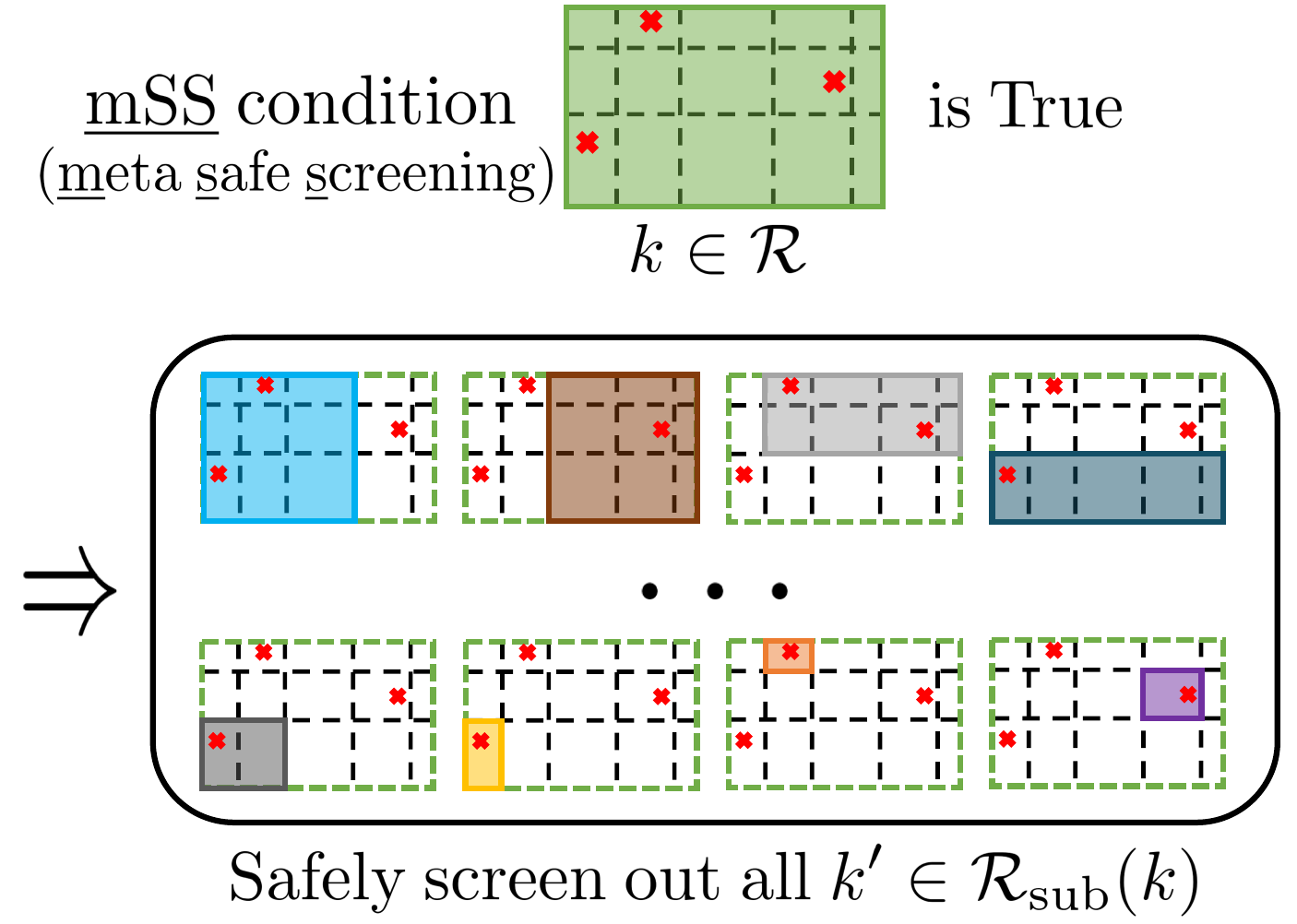}
\end{center}
\vspace{-1.5em}
\caption{Schematic formulation of meta safe screening (mSS) in the proposed SRF method.}
\label{fig:mssc_ssc}
\end{figure}

In this paper we propose a method called {\em Safe RuleFit} (SRF)
for learning optimal sparse rule models.
%
SRF is formulated as a convex optimization problem with an extremely large number of binary features corresponding to all possible rules.
Although a variety of sparse learning algorithms for high-dimensional data have been proposed and explored (e.g., \cite{bach2012optimization}), none of them can handle an extremely large number of features, as per what is considered herein.
To tackle this difficulty, we employ the idea of safe screening (SS) \cite{ghaoui2012safe}.
SS is a method for reducing the number of features to be optimized before and/or during optimization without jeopardizing the optimality of the fitted model.
If a feature satisfies a condition called \emph{SS condition}\footnote{
In the literature, \emph{safe screening conditions} are usually referred to as \emph{safe screening rules}. Since we use ``rule'' with a different meaning in this paper, we denote safe screening rules as safe screening conditions.
}, the feature is guaranteed to be inactive at the optimal sparse model, and hence the feature can be screened out before or/and during optimization.
Unfortunately, however, we cannot apply existing SS methods to our problem because it is impossible to check the SS condition for each of the extremely large number of features.
Our basic idea is to exploit the inclusion-relations of hyper-rectangles in the input space,
and derive \emph{meta safe screening (mSS) condition}
which allows us to screen out multiple rules together (Fig. \ref{fig:mssc_ssc}).
%
If the mSS condition for a certain hyper-rectangle is satisfied,
then
SS conditions of
all the hyper-rectangles included in the above hyper-rectangle are satisfied, 
meaning that all the corresponding rules can be safely screened out together.
By considering a tree structure representing the inclusion-relations of all possible rules (see \S~\ref{sec:tree}), the mSS condition can be effectively used for developing an efficient pruning strategy of the tree. 
SRF can be used for fitting a wide class of regularized convex learning problems including $L_1$-penalized least-square regression and $L_1$-penalized logistic regression as well as their extensions to sparse group penalty versions.
In addition, we also show that SRF can be applicable to an extended penalty function called the {\em sparse group LASSO (SGL) penalty} \cite{simon2013sparse}, that reduces the variety of feature combinations and thus helps the interpretability of the set of identified rules (\S\ref{sec:extension}).

\subsection{Related works} \label{sec:related}

Prediction models with rules have been intensively studied in the literature
\cite{dembczynski2008maximum,dembczynski2008solving,friedman2008predictive,aho2012multi,letham2015interpretable}.
In these existing methods, the rule {\em dictionary} is constructed based on recursive partitioning in the input space (e.g., Decision Tree or Random Forest).
%
%
%
In addition, a more direct approach to identify hyper-rectangles has been proposed for the two-class classification problem (see, e.g., \cite{goh2014box} and the references therein).
As already noted, these existing methods are suboptimal in the sense that all possible rules are not considered.
To the best of our knowledge, among the many rule model studies, the only approaches that consider all possible rules are \cite{eckstein2017rule} and \cite{wei2019generalized}. In these two methods, a feature in the form of a rule is added one by one sequentially by searching over the tree structure similar to the one we use in our proposed SRF method.
The main drawback of these methods is that it must repeat the tree search many times, in contrast to SRF because it only involves a single tree search.
Furthermore, \cite{eckstein2017rule} is applicable only to regression, in contrast to SRF which can be used for a wide class of convex learning problems including not only regression but also classification.
Since these two methods \cite{eckstein2017rule,wei2019generalized} behave very similarly, we only compare our proposed SRF method with {\em REPR} proposed in \cite{eckstein2017rule} in \S\ref{sec:experiment}.

Safe screening (SS) methods have been proposed as efficient sparse learning algorithms
\cite{ghaoui2012safe,wang2013lasso,liu2014safe,xiang2014screening,wang2014safe,bonnefoy2014dynamic,fercoq2015mind,ndiaye2015gap,ndiaye2016gap}.
%
%
%
%
In SS, if a feature satisfies the SS condition, then the feature is guaranteed to be inactive at the optimal sparse model, and hence it can be safely screened out.
%
%
%
As already noted, we introduce the mSS condition by which not only a single feature but also multiple features (rules) can be safely screened out together. 
%
%
To the best of our knowledge, there are no other existing SS methods that can screen out multiple features except \cite{nakagawa2016safe},
in which the authors studied high-order interaction models and derived an SS condition such that, once a feature is screened out, then its higher-order features can also be safely screened out. 
%
Although this work is similar in spirit to SRF, and it can enumerate combinations of features, it is still difficult to handle our setup that considers both the combinations of features and of intervals simultaneously (i.e., hyper-rectangle).

A penalty function called the {\em tree-structured group LASSO (TGL) penalty} \cite{kim2010treeguided}, and the SS condition for the TGL penalty has been studied \cite{jie2014twolayer,jie2015multilayer}. Although both the TGL penalty and the proposed SRF use a tree representation defined over the features, the roles of the tree representations are totally different. The tree representation for the TGL penalty is used to incorporate prior knowledge about the hierarchical structures of the features. On the other hand, the tree representation for the SRF is naturally defined and used for efficient computation based on tree pruning. It is also important to note that the tree is actually constructed in TGL penalty, whereas the tree is not actually constructed in the SRF because the tree is effectively pruned by the mSS condition. This is why the SRF can handle a huge number of all possible rules.

\subsection{Notation}
%
For any real number $u$, $(u)_+ := \max\{0, u\}$.
For any natural number $n$, $[n]:=\{1,\ldots,n\}$. 
For $n \times d$ matrix $\bm{A}$, $\bm{A}_{:j}~j\in[d]$ and $A_{ij}$ represent the $j$-th column vector and $(i,j)$ element of $\bm{A}$, respectively. 
Let the norms of matrix $\|\bm{A}\|_2$ and $\|\bm{A}\|_F$ be the spectral norm (the maximum singular value of $\bm{A}$) and the Frobenius norm ($\|\bm{A}\|_F:=\sqrt{\sum_{i\in[n]}\sum_{j\in[d]}|A_{ij}|^2}$), respectively.
For any $d$ dimensional vector $\bm{v}$, we denote $L_1$ norm, $L_2$ norm and $L_{\infty}$ norm by $\|\bm{v}\|_1:=\sum_{j\in[d]}|v_j|$, $\|\bm{v}\|_2:=\sqrt{\sum_{j\in[d]}v_j^2}$ and $\|\bm{v}\|_{\infty}:=\max_{j\in[d]} |v_j|$, respectively. 
The indicator function is written as $I(\cdot)$, i.e., $I(u)=1$ if $u$ is true, or $I(u)=0$ otherwise. 
For any function $f$, $\dom(f)$ represents the domain of $f$.

%% file: Sec2.tex
\section{Problem Formulation}
\label{sec:formulation}
Consider a training set with $n$ instances and $d$ features denoted by $\{(\bm{x}_i,y_i)\}_{i\in[n]}$, where $\bm{x}_i\in\bbR^d$ is a $d$-dimensional input vector and $y_i$ is the label, where $y_i\in\bbR$ for regression and $y_i\in\{-1,+1\}$ for binary classification.
We define the matrix consisting of all input vectors as $\bm{X}:=[\bx_1,\ldots,\bx_n]^{\top}$.
%
%
In this paper, we interpret rules as new features in the prediction model.
For clarity, we denote the original features as the \emph{input features}.

\subsection{Representation of Rules} \label{sec:segment}
%
In this section we formally define the ``rules'' stated in \S\ref{sec:introduction}.
By specifying a pair of $d$-dimensional vectors $\bm{\ell}, \bm{u}\in(\bbR\cup\{-\infty, +\infty\})^d$, a $d$-dimensional hyper-rectangle, denoted by $[\bm{\ell}, \bm{u}]$, is defined.
For two hyper-rectangles $[\bm{\ell}, \bm{u}]$ and $[\bm{\ell}^\prime, \bm{u}^\prime]$, we state that the former includes the latter if $\ell_j \leq \ell^\prime_j$ and $u^\prime_j \leq u_j$ hold for all $j\in[d]$, and denote the inclusion relation as $[\bm{\ell}, \bm{u}] \sqsupseteq [\bm{\ell}^\prime, \bm{u}^\prime]$.
Given a hyper-rectangle $[\bm{\ell}, \bm{u}]$, the corresponding {\em rule} $r_{[\bm{\ell}, \bm{u}]}$ is defined as the indicator function returning 1 if an instance $\bm{x}\in\bbR^d$ is in the hyper-rectangle $[\bm \ell, \bm u]$, or 0 otherwise, i.e., 
\begin{align*}
\textstyle r_{[\bm{\ell},\bm{u}]}(\bm{x}):=\prod_{j\in[d]}I(\ell_{j}\leq x_{j}\leq u_{j}).
\end{align*}

\begin{figure}[tp]
\begin{center}
\includegraphics[width=0.55\hsize]{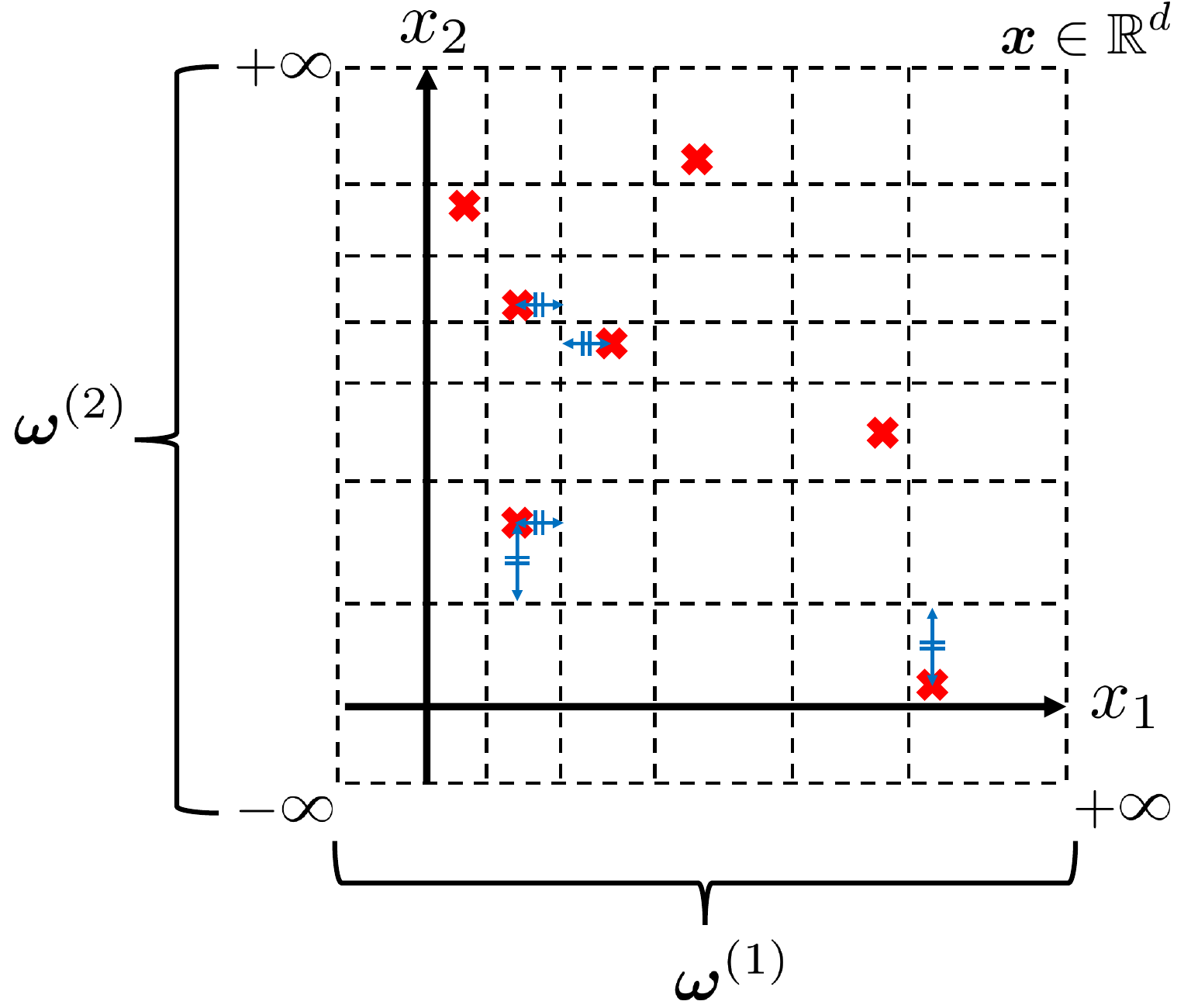}
\end{center}
\vspace{-1em}
\caption{
 Illustrative example of all possible rules to be considered in the case of a two-dimensional dataset represented by red crosses. 
 Here, the vertical lines indicate members of $\bm \omega^{(1)}$, while the horizontal lines indicate members of $\bm \omega^{(2)}$.
 Any rectangles defined by selecting any two vertical lines (any two elements of $\bm \omega^{(1)}$) and any two horizontal lines (any two elements of $\bm \omega^{(2)}$) correspond to the rules to be considered. 
 }
 \label{fig:segment}
\end{figure}

Given a training input matrix $\bm X \in \RR^{n \times d}$, a set of all possible rules to be considered is defined by a set of corresponding hyper-rectangles
$[\bm \ell, \bm u]$s
such that
the 
$j^{\rm th}$
elements of
$\bm \ell$
and
$\bm u$
satisfy
$\ell_j < u_j \text{ and } \ell_j, u_j \in \bm \omega^{(j)}$,
where
$\bm \omega^{(j)}$
is the set of the midpoints of all possible neighboring values in $\bm X_{:j}$ and $\{\pm \infty\}$ for $j \in [d]$.
Fig. \ref{fig:segment} illustrates the set of all possible hyper-rectangles $[\bm \ell, \bm u]$s to be considered. 
%
Here, we can refrain from using all $\bm{\omega}^{(j)}$ and instead focus on a subset so as not to generate an excessively large number of rules\footnote{Note that $\bm{\omega}^{(j)}$ must include $-\infty$ and $+\infty$ even after the subset is taken.}.
We denote the procedure for selecting a subset of $\bm{\omega}^{(j)}$ as {\em discretization} in the sense that, if we do not use the original data values but the discretized alternatives, then we would have the set of such rules. In fact, discretizing data values before obtaining $\bm{\omega}^{(j)}$ will provide a similar result, which is known as the {\em binning} of data values.
In the experiment in \S\ref{sec:experiment} we adopted two discretization methods (see Appendix \ref{sec:discretization}).
Given $\{\bm{\omega}^{(j)}\}_{j\in[d]}$, the number of all possible rules are calculated as follows:
\begin{align}
\prod_{j\in[d]}\left(\begin{array}{c} |\bm{\omega}^{(j)}|-1 \\ 2 \end{array}\right)
= \prod_{j\in[d]}\frac{|\bm{\omega}^{(j)}| (|\bm{\omega}^{(j)}|-1)}{2}.
\label{eq:num-all-rules}
\end{align}

Hereafter the rules and corresponding hyper-rectangles are indexed by $k = 1, 2, ...$ in an arbitrary order.
We denote the set of all possible rules as $\cR$, and write the hyper-rectangle corresponding to the rule $k \in \cR$ as $[\bm \ell^{(k)}, \bm u^{(k)}]$. Furthermore, with a slight abuse of notation, a rule $r_{[\bm{\ell}^{(k)},\bm{u}^{(k)}]}(\bm{x})$ is simply written as $r_k(\bm{x})$.

\subsection{Sparse Learning with Rules} \label{sec:sparse_learning}
We fit the following linear regression or classification model
consisting of linear terms and rule terms:
\begin{align}
f(\bx):=b+\bx^{\top}\hbw+\sum_{k\in\cR}r_k(\bm{x})\chw_k,
\label{eq:linear_model}
\end{align}
where the model parameters to be learned are $\hbw\in\bbR^d$, $\chbw\in\bbR^{|\cR|}$ and $b\in\bbR$. 
We consider a model in the form of \eqref{eq:linear_model} by following the conventional studies on prediction models with rules \cite{friedman2008predictive,eckstein2017rule}, in which it is stated that using not only the rule terms $\sum_{k\in\cR}r_k(\bm{x})\chw_k$ but also the linear terms $\bx^{\top}\hbw$ is beneficial for rendering the entire model simpler and more interpretable.
\begin{remark}
We add that the proposed method SRF works even for the model without linear terms,
namely, $f(\bx):=b+\sum_{k\in\cR}r_k(\bm{x})\chw_k$ rather than \eqref{eq:linear_model}.
The changes of SRF by removing linear terms is presented in Appendix \ref{app:no-linear-term},
and the experimental results between these two models are presented in \S\ref{sec:experiment}, Exp.5.
\end{remark}

We learn these model parameters by the following $L_1$-penalized empirical risk minimization:
\begin{align}
& \min_{\hbw\in\bbR^d,\chbw\in\bbR^{|\cR|}, b\in\bbR} P_{\lambda}(\hbw,\chbw,b),
	\quad\text{where} \label{eq:primal}\\
& P_{\lambda}(\hbw,\chbw,b) := \sum_{i\in[n]}\ell_i(y_i,f(\bx_i))+\lambda(\|\hbw\|_1+\|\chbw\|_1),
	\nonumber
\end{align}
and $\lambda>0$ is the regularization hyperparameter.
We call \eqref{eq:primal} the {\em primal} problem in contrast to the dual problem defined later in \eqref{eq:dual}.
Let $\hbw^*$, $\chbw^*$ and $b^*$ be the optimal solution of \eqref{eq:primal}.
We assume that the loss function $\ell_i:\bbR\times\bbR\rightarrow\bbR$ is convex and smooth.
The $L_1$ penalty is known to make the solution vector sparse, i.e., most of the elements would be zero. 
%
%
For regression, the formulation \eqref{eq:primal} is known as LASSO when $\ell_i(y_i,u_i):=\frac{1}{2}(y_i-u_i)^2~\forall i\in[n]$ (squared loss; $y_i\in\bbR$). For binary classification, $L_1$-penalized logistic regression is commonly used, where $\ell_i(y_i,u_i):=\log(1+\exp(-y_iu_i))~\forall i\in[n]$ (logistic loss; $y_i\in\{-1, +1\}$).
Many other useful loss functions (e.g. squared-hinge SVM) can be used here. 
For notational simplicity, we define $\bm{\xi}\in\bbR^n$ as $\bm{\xi}:=\bm{1}$ for the squared loss and $\bm{\xi}:=\bm{y}$ for the logistic loss, and define $\hbZ\in\bbR^{n\times d}$ and $\chbZ\in\bbR^{n\times |\cR|}$
as
$\hZ_{ij}:=\xi_ix_{ij}$
and
$\chZ_{ik}:=\xi_ir_{k}(\boldx_i)$,
respectively,
for
$i \in [n]$,
$j \in [d]$,
and
$k \in \cR$. 

Then, let us consider the {\em dual} problem of \eqref{eq:primal}
(see, e.g., Theorem 31.3 of \cite{rockafellar1970convex}):
\begin{align}
&\max_{\btheta\in\bm{\Delta}} D_{\lambda}(\btheta),
	\quad\text{where}\quad
	D_{\lambda}(\btheta) = -\sum_{i\in[n]}\ell_i^*(-\lambda\theta_i), \label{eq:dual} \\
&\bm{\Delta}:=\{\btheta\in\bbR^n:\|\hbZ^{\top}\btheta\|_{\infty}\leq 1,\|\chbZ^{\top}\btheta\|_{\infty}\leq 1, \nonumber \\
&\phantom{\bm{\Delta}:=\{} \bxi^{\top}\btheta=0,\btheta\in\dom(\ell^*)\}, \nonumber
\end{align}
and $\ell_i^*: \bbR\rightarrow\bbR$ is the {\em convex conjugate} of $\ell_i$ with respect to the second argument.
For the squared loss and the logistic loss, $\ell_i^*(-\lambda\theta_i):=\frac{\lambda^2}{2}\theta_i^2-\lambda y_i\theta_i$ and $\ell_i^*(-\lambda\theta_i):=(1-\lambda\theta_i)\log(1-\lambda\theta_i)+\lambda\theta_i\log(\lambda\theta_i)$ with $0<\theta_i<\frac{1}{\lambda}$ for all $i\in[n]$, respectively.
Furthermore, $\bm{\Delta}\subset\bbR^n$ represents the domain of $D_{\lambda}$. Let $\btheta^*$ be the optimal solution of \eqref{eq:dual}.
Using the definitions above, for any $j\in[d]$ and $k\in\cR$, we can prove that the following relationships hold between the primal and dual optimal solutions $\hbw^*,\chbw^*$ and $\btheta^*$:
\begin{lemma} \label{lem:kkt}
For any $j\in[d]$ and any $k\in\cR$,
\begin{align}
& \theta^*_i = -\frac{1}{\lambda}\cdot \left[\frac{\partial}{\partial u}\ell_i(y, u)\right]_{y\gets y_i^*, u\gets u_i^*},
	\label{eq:kkt-primal2dual} \\
& |\hbZ_{:j}^{\top}\btheta^*|<1\Rightarrow \hw_j^*=0,\quad |\chbZ_{:k}^{\top}\btheta^*|<1\Rightarrow \chw_k^*=0.
	\label{eq:kkt-dual2primal}
\end{align}
\end{lemma}
The proof of Lemma \ref{lem:kkt} is presented in Appendix \ref{sec:proof_kkt}.

\subsection{Safe Screening (SS)} \label{sec:safe_screening}
In this section, we describe the safe screening (SS) framework outlined in \S \ref{sec:introduction}. 
As stated in \S \ref{sec:sparse_learning}, we expect many elements of $\hbw^*$ and $\chbw^*$ to be zero by the $L_1$ penalty. 
{\em Safe screening} (SS) 
can detect parts of $j\in[d]$ and/or $k\in\cR$ such that $\hw_j^*=0$ and/or $\chw_k^*=0$, which we call \emph{inactive} features, before the optimal $\hbw^*$ and $\chbw^*$ are available, as stated in the following lemma:
\begin{lemma}[Theorem 3 in \cite{ndiaye2015gap}] \label{lem:sphere_bound}
Let the loss function of the primal problem \eqref{eq:primal} be a convex and $1/\gamma$-smooth function. Furthermore, let $(\hbw',\chbw',b')\in\dom(P_{\lambda})$ and $\btheta'\in\dom(D_{\lambda})$ be arbitrary feasible solutions of the primal and dual problems, respectively. Then, the optimal solution of the dual problem $\btheta^*$ is always within the following hypersphere whose center is $\btheta'$ and whose radius is $r_{\lambda}$  in the dual solution space $\btheta\in\bbR^n$:
\begin{align}
&\bm{\Theta}_{\btheta^*}:=\{\btheta \in \bbR^n : \|\btheta'-\btheta\|_2\leq r_{\lambda}\}, \notag \\
\text{where}~&r_{\lambda}:=\sqrt{2\gamma^{-1}(P_{\lambda}(\hbw',\chbw',b')-D_{\lambda}(\btheta'))}/\lambda. \label{eq:safe_radius}
\end{align}
\end{lemma}
The value $P_{\lambda}(\hbw',\chbw',b')-D_{\lambda}(\btheta')$ is called the {\em duality gap}.
We can apply Lemma \ref{lem:sphere_bound} to Lemma \ref{lem:kkt} to detect inactive features without knowing the dual optimal solution $\btheta^*$. 
To this end, we compute upper bounds of $|\hbZ_{:j}^{\top}\btheta^*|~\forall j\in[d]$ and $|\chbZ_{:k}^{\top}\btheta^*|~\forall k\in\cR$ in Lemma \ref{lem:kkt}, which we denote as $\hSSUB(j)$ and $\chSSUB(k)$, such that
\begin{subequations}
\label{eq:ub_jk}
\begin{align}
&\hSSUB(j) \geq |\hbZ_{:j}^{\top}\btheta|\quad\forall\btheta\in\bbR^n:~\btheta\in\bm{\Theta}_{\btheta^*},~\bxi^{\top}\btheta=0, \label{eq:ub(j)} \\
&\chSSUB(k) \geq |\chbZ_{:k}^{\top}\btheta|\quad\forall\btheta\in\bbR^n:~\btheta\in\bm{\Theta}_{\btheta^*},~\bxi^{\top}\btheta=0. \label{eq:ub(k)}
\end{align}
\end{subequations}
We can compute such upper bounds as follows:
\begin{lemma} \label{lem:val_ub}
The following $\hSSUB(j)$ and $\chSSUB(k)$ satisfy the condition \eqref{eq:ub_jk}:
\begin{align}
&\hSSUB(j):=|\hbZ_{:j}^{\top}\btheta'|+r_{\lambda}\|\hbZ_{:j}-\Pi_{<\bxi>}(\hbZ_{:j})\|_2, \label{eq:val_ub(j)} \\
&\chSSUB(k):=|\chbZ_{:k}^{\top}\btheta'|+r_{\lambda}\|\chbZ_{:k}-\Pi_{<\bxi>}(\chbZ_{:k})\|_2, \label{eq:val_ub(k)}
\end{align}
where $\Pi_{<\bm{u}>}(\bm{v}):=\frac{\bm{v}^{\top}\bm{u}}{\|\bm{u}\|_2^2}\bm{u}$ is the projected vector of $\bm{v}\in\bbR^n$ onto $\bm{u}\in\bbR^n$.
\end{lemma}
%
The proof of the lemma is presented in Appendix \ref{sec:proof_ub},
and the calculations of Lemma \ref{lem:val_ub} for two loss functions in \S\ref{sec:sparse_learning} are given in Appendix \ref{app:SS-loss-functions}.

From Lemmas \ref{lem:kkt} and \ref{lem:val_ub}, we have
\begin{align}
\hSSUB(j)<1\Rightarrow \hw_j^*=0,~\chSSUB(k)<1\Rightarrow \chw_k^*=0.
\label{eq:screening}
\end{align}
We denote the precedent parts
$\hSSUB(j)<1$
and
$\chSSUB(k)<1$
as \emph{safe screening (SS) conditions}.
The conditions \eqref{eq:screening} can be used to detect inactive features without knowing the optimal solutions $\hbw^*,\chbw^*,b^*$ nor $\btheta^*$.
This renders the optimization more efficient while guaranteeing the optimality of the solution.
Note, however, that it is computationally infeasible to compute $\chSSUB(k)$ for each rule $k\in\cR$ since the number of all possible rules $|\cR|$ grows exponentially with $d$.
%
To tackle this problem, we develop {\em meta safe screening (mSS)} which allows us to identify multiple inactive features simultaneously, and screen them out together.

%% file: Sec3.tex
\section{Proposed Method: Safe RuleFit (SRF)}
\label{sec:propsed}
In this section, we propose \emph{Safe RuleFit (SRF)} that enables us to learn the optimal sparse rule model even if the number of all possible rules to be considered is extremely large. 
Our main idea is to develop \emph{meta safe screening (mSS)} by which multiple rules can be safely screened out at once. 
The mSS method is developed based on the inclusion-relations of hyper-rectangles in the input space. 
For a rule
$k\in\cR$,
let us define a set of rules
$\cR_{\rm sub}(k) := \{ k^\prime\in\cR : [\bm \ell^{(k^\prime)}, \bm u^{(k^\prime)}] \sqsubseteq [\bm \ell^{(k)}, \bm u^{(k)}]\}$
whose hyper-rectangles are included in the hyper-rectangle of the rule $k$.
In the mSS method, we introduce a novel \emph{mSS condition} such that, if the mSS condition for a rule $k$ is met, then SS condition of rule $k^\prime$ for all $k^\prime \in \cR_{\rm sub}(k)$ is guaranteed to be met (see Fig.\ref{fig:mssc_ssc}). 
Thus, just by checking the mSS condition for a rule $k$, all the rules in $\cR_{\rm sub}(k)$ can be safely screened out at once. 
Based on the inclusion-relations of hyper-rectangles, we introduce a tree representation of rules, in which each node corresponds to a rule, and ancestor-descendant relations in the tree represent the inclusion-relations of the corresponding hyper-rectangles. 
In SRF, mSS is effectively used for pruning the branches of the tree, by which we can efficiently handle an extremely large number of rules.

\subsection{Meta Safe Screening (mSS)} \label{sec:pruning}
Let us denote the set of active rules in the optimal solution of \eqref{eq:primal} by $\cR^*:=\{k\in\cR : \chw_k^*=0\}$. 
In SRF, mSS is effectively used for finding a superset $\tilde{\cR}\supseteq\cR^*$ by efficiently searching over the tree representation of the rules. 
Once we obtain such a superset $\tilde{\cR}$, and the size of the $\tilde{\cR}$ is moderate, the optimal sparse model can be obtained by solving the problem \eqref{eq:primal} over the set of rules in $\tilde{\cR}$ as stated in the following property.
\begin{property} \label{pr:optimize-for-subset}
Let $\tilde{\cR}$ be a set such that $\cR^*\subseteq\tilde{\cR}\subseteq\cR$, and $\chbw_{\tilde{\cR}}\in\bbR^{|\tilde{\cR}|}$ and $\chbw_{\cR\setminus\tilde{\cR}}\in\bbR^{|\cR\setminus\tilde{\cR}|}$ be $\chbw_{\tilde{\cR}}:=\{\chw_k\}_{k\in\tilde{\cR}}$ and $\chbw_{\cR\setminus\tilde{\cR}}:=\bm{0}$, respectively. Then, the optimal solution of \eqref{eq:primal} is given by
\begin{align}
(\hbw^*,\chbw_{\tilde{\cR}}^*,b^*)&\in\argmin_{\hbw\in\bbR^d,\chbw_{\tilde{\cR}}\in\bbR^{|\tilde{\cR}|},b\in\bbR}P_{\lambda}(\hbw,\chbw_{\tilde{\cR}},b), \label{eq:primal_only_active} \\
\chbw_{\cR\setminus\tilde{\cR}}^*&=\bm{0}. \notag
\end{align}
\end{property}
Next, let us describe how mSS is used for finding such a superset $\tilde{\cR}$.
%
%
%
%
%
The core of the mSS is stated in the following theorem. 
\begin{theorem} \label{thm:pruning}
Let the loss function of the primal problem \eqref{eq:primal} be a convex and $1/\gamma$-smooth function. Given an arbitrary pair of primal and dual feasible solutions $(\hbw^\prime,\chbw^\prime,b^\prime)\in\dom(P_{\lambda}),~\btheta^\prime\in\dom(D_{\lambda})$ and \eqref{eq:safe_radius}, for any $k^\prime\in\cR_{\rm sub}(k)$ the following relationship holds:
\begin{align*}
&\chmSSUB(k):=\eta_k+r_{\lambda}\|\chbZ_{:k}\|_2<1\Rightarrow \chSSUB(k^\prime)<1, \\
&\text{where}~\eta_k:=\max\Bigl\{\sum_{i:\xi_i\theta_i^\prime>0}\chZ_{ik}\theta_i^\prime,-\sum_{i:\xi_i\theta_i^\prime<0}\chZ_{ik}\theta_i^\prime\Bigr\}.
\end{align*}
for all $(k,k^\prime)\in\cR\times\cR$ such that $k^\prime\in\cR_{\rm sub}(k)$.
\end{theorem}
%
The proof of this theorem is shown in Appendix \ref{sec:proof_pruning}. 
We denote the precedent part ${\rm mSS_{UB}}(k) < 1$ as \emph{meta safe screening (mSS) condition}. 
A nice property of mSS is that, without using any information on $k^\prime\in\cR_{\rm sub}(k)$, it can tell whether the SS condition for the rule $k^\prime$ can be met or not. 
This property enables us to screen out all rules in $\cR_{\rm sub}(k)$ together just by checking a single mSS condition for the rule $k$. 
To prove Theorem \ref{thm:pruning}, we use the following property for any pair of two rules such that $k^\prime\in\cR_\mathrm{sub}(k)$:
\begin{align}
\forall \bx\in\bbR^d:\quad & r_{k}(\bx) = 0 \Rightarrow r_{k^\prime}(\bx) = 0, \nonumber\\
& r_{k^\prime}(\bx) = 1 \Rightarrow r_{k}(\bx) = 1.
	\label{eq:property-rule}
\end{align}
Moreover, with \eqref{eq:property-rule} we have the following corollary:
\begin{corollary} \label{coro:deeper}
For any rules $k\in\cR$ and $k^\prime\in\cR_{\rm sub}(k)$, $\chmSSUB(k)\geq\chmSSUB(k^\prime)$.
\end{corollary}
The corollary states that, the smaller the hyper-rectangle is, the more the mSS condition would be met. 
The proof of Corollary \ref{coro:deeper} is presented in Appendix \ref{sec:proof_corollary_deeper}.
\subsection{Tree Representation of Rules} \label{sec:tree}
To effectively use mSS in SRF, we introduce a tree representation of all possible rules in $\cR$.
The main principle is that each node in the tree represents a rule, and if a node $k^\prime$ is a descendant of $k$ then the hyper-rectangle of the rule $k^\prime$ must be included in the hyper-rectangle of the rule $k$.
By constructing such a tree, whenever we add a node, we check if the mSS condition is met. 
Then, if the condition is met, we can screen out all the rules corresponding to its descendant nodes by pruning the branch.

To appropriately enumerate all rules in $\cR$, we use the algorithm presented in \cite{kaytoue2011revisiting}.
The outline of the algorithm is as follows:

\begin{property} \label{pr:rule-enumeration} \cite{kaytoue2011revisiting}
Suppose that each node in a tree represents a rule.
Given $\{\bm{\omega}^{(j)}\}_{j\in[d]}$ (\S\ref{sec:segment}), all possible rules can be enumerated as follows.
Initialize a tree $T$ with only one root node $[-\infty, +\infty]_{j\in[d]}$, then do the followings for any leaf node $[\bm{\ell}^\prime, \bm{u}^\prime]$ in $T$ until no more child nodes can be added:
\begin{itemize}
\item[(a)] Let $\tau = \max_{j\in[d], [\ell^\prime_j, u^\prime_j] \neq [-\infty, +\infty]} j$. \\
	\{To avoid duplications, we shrink intervals for input features of smaller index to larger.\}
\item[(b)] For each $j\in\{\tau, \tau+1, \dots, d\}$, let $\ell^{\prime\mathit{next}}_{j}$ be the next larger element in $\bm{\omega}^{(j)}$ than $\ell^\prime_{j}$.
			If $\ell^{\prime\mathit{next}}_{j} < u^\prime_{j}$, then add a copy of $[\bm{\ell}^\prime, \bm{u}^\prime]$ as a child node except that $\ell^\prime_{j}$ is replaced with $\ell^{\prime\mathit{next}}_{j}$.
\item[(c)] For each $j\in\{\tau, \tau+1, \dots, d\}$, let $u^{\prime\mathit{prev}}_{j}$ be the next smaller element in $\bm{\omega}^{(j)}$ than $u^\prime_{j}$.
	If $\ell^\prime_{j} = -\infty < u^{\prime\mathit{prev}}_{j}$, then add a copy of $[\bm{\ell}^\prime, \bm{u}^\prime]$ as a child node except that $u^\prime_{j}$ is replaced with $u^{\prime\mathit{prev}}_{j}$. \\
	\{To avoid duplications, we request $\ell^\prime_{j} = -\infty$ here in order to, for every input feature, shrink the upper bound of the rule before the lower bound.\}
\end{itemize}
\end{property}

The complete algorithm of enumerating all the rules in $\cR$, including the implementations of issues in \S\ref{sec:pruning-in-enumeration} and \S\ref{sec:closed-rules}, is presented in Algorithm \ref{alg:enumerate_all_closed}.
The time complexity of the algorithm is $O(Nnd^2)$ if $\mathrm{OnlyClosed} = \mathbf{true}$ or $O(Nd(n+d))$ otherwise, where $N$ is the number of examined rules (i.e., number of executions of the function \textsc{MakeSubtree}). In the function \textsc{MakeSubtree}, for the $O(d)$-time loop, we need $O(n+d)$ time to create the ``NewNode''. Moreover, to conduct the function \textsc{CheckClosure} we need $O(nd)$ time.

\subsubsection{Application of pruning conditions in the rule enumeration} \label{sec:pruning-in-enumeration}

For the strategy of Property \ref{pr:rule-enumeration} that just enumerates all possible rules for given $\{\bm{\omega}^{(j)}\}_{j\in[d]}$,
we introduce mSS (\S\ref{sec:pruning}) to reduce the number of enumerated rules.
Theorem \ref{thm:pruning} states that, if $\chmSSUB(k) < 1$ for a rule $k$, then any rules represented as descendant in the tree are inactive (pruned). So, whenever we add a node (rule) $k$ to a tree, we calculate $\chmSSUB(k)$ and stop adding nodes from here if $\chmSSUB(k) < 1$. Note that we can compute $r_\lambda$ used in $\chmSSUB(k)$ are independent of the rule, that is, we can compute it before enumerating rules.

Moreover, like frequent pattern mining algorithms do, we may use two additional conditions to prune the tree.
One is to specify the {\em minimum support} in frequent pattern mining algorithms, denoted by ``min\_sup''. More specifically, we enumerate only rules such that the number of training instance in the rule (called {\em support}) is at least the predetermined threshold.
The other is to specify the maximum number of input features used in the rule, denoted by ``max\_efs''. More specifically, for a rule $k\in\cR$, we define the \emph{effective feature set (EFS)} of it as the set of input features that are relevant to judging whether an instance is in the rule. EFS can be defined as
\begin{align}
\mathrm{EFS}(k):=\{j\in[d]:[\ell^{(k)}_j,u^{(k)}_j]\neq[-\infty,+\infty]\}.
\label{eq:EFS}
\end{align}
Note that, since the support is non-increasing and the EFS is non-decreasing as we go the tree deeper (i.e., the hyper-rectangle of the rule becomes smaller), we can prune the node if the support becomes smaller than ``min\_sup'' or the EFS becomes larger than ``max\_efs''.
\begin{remark}
In training a rule model, there are many cases where we want to add conditions such as ``min\_sup'' and/or ``max\_efs'' in addition to the criterion of minimizing the penalized loss in equation \eqref{eq:primal}. Our proposed SRF can easily incorporate these additional conditions on the rules and obtain the optimal rule model under these conditions.
Furthermore, if the additional conditions can be checked efficiently in the tree structure such as ``min\_sup'' and ``max\_efs'', a more efficient search over the tree can be possible by combining the meta safe screening rule with these conditions.
\end{remark}

\subsubsection{Avoiding redundant rules} \label{sec:closed-rules}

Finally let us discuss the redundancy of rules.
If there are two rules $k_1$ and $k_2$ such that $r_{k_1}(\bx_i)=r_{k_2}(\bx_i)~\forall i\in[n]$, then these two rules are redundant because they are indistinguishable in the prediction model.
When there are multiple such indistinguishable rules, it is reasonable to pick up only one of them. 
This phenomenon has been studied in the pattern mining literature, and the problem of finding a set of non-redundant patterns is called \emph{closed pattern mining} \cite{pasquier1999efficient,aggarwal2014frequent}.

\begin{figure}[t]
\vskip 0.2in
\begin{center}
\centerline{\includegraphics[width=0.6\columnwidth]{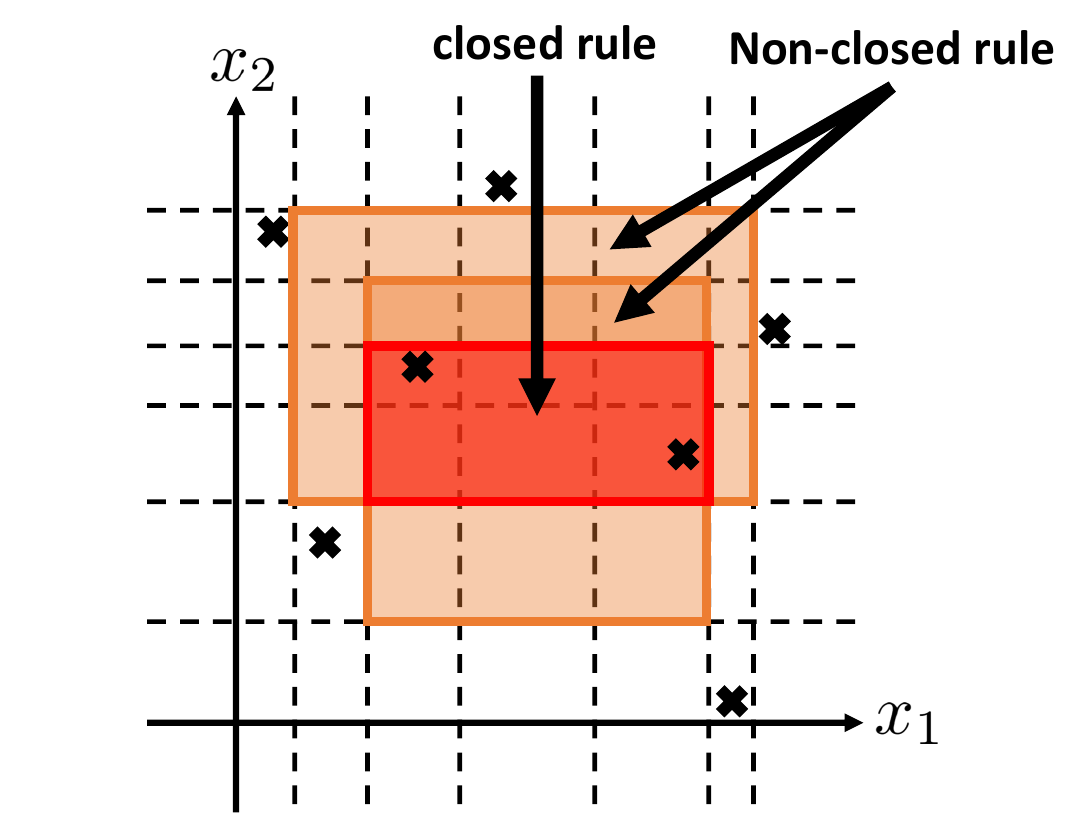}}
\caption{Example of a closed rule (in red) with non-closed rules (in orange).}
\label{fig:closed_rule}
\end{center}
\vskip -0.2in
\end{figure}

We define the {\em closed rule} (or {\em closure}) of $k$, denoted by $k^{cl}$, as the rule with the ``smallest'' hyper-rectangle within the set of rules which are indistinguishable from $k$ (Fig. \ref{fig:closed_rule}). Formally,
\begin{definition}
Given a data matrix $X\in\mathbb{R}^{n\times d}$,
a discretization $\{ \bm{\omega}^{(j)} \}_{j\in[d]}$
and a rule $k$ upon the discretization,
let ${\cal I}_X(k) := \{ i \mid r_k(X_{i:}) = 1 \}$
be the indices of the instances in $X$ that are in the rule.\\
Then, the {\em closed} rule of $k$,
denoted by $k^{cl}$ and its hyper-rectangle being denoted by $[\bm{\ell}^{(k^{cl})}, \bm{u}^{(k^{cl})}]$,
is defined as follows:
\begin{align*}
& \ell^{(k^{cl})}_j := \max\{ t \mid t\in\bm{\omega}^{(j)},~t < X_{ij}~\forall i\in {\cal I}_X(k) \}, \\
& u^{(k^{cl})}_j    := \min\{ t \mid t\in\bm{\omega}^{(j)},~t > X_{ij}~\forall i\in {\cal I}_X(k) \}.
\end{align*}
\end{definition}

The calculation of the closure is presented in the ``\textsc{CheckClosure}'' procedure in Algorithm \ref{alg:enumerate_all_closed}. The procedure also includes the check of the closure: With a similar idea to Property \ref{pr:rule-enumeration}, we can avoid duplicated enumerations of rules.

\begin{algorithm*}[tp]
\caption{Algorithm to enumerate rules with mSS, that is, only rules that can possibly be active in \eqref{eq:primal} for the specified $\bm{X}$, $\bm{y}$ and $\lambda$. Marks (G1) and (G2) denote the processes that we need to modify to apply GSRF (\S\ref{sec:extension}).}
\label{alg:enumerate_all_closed}
\begin{tabular}{c|c}
\begin{minipage}[t]{0.48\hsize}
{\footnotesize
\begin{algorithmic}
\REQUIRE{$\{\bm{\omega}^{(j)}\}_{j\in[d]}$} \COMMENT{See \S\ref{sec:segment}}
\REQUIRE{$\bm{X}\in\mathbb{R}^{n\times d}$, $\bm{y}\in\mathbb{R}^n$, $\lambda > 0$~${}^{\text{(G1)}}$}
\REQUIRE{$\btheta^\prime\in\dom(D_{\lambda})$, $r_\lambda\in\mathbb{R}$}
\STATE \COMMENT{Values needed to compute mSS in Theorem \ref{thm:pruning}.}
\REQUIRE{$\mathrm{min\_sup}\in[n]$}
\STATE \COMMENT{It enumerates only rules such that at least ``$\mathrm{min\_sup}$''}
\STATE \COMMENT{instances in $X$ are in it. To enumerate rules without}
\STATE \COMMENT{this limit, set $\mathrm{min\_sup} \gets 1$.}
\REQUIRE{$\mathrm{max\_efs}\in[d]$}
\STATE \COMMENT{It enumerates only rules whose size of EFS \eqref{eq:EFS}}
\STATE \COMMENT{is ``max\_efs'' or less. To enumerate rules}
\STATE \COMMENT{without this limit, set $\mathrm{max\_efs} \gets d$.}
\REQUIRE{$\mathrm{OnlyClosed}\in\{\mathbf{false}, \mathbf{true}\}$}
\STATE \COMMENT{Whether to enumerate only closed rules or not.}
\ENSURE{$\tilde{\cR}\subseteq\cR$}
\STATE \COMMENT{Candidates of active rules. See Property \ref{pr:optimize-for-subset}.}
\STATE ~
\STATE \hspace{-1em}{\bf function}~\textsc{Enumerate}
\STATE $\mathrm{CandidatesOfActiveRules}\gets\emptyset$
\STATE $\text{RootNode.Rule} \leftarrow (\text{L}[j]=-\infty, \text{U}[j]=+\infty)_{j\in[d]}$
\STATE $\text{RootNode.LastChanged} \leftarrow 1$
\STATE $\text{RootNode.LowerChanged} \leftarrow \mathbf{false}$
\STATE $\text{RootNode.Instances} \leftarrow [n]$
\STATE \COMMENT{The set of training instances in the rule}
\STATE \textsc{MakeSubtree}(RootNode)
\STATE \hspace{-1em}{\bf end function}
\STATE ~
\STATE \hspace{-1em}{\bf function}~$prev_j(t) := \max\{ t^\prime : t^\prime\in\bm{\omega}^{(j)}, t^\prime < t \}$
\STATE \hspace{-1em}{\bf function}~$next_j(t) := \min\{ t^\prime : t^\prime\in\bm{\omega}^{(j)}, t^\prime > t \}$
\STATE \COMMENT{The previous or the next element of $t$ in $\bm{\omega}^{(j)}$.}
\STATE \COMMENT{In the implementation, rather than the values $t$ and $t^\prime$}
\STATE \COMMENT{themselves, let us store the indices $o_t$ and $o_{t^\prime}$}
\STATE \COMMENT{($t$ is the $o_t$-th smallest element in $\bm{\omega}^{(j)}$) instead.}
\STATE \COMMENT{Then these operations can be implemented by just}
\STATE \COMMENT{subtracting or adding 1.}
\STATE ~
\STATE \hspace{-1em}{\bf function}~\textsc{CheckClosure}(Node)
\STATE \COMMENT{Compute the closure of the rule}
\FOR{$j\leftarrow[d]$}
	\STATE $\text{Closure.L}[j] \gets$
	\STATE \hfill$\max\{ t \mid t\in\bm{\omega}^{(j)},~\forall i\in\text{Node.Instances}:~t<X_{ij} \}$
	\STATE $\text{Closure.U}[j] \gets$
	\STATE \hfill$\min\{ t \mid t\in\bm{\omega}^{(j)},~\forall i\in\text{Node.Instances}:~t>X_{ij} \}$
\ENDFOR
\STATE ~
\IF{($\forall j\in[\text{Node.LastChanged}-1]$:\\
	\qquad$\text{Node.Rule.L}[j]=\text{Closure.L}[j]$,\\
	\qquad$\text{Node.Rule.U}[j]=\text{Closure.U}[j]$)}
	\STATE {\bf return} Closure
\ELSE
	\STATE {\bf return} {\bf null}
	\COMMENT{Enumerating this causes a duplication}
\ENDIF
\STATE \hspace{-1em}{\bf end function}
\end{algorithmic}
}
\end{minipage}
&
\begin{minipage}[t]{0.48\hsize}
{\footnotesize
\begin{algorithmic}
\STATE \hspace{-1em}{\bf function}~\textsc{MakeSubtree}(Node)
\STATE {\bf if}~$|\text{EFS}(\text{Node.Rule})| > \mathrm{max\_efs}$~{\bf then}~{\bf return}
\STATE {\bf if}~$|\text{Node.Instances}| < \mathrm{min\_sup}$~{\bf then}~{\bf return}
\STATE ~
\STATE \COMMENT{Check conditions of mSS (Theorem \ref{thm:pruning}) and SS \eqref{eq:screening}.}
\STATE Compute $\eta_k$ and then $\chmSSUB(k)$ in  for
\STATE \hfill$k\gets\text{Node.Rule}$, using $\text{Node.Instances}$ and $r_\lambda$~${}^{\text{(G2)}}$
\STATE {\bf if}~$\chmSSUB(k) < 1$~{\bf then}~{\bf return}
\STATE {\bf if}~$\chSSUB(k)\geq 1$~{\bf then}~$\tilde{\cR}\gets\tilde{\cR}\cup\{k\}$~${}^{\text{(G2)}}$
\STATE ~
\STATE \COMMENT{Enumeration of rules (Property \ref{pr:rule-enumeration})}
\STATE $\tau \leftarrow \text{Node.LastChanged}$ \COMMENT{Property \ref{pr:rule-enumeration}(a)}
\FOR{$j\leftarrow\tau,\tau+1,\ldots,d$}
	\IF{$\mathit{next}_{j}(\text{Node.Rule.L}[j]) = \text{Node.Rule.U}[j]$}
		\STATE {\bf continue}
	\ENDIF
	\IF{$j>\tau$~{\bf or}~$\text{Node.LowerChanged} = \mathbf{false}$}
		\STATE \COMMENT{Property \ref{pr:rule-enumeration}(c)}
		\STATE $\text{NewNode}\leftarrow\text{(a new node)}$
		\STATE $\text{NewNode.Rule}\leftarrow\text{(copy of Node.Rule)}$
		\STATE $\text{NewNode.Rule.U}[j]\leftarrow \mathit{prev}_{j}(\text{Node.Rule.U}[j])$
		\STATE $\text{NewNode.LastChanged} \leftarrow j$
		\STATE $\text{NewNode.LowerChanged} \leftarrow \mathbf{false}$
		\STATE $\text{NewNode.Instances}\leftarrow \{ i \mid i\in\text{Node.Instances},$
		\STATE \hfill$X_{ij} < \text{NewNode.Rule.U}[j]\}$
		\IF{$\mathrm{OnlyClosed} = \mathbf{true}$}
			\STATE $\text{Closure} \leftarrow \textsc{CheckClosure}(\text{NewNode.Rule}, j)$
			\IF{$\text{Closure}\neq{\bf null}$}
				\STATE $\text{NewNode.Rule} \leftarrow \text{Closure}$
				\STATE \textsc{MakeSubtree}(NewNode)
			\ENDIF
		\ELSE
			\STATE \textsc{MakeSubtree}(NewNode)
		\ENDIF
	\ENDIF
	\STATE ~
	\STATE \COMMENT{Property \ref{pr:rule-enumeration}(b)}
	\STATE $\text{NewNode}\leftarrow\text{(a new node)}$
	\STATE $\text{NewNode.Rule}\leftarrow\text{(copy of Node.Rule)}$
	\STATE $\text{NewNode.Rule.L}[j] \leftarrow \mathit{next}_{j}(\text{Node.Rule.L}[j])$
	\STATE $\text{NewNode.LastChanged} \leftarrow j$
	\STATE $\text{NewNode.LowerChanged} \leftarrow \mathbf{true}$
	\STATE $\text{NewNode.Instances}\leftarrow \{ i \mid i\in\text{Node.Instances},$
	\STATE \hfill$X_{ij} > \text{NewNode.Rule.L}[j]\}$
	\IF{$\mathrm{OnlyClosed} = \mathbf{true}$}
		\STATE $\text{Closure} \leftarrow \textsc{CheckClosure}(\text{NewNode})$
		\IF{$\text{Closure}\neq{\bf null}$}
			\STATE $\text{NewNode.Rule} \leftarrow \text{Closure}$
			\STATE \textsc{MakeSubtree}(NewNode)
		\ENDIF
	\ELSE
		\STATE \textsc{MakeSubtree}(NewNode)
	\ENDIF
\ENDFOR
\STATE \hspace{-1em}{\bf end function}
\end{algorithmic}
}
\end{minipage}
\end{tabular}
\end{algorithm*}

\subsection{Use of mSS for regularization paths} \label{sec:regularization_path}

In this section we present a specific application of mSS: {\em regularization paths}.
A regularization path is to compute the optimal models for multiple $\lambda$'s in \eqref{eq:primal}. Doing so with $L_1$-penalty, it is known to be efficient to compute from large $\lambda$ to small.
Regularization paths are used for model selection (choosing the best $\lambda$ in the prediction performance), controlling the number of active features, faster computations, and so on.

Let us compute the optimal solutions in the order of $\lambda_0$, $\lambda_1$, ..., $\lambda_T$, where $\lambda_0>\lambda_1>\cdots>\lambda_T$.
A basic idea is that, when we compute $\chmSSUB(k)$ for $\lambda_t$ ($t > 0$) in Theorem \ref{thm:pruning}, we use the previously computed optimal solution (i.e., solution for $\lambda_{t-1}$) as the ``primal and dual feasible solution'' $(\hbw^\prime,\chbw^\prime,b^\prime)\in\dom(P_{\lambda}),~\btheta^\prime\in\dom(D_{\lambda})$.
This is reasonable because, to facilitate the pruning, the specified feasible solution has to be near to the true optimal solution. In fact, the duality gap, which determines the radius $r_{\lambda}$ (\S\ref{sec:safe_screening}), tends to be smaller when the specified feasible solution is nearer to the true optimal solution.
We expect that, if $\lambda_{t-1}$ is close to $\lambda_t$, their optimal solutions are also close. Moreover, in such a situation, we can initiate the optimization at $\lambda_t$ from the optimal solution at $\lambda_{t-1}$: it is well-known that this speeds up the optimization (called the {\em warm-start}).

\subsubsection{Computation of $\lammax$ for SRF} \label{sec:lammax_srf}
With the $L_1$ penalty, it is known that $\lammax$, the smallest $\lambda$ when all features are inactive (i.e., $\hbw^*=\bm{0}$ and $\chbw^*=\bm{0}$) is easily computed. 
Thus $\lammax$ is usually set as $\lambda_0$.
With the formulation for SRF, $\lammax$ is known to be computed as:
\begin{align}
&\lammax:=\max\left\{\|\hbZ^{\top}\bm{\phi}\|_{\infty},\|\chbZ^{\top}\bm{\phi}\|_{\infty}\right\}, \label{eq:lammax_srf} \\
&\text{where}~\forall i\in[n] \notag \\
&\phi_i:=
\begin{cases}
y_i-\frac{1}{n}\sum_{i'\in[n]}y_{i'} & (\text{squared~loss}) \\
\frac{1}{1+\exp(y_ic)},~c=\log\frac{\sum_{i':y_{i'}=+1}y_{i'}}{\sum_{i':y_{i'}=-1}y_{i'}} & (\text{logistic~loss})
\end{cases}
. \label{eq:def_phi}
\end{align}
Note that \eqref{eq:lammax_srf} is the maximum among $d+|\cR|$ elements, which is exponentially large.
When we take the maximum for $|\cR|$ elements, in reality, we can prune a part of $|\cR|$ elements in the tree (Algorithm \ref{alg:enumerate_all_closed} in \S\ref{sec:tree}) due to the following property.
By Lemma \ref{lem:rule_bound} (Appendix \ref{sec:proof_pruning}) we have
\begin{align}
&|\chbZ_{:k'}^{\top}\bm{\phi}| \label{eq:lammax_score_k} \\
&\leq\max\left\{\sum_{i:\xi_i\phi_i>0}\xi_ir_k(\boldx_i)\phi_i,-\sum_{i:\xi_i\phi_i<0}\xi_ir_k(\boldx_i)\phi_i\right\} \label{eq:lammax_bound} \\
&:=\max\{p(k),q(k)\},\quad k'\in\cR_{\text{sub}}(k). \notag
\end{align}
Therefore, \eqref{eq:lammax_srf} can be computed for $|\cR|$ elements with pruning as follows:
Let $\lambda_{\mathrm{tmp}}$ be the largest $|\chbZ_{:h}^{\top}\bm{\phi}|$, where $h$ is chosen from the rules enumerated so far. Given a new rule $k$, if $\max\{p(k),q(k)\} < \lambda_{\mathrm{tmp}}$, then all rules in $k'\in\cR_{\text{sub}}(k)$ cannot be larger than $\lambda_{\mathrm{tmp}}$ and therefore we can prune $k$.
To implement this, in Algorithm \ref{alg:enumerate_all_closed} we have only to check the pruning condition above instead of checking $\chmSSUB(k) < 1$.

\subsubsection{Regularization Path Computation Algorithm} \label{sec:path_alg}
Algorithm \ref{alg:path_alg} presents the optimization procedure with the regularization path.
To compute dual feasible solutions, we used the {\em dual scaling} method \cite{bonnefoy2014dynamic,fercoq2015mind}.
After pruning, we used the coordinate descent \cite{tseng2009coordinate} to compute the optimal coefficients.
If we learn GSRF in \S\ref{sec:extension}, we used the block coordinate descent \cite{simon2013sparse} instead.
Note that safe screenings can also be conducted during these optimization algorithms, referred to as {\em dynamic safe screening} \cite{bonnefoy2014dynamic,fercoq2015mind}. The outline of dynamic safe screening in our implementation is presented in Appendix \ref{app:coordinate-descent}.
\begin{algorithm}[t]
   \caption{Regularization Path Computation Algorithm of SRF. Marks (G3) and (G4) denote the processes that we need to modify to apply GSRF (\S\ref{sec:extension}).}
   \label{alg:path_alg}
\begin{algorithmic}
   \STATE {\bfseries Input:} $\{(\bx_i,y_i)\}_{i\in[n]},T,\{\lambda_t\}_{t\in[T]}$
   \STATE {\bfseries Output:} $\{(\hbw^{*(\lambda_t)},\chbw^{*(\lambda_t)},b^{*(\lambda_t)})\}_{t\in[T]}$
   \STATE Compute $\lambda_0$ by \eqref{eq:lammax_srf}~${}^{\text{(G3)}}$
   \FOR{$t\in[T]$}
   \STATE $(\hbw^{'(\lambda_t)},\chbw^{'(\lambda_t)},b^{'(\lambda_t)}) \leftarrow (\hbw^{*(\lambda_{t-1})},\chbw^{*(\lambda_{t-1})},b^{*(\lambda_{t-1})})$
   \STATE Compute $\btheta^{'(\lambda_t)}$ by dual scaling (\S\ref{sec:path_alg})
   \STATE Compute $r_\lambda$ by \eqref{eq:safe_radius}
   \STATE \hfill with $(\hbw^{'(\lambda_t)},\chbw^{'(\lambda_t)},b^{'(\lambda_t)})$ and $\btheta^{'(\lambda_t)}$
   \STATE ~
   \STATE \COMMENT{Compute $\tilde{\cR}^{(\lambda_t)}$ of Property \ref{pr:optimize-for-subset} by Theorem \ref{thm:pruning}} \\
   \STATE Run ``\textsc{Enumerate}'' in Algorithm \ref{alg:enumerate_all_closed} with $\btheta^\prime\gets\btheta^{\prime(\lambda_t)}$
   \STATE \qquad and $r_\lambda$. Let $\tilde{\cR}^{(\lambda_t)} \gets (\tilde{\cR}~\text{in the algorithm})$.
   \STATE ~
   \STATE Solve \eqref{eq:primal_only_active} with $\tilde{\cR}=\tilde{\cR}^{(\lambda_t)}$ by arbitrary algorithm
   \STATE \qquad (e.g., coordinate descent method), and obtain\\
   \STATE \qquad $(\hbw^{*(\lambda_{t})},\chbw^{*(\lambda_{t})},b^{*(\lambda_{t})})$~${}^{\text{(G4)}}$
   \ENDFOR
\end{algorithmic}
\end{algorithm}

%% file: Sec4.tex
\section{Extension: Group Sparsity}
\label{sec:extension}
\begin{figure*}[t]
\vskip 0.2in
\begin{center}
\centerline{\includegraphics[width=1.5\columnwidth]{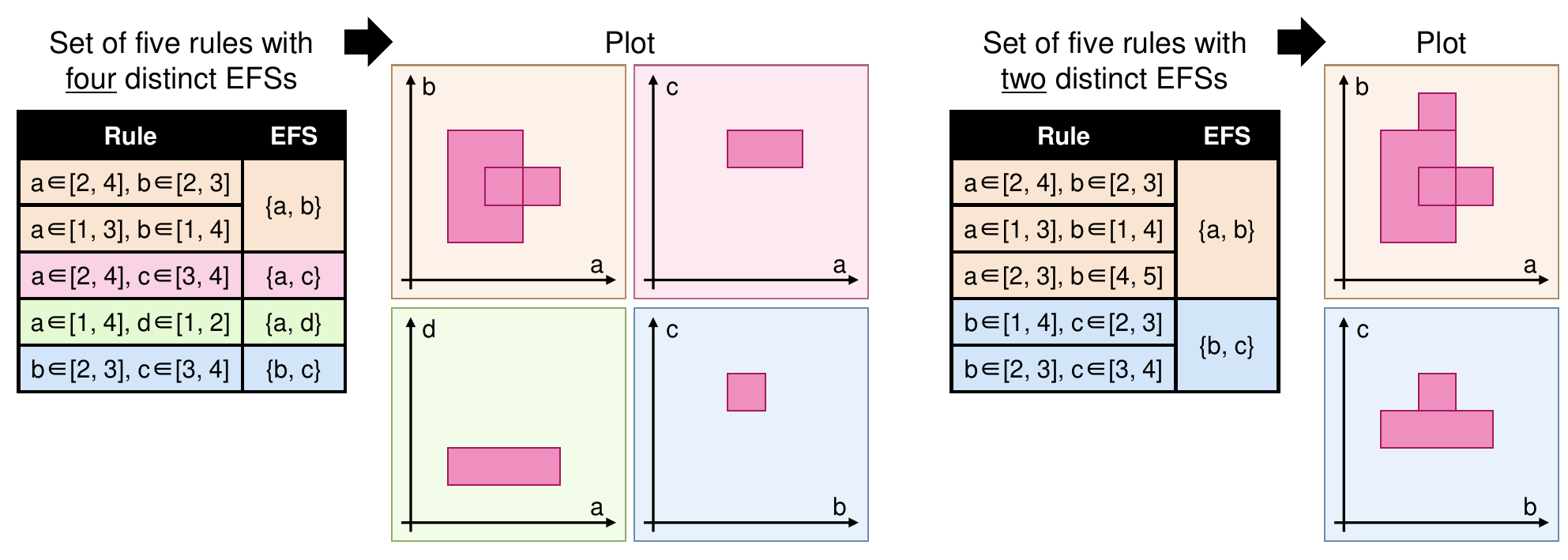}}
\caption{Rules with various combinations of input features, formulated as \emph{effective feature set (EFS)}. Given a set of rules, fewer number of distinct EFSs means the fewer number of plots representing the region the rule is satisfied, that is, easier to interpret.}
\label{fig:EFS}
\end{center}
\vskip -0.2in
\end{figure*}

One of the motivations for introducing rules in prediction models is the interpretability. 
However, if the fitted model consists of a variety of rules with various combinations of input features (i.e., various EFSs defined in \S\ref{sec:pruning-in-enumeration}), it is not easy to interpret (Fig. \ref{fig:EFS}). 
A possible remedy is to introduce \emph{sparse group LASSO (SGL) penalty} \cite{simon2013sparse} so that the active rules in the fitted model have less variety in their input feature combinations. 
In this section, we present an extension of SRF that employs SGL penalty, called \emph{Group Safe RuleFit (GSRF)}, and show that the GSRF can still enjoy the mSS framework.

\begin{figure}[t]
\vskip 0.2in
\begin{center}
\centerline{\includegraphics[width=0.99\columnwidth]{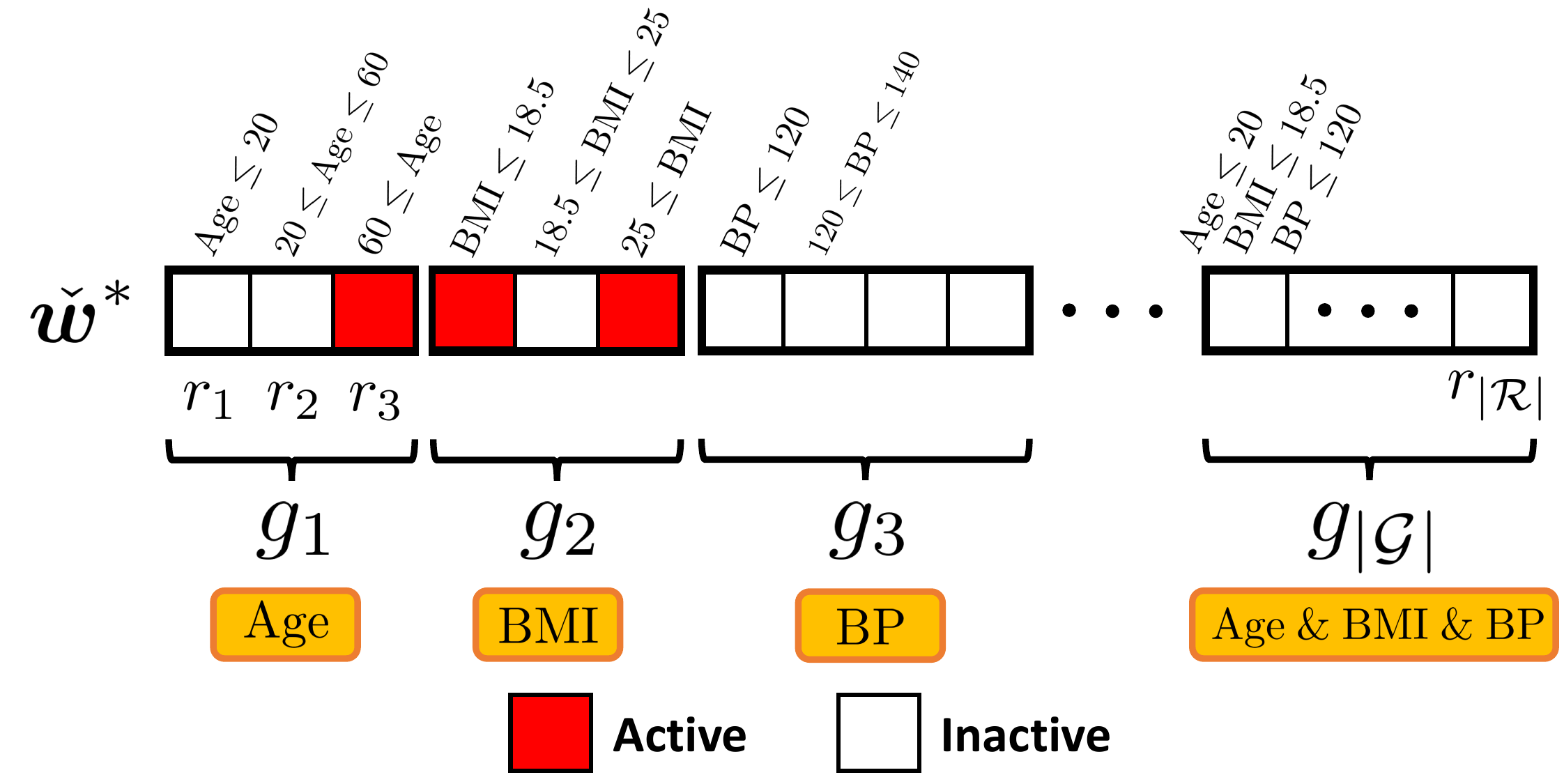}}
\caption{Example of groups in GSRF. Rules in the same group have the same EFS.}
\label{fig:group}
\end{center}
\vskip -0.2in
\end{figure}

Here, we make groups of rules with the same EFS (Fig. \ref{fig:group}).
The SGL penalty produces models with a small number of \emph{active} groups, i.e., rules with less variety in their input feature combinations. 

Let $\cG$ be the set of groups of rules having identical EFSs. 
For a group $g\in\cG$, we define $\bm{a}_g \in \bbR^{|g|}$ as the subvector of $\bm{a}\in\bbR^{|\cR|}$ consisting only of the dimensions in $g$. 
Similarly, we define $\bm{A}_g\in\bbR^{n\times|g|}$ as the submatrix of $\bm{A}\in\bbR^{n\times|\cR|}$ consisting only of the columns in $g$.
\par
The primal problem of GSRF is written as follows:
\begin{align}
&\min_{\hbw\in\bbR^d,\chbw\in\bbR^{|\cR|}, b\in\bbR}P_{\lambda,\htau,\chtau}(\hbw,\chbw,b) \notag \\
&:=\sum_{i\in[n]}\ell_i(y_i,f(\bx_i))+\lambda(\htau\|\hbw\|_1+\Omega(\chbw)), 
\label{eq:primal_gsrf}
\end{align}
where $\Omega(\chbw):=\chtau\|\chbw\|_1+(1-\chtau)\sum_{g\in\cG}\|\chbw_g\|_2$ is the SGL penalty function, and $\chtau\in(0,1)$ is the hyperparameter for controlling the balance between the two penalty terms in $\Omega$ (first term for sparsity of rules, and second term for sparsity of groups). 
Another hyperparameter $\htau>0$ is used to separately control the penalties for $\hbw$ and $\chbw$.
The dual problem of \eqref{eq:primal_gsrf} is written as \cite{ndiaye2016gap}:
\begin{align}
&\max_{\btheta\in\bm{\Delta}_{\htau,\chtau}}D_{\lambda}(\btheta):=-\sum_{i\in[n]}\ell_i^*(-\lambda\theta_i),
	\quad\text{where} \label{eq:dual_gsrf}\\
&\bm{\Delta}_{\htau,\chtau}:=\{\btheta\in\bbR^n:\|\hbZ^{\top}\btheta\|_{\infty}\leq \htau,\Omega^D(\chbZ^{\top}\btheta)\leq 1, \notag \\
&\phantom{\bm{\Delta}_{\htau,\chtau}:=\{}
	\bxi^{\top}\btheta=0,\btheta\in\dom(\ell^*)\}. \nonumber
\end{align}
Here, $\Omega^D:\bbR^{|\cR|}\rightarrow\bbR$ is called the {\em dual norm} of $\Omega$ defined as $\Omega^D(\chbZ^{\top}\btheta):=\max_{g\in\cG}\|\chbZ_g^{\top}\btheta\|_{1-\chtau}$, and $\|\bm{v}\|_{\epsilon}$ ($\epsilon\in(0,1)$, $\bm{v}\in\bbR^d$), referred to as the {\em $\epsilon$-norm}, is the (unique) nonnegative solution of $\sum_{i\in[d]}(|v_i|-(1-\epsilon)\nu)_{+}^2=(\epsilon\nu)^2$ w.r.t. $\nu$ \cite{burdakov1988new,burdakov2001new}. See, for example, Algorithm 4 in \cite{burdakov2001new} for computing the $\epsilon$-norm.
Since the SGL penalty provides both rule-wise sparsity and group-wise sparsity, SS can be applied both to the rule-level and the group-level. 
In the formulations above, the SS conditions are written as follows:
%
\begin{theorem}
\label{thm:group_screening}
Let the loss function of the primal problem \eqref{eq:primal_gsrf} be a convex and $1/\gamma$-smooth function. Given an arbitrary pair of primal and dual feasible solutions $(\hbw',\chbw',b')\in\dom(P_{\lambda,\htau,\chtau}),~\btheta'\in\dom(D_{\lambda})$, and the radius \eqref{eq:safe_radius} computed from \eqref{eq:primal_gsrf} and \eqref{eq:dual_gsrf}, the following group- and rule-level safe screening conditions are available:
\begin{align}
&(\text{{\bf Group}})~\forall g\in\cG,~\mathrm{SS}_\mathrm{UB}^\mathrm{G}(g)<1-\chtau\Rightarrow\chbw_g^*=\bm{0}, \label{eq:group_safe} \\
&\text{where}~\mathrm{SS}_\mathrm{UB}^\mathrm{G}(g) \notag \\
&:=
\begin{cases}
\|\bm{S}_{\chtau}(\chbZ_g^{\top}\btheta')\|_2+r_{\lambda}\|\bm{\Gamma}_g\|_2 & \text{if}\quad\|\chbZ_g^{\top}\btheta'\|_{\infty}>\chtau, \\
(\|\chbZ_g^{\top}\btheta'\|_{\infty}+r_{\lambda}\|\bm{\Gamma}_g\|_2-\chtau)_{+} & \text{otherwise},
\end{cases}
\notag \\
 &(\text{{\bf Rule}})~\forall g\in\cG,~\forall k\in g,~\chSSUB(k)<\chtau\Rightarrow\chw_k^*=0, \label{eq:group_rule_safe}
\end{align}
where $\bm{\Gamma}$ is defined as $\bm{\Gamma}:=\chbZ-\Pi_{<\bm{\Xi}>}(\chbZ)$, $\bm{\Xi}:=[\bxi,\ldots,\bxi]\in\bbR^{n\times |\cR|}$ and $\Pi_{<\bm{\Xi}>}(\chbZ):=\frac{\mathrm{tr}(\bm{\Xi}^{\top}\chbZ)}{\|\bm{\Xi}\|_F^2}\bm{\Xi}$.
In addition, for any $\bm{v}\in\bbR^d$ and $\tau>0$,
$\bm{S}_{\tau}(\bm{v})$ is the elementwise soft thresholding function: $[S_{\tau}(\bm{v})]_j:=\mathrm{sign}(v_j)(|v_j|-\tau)_{+},~j\in[d]$.
\end{theorem}
We show the proof for the group-level condition \eqref{eq:group_safe} in Appendix \ref{sec:proof_group_screening}.
%
The proof is similar to that in \cite{ndiaye2016gap} except the intercept term $b$.
Since the rule-level condition \eqref{eq:group_rule_safe} is almost the same as that in SRF
(the right-hand side is changed from 1 to $\check{\tau}$), 
we can apply the mSS framework to the extended problem \eqref{eq:primal_gsrf} as
\begin{align}
&\forall g\in\cG,~\forall k\in g, \notag \\
&\quad\chmSSUB(k)<\chtau\Rightarrow\chw_{k'}^*=0,~\forall k'\in\cR_{\mathrm{sub}}(k).
\label{eq:prune_group}
\end{align}
The above extension illustrates that our basic idea on mSS can be generally applied to wide class of penalized empirical risk minimization problems.
%

\subsection{Regularization paths for GSRF} \label{sec:lammax_gsrf}
To implement the regularization paths for GSRF like in \S\ref{sec:regularization_path} for SRF,
we need to compute $\lammax$ for the SGL penalty.
Then the other processes are almost the same as the SRF case
(see Algorithm \ref{alg:path_alg}).

For the SGL penalty, $\lammax$ is computed as follows:
\begin{align}
&\lammax:=\max\left\{\htau^{-1}\|\hbZ^{\top}\bm{\phi}\|_{\infty},\Omega^D(\chbZ^{\top}\bm{\phi})\right\}, \notag \\
&\text{where}~\Omega^D(\chbZ^{\top}\bm{\phi})=\max_{g\in\cG}\|\chbZ_g^{\top}\bm{\phi}\|_{1-\chtau}.
\label{eq:lammax_dual_norm}
\end{align}
To compute \eqref{eq:lammax_dual_norm}, it is difficult to have an inequality corresponding to \eqref{eq:lammax_bound} in SRF because we use the $\epsilon$-norm for GSRF.
Here we instead compute an upper bound of $\lammax$, denoted by $\tilde{\lambda}_{\max}$, via an upper bound of $\Omega^D(\chbZ^{\top}\bm{\phi})$. Let us derive the upper bound, denoted by $\UB(\Omega^D(\chbZ^{\top}\bm{\phi}))$. Then we have
\begin{align}
\lammax &\leq \max\left\{\htau^{-1}\|\hbZ^{\top}\bm{\phi}\|_{\infty},\UB(\Omega^D(\chbZ^{\top}\bm{\phi}))\right\} \label{eq:ub_omega_D} \\
&:=\tilde{\lambda}_{\max}. \notag
\end{align}
The $\UB(\Omega^D(\chbZ^{\top}\bm{\phi}))$ in \eqref{eq:ub_omega_D} can be calculated as follows:
\begin{theorem} \label{thm:dual_norm_bound}
For $\Omega^D(\chbZ^{\top}\bm{\phi})$,
\begin{align*}
\Omega^D(\chbZ^{\top}\bm{\phi}) \leq \UB(\Omega^D(\chbZ^{\top}\bm{\phi})):=\chtau^{-1} \|\chbZ^{\top}\bm{\phi}\|_{\infty}.
\end{align*}
\end{theorem}
See Appendix \ref{app:dual_norm_bound} for the proof.
Then we use $\lambda_0\gets\tilde{\lambda}_{\max}$ below for the GSRF regularization path:
\begin{align}
\tilde{\lambda}_{\max}:=\max\left\{\htau^{-1}\|\hbZ^{\top}\bm{\phi}\|_{\infty},\chtau^{-1}\|\chbZ^{\top}\bm{\phi}\|_{\infty}\right\}
\label{eq:lammax_gsrf}
\end{align}

\subsection{Algorithm of GSRF} \label{sec:algorithm_gsrf}

Algorithms \ref{alg:enumerate_all_closed} and \ref{alg:path_alg}, which are described for SRF, are almost the same for GSRF, since the pruning condition is almost the same.
After modifying the following processes, Algorithms \ref{alg:enumerate_all_closed} and \ref{alg:path_alg} works for GSRF.
\begin{itemize}
\item Algorithm \ref{alg:enumerate_all_closed}, (G1): If we apply GSRF, as additional inputs for Algorithm \ref{alg:enumerate_all_closed}, we need to specify $\htau$ and $\chtau$ in \eqref{eq:primal_gsrf}.
\item Algorithm \ref{alg:enumerate_all_closed}, (G2): To derive the pruning condition for SRF, we use Theorem \ref{thm:pruning}. For GSRF we need to use \eqref{eq:prune_group} instead. Similarly, for the screening condition we use \eqref{eq:screening} for SRF while \eqref{eq:group_rule_safe} for GSRF.
\item Algorithm \ref{alg:path_alg}, (G3): To compute $\lambda_0$ for SRF, we use \eqref{eq:lammax_srf}. For GSRF we need to use \eqref{eq:lammax_gsrf} instead.
\item Algorithm \ref{alg:path_alg}, (G4): To find the optimal model parameters $(\hbw^{*(\lambda_{t})},\chbw^{*(\lambda_{t})},b^{*(\lambda_{t})})$, we can use the coordinate descent method for SRF. For GSRF we need to use the block coordinate descent method instead.
\end{itemize}

%% file: Sec5.tex
\section{Experiments} \label{sec:experiment}
In this section we examine the performances and advantages of SRF by six experiments.
From Exp.1 to Exp.3 we compare the proposed SRF with conventional methods.
\begin{itemize}
\item In Exp.1 we compare the prediction accuracy of SRF with other methods with guaranteed optimality.
	Specifically, we compare SRF with the linear model (without rule terms)
	as a simpler model, and the kernel linear model as a non-linear but less interpretable model.
\item In Exp.2, we examine the optimality and the stability of learned rules in the
	well-known conventional methods (RuleFit and Random Forest).
	Since these methods are suboptimal (\S\ref{sec:related}) and randomized,
	we consider that SRF is advantageous since it can learn a model with guaranteed optimality.
	Therefore we examine how the learned rules by these two methods are suboptimal and unstable.
\item In Exp.3, we compare SRF with REPR, a conventional rule model with guaranteed optimality
	(\S\ref{sec:related}), in the computational cost.
\end{itemize}

On the other hand, from Exp.4 to Exp.6 we examine SRF in various aspects.
\begin{itemize}
\item In Exp.4, we examine the scalability of SRF, specifically, growth of computational costs and number of enumerated rules for various parameters (number of discretizations $M$, number of input features $d$ and the ``max\_efs'').
\item In Exp.5, we examine the effect of the introduction of linear terms (see \S\ref{sec:sparse_learning})
	in the prediction accuracy and the number of rules.
	We expect that, if we employ linear terms, then rules that can be substituted by linear terms
	are not used in the learned rules; so we expect that the number of rules are reduced.
\item In Exp.6, we examine the effect of the group sparsity constraint by GSRF (see \S\ref{sec:extension})
	compared to original SRF. As a result of using GSRF, we expect that we can retrieve
	more interpretable set of rules, at the expense of a small amount of the prediction accuracy.
\end{itemize}

Tables \ref{tab:datasets124} and \ref{tab:datasets3} show the benchmark datasets used in the experiments\footnote{
We selected benchmark datasets with moderate dimension $d$ partly because a rule is easy to interpret when the number of input features involved in the rule is not too large. Furthermore, as specified as \eqref{eq:num-all-rules}, even with effective use of mSS and SS, the computational cost of SRF increases rapidly with $d$.
} (see Appendix \ref{sec:preprocesses} for details of the preprocessing employed in the experiments).
We use the squared loss for regression and the logistic loss for binary classification. 
The (G)SRF is implemented by combining mSS and (ordinal) SS. 
After we obtained a set of rules $\tilde{\cR}$ by mSS, we fit the sparse rule model
(``Solve \eqref{eq:primal_only_active} with ...'' in Algorithm 2).
As stated in \S\ref{sec:path_alg}, we use the (block) coordinate descent algorithm to fit the model
until the stopping criterion is met.
During the fitting, we apply SS (not mSS) to make the fitting faster.
Details are presented in Appendix \ref{app:coordinate-descent}.

%
%

The (G)SRF was implemented in C++, and executed on a computer with CPU Intel(R) Xeon(R) CPU E5-2687W v4 (3.00GHz) and RAM 256GB.
\begin{table}[tp]
\begin{center}
\caption{Datasets used in experiments except for Exp.3 (available at UCI machine learning repository \cite{Dua:2017}).}
\label{tab:datasets124}
\begin{tabular}{cccc}
\hline
Task & Name & $n$ & $d$ \\
\hline
Regression
& {\tt CONCRETE} & 1,030 & 8 \\
& {\tt ABALONE} & 4,177 &  10 \\
& {\tt WHITEWINE} & 4,898 & 11 \\
& {\tt CASP} & 45,730 &  9 \\
\hline
Binary
& {\tt CONTRACEPTIVE} & 1,473 & 17 \\
classification
& {\tt REDWINE} & 1,599 &  11 \\
& {\tt PAGEBLOCKS} & 5,473 &  10 \\
& {\tt MAGIC} & 19,020 & 10 \\
\hline
\end{tabular}
\end{center}
\end{table}

\begin{table}[tp]
\begin{center}
\caption{Datasets used in Exp.3 (all for regressions, available at UCI machine learning repository \cite{Dua:2017}).
The values of $\delta$, which controls the discretization (Appendix \ref{sec:discretize-interval}), are set to be the same as those in the existing method REPR \cite{eckstein2017rule}.}
\label{tab:datasets3}
\begin{tabular}{ccccc}
\hline
Name & $n$ & $d$ & $|\cR|$ & $\delta$ \\ \hline
{\tt SERVO} & 167 & 10 & 9.8e05 & 0 \\
{\tt YACHT} & 308 & 6 & 2.6e10 & 0 \\
{\tt COOL} & 768 & 8 & 1.1e10 & 0.005 \\
{\tt HEAT} & 768 & 8 & 1.1e10 & 0.005 \\
{\tt AIRFOIL} & 1,503 & 5 & 1.0e11 & 0.005 \\ \hline
\end{tabular}
\end{center}
\end{table}
\subsection*{Exp.1: Comparison with Linear and Nonlinear (kernel) Models}

In this experiment we confirm the advantage of introducing rules in prediction modeling. 
(A) First we compare the accuracy of SRF with an ordinary $L_1$-regularized linear model to confirm that SRF represents the data better.
(B) Then we compare the accuracy of SRF with the optimality in the learning and a nonlinear (and less interpretable) model to show that SRF is as accurate as the nonlinear model.
For regression, we used LASSO for (A) and kernel ridge regression for (B). 
For binary classification, we used $L_1$-regularized logistic regression for (A) and kernel SVM for (B).
See Appendix \ref{chap:implementation-details} for details of these baseline methods.

For SRF we set the number of discretizations as $M\in\{5,8\}$ (see Appendix \ref{sec:discretize-quantile}), and used only the rules with $\text{max\_efs}\leq 3$ (see \S\ref{sec:pruning-in-enumeration}).
%
%
%
For each dataset, we took a third of the instances as the test set, and conducted two-fold cross-validation with the rest of the instances for selecting the best hyperparameters (the regularization parameter $\lambda$ for (A) and SRF, while the hyperparameters of kernel functions for (B)).
For $\lambda$, we selected it from $\lammax$ to $0.01\lammax$, divided into 100 $\lambda$'s in logarithmic scale.
See Appendix \ref{chap:hyperparameters} for other details.
%
As measures of accuracy, we used the MSE (mean squared error) for regression and AUC (area under curve) for classification.

We show the results in Table \ref{tab:result1}.
In many datasets, the proposed SRF was much more accurate (smaller MSE or higher AUC) than the linear model (A), while the accuracy of SRF is comparative to nonlinear models (B).
The results indicate that the sparse rule model is favorable in terms of both prediction accuracy and interpretability.

\begin{table}[tp]
\begin{center}
\caption{Accuracy of the proposed SRF, linear models (LASSO, Logistic regression), and nonlinear models (Kernel Ridge, Kernel SVM). {\bf Bold} numbers indicate the best results.}
\label{tab:result1}
{\tabcolsep=1mm
\begin{tabular}{ccccc}
\hline
\multicolumn{5}{c}{MSE (regression datasets)} \\
\hline
        & SRF   & SRF   &       & Kernel \\
Dataset & {\small $M=5$} & {\small $M=8$} & LASSO & Ridge  \\
\hline
{\small {\tt CONCRETE}}  & 0.149 & {\bf 0.104} & 0.439 & 0.138 \\
{\small {\tt ABALONE}}   & 0.397 & 0.391 & 0.414 & {\bf 0.373} \\
{\small {\tt WHITEWINE}} & 0.579 & {\bf 0.568} & 0.680 & 0.573 \\
{\small {\tt CASP}}      & 0.630 & 0.599 & 0.735 & {\bf 0.375} \\
\hline
\multicolumn{5}{c}{AUC (binary classification datasets)} \\
\hline
        & SRF   & SRF   &          & Kernel \\
Dataset & {\small $M=5$} & {\small $M=8$} & {\small Logistic} & SVM \\
\hline
{\small {\tt CONTRACEPTIVE}} & 0.729 & {\bf 0.749} & 0.652 & 0.737 \\
{\small {\tt REDWINE}}       & 0.826 & 0.821 & 0.729 & {\bf 0.830} \\
{\small {\tt PAGEBLOCKS}}    & {\bf 0.986} & 0.983 & 0.811 & 0.985 \\
{\small {\tt MAGIC}}         & 0.919 & 0.922 & 0.749 & {\bf 0.927} \\
\hline
\end{tabular}
} 
\end{center}
\end{table}

\subsection*{Exp.2: Comparison with RuleFit and Random Forest}

In this experiment we observe the advantage of SRF over RuleFit \cite{friedman2008predictive} and Random Forest \cite{breiman2001random} as the methods for selecting rules. 
As discussed in \S \ref{sec:introduction}, these existing methods are suboptimal in the sense that they do not consider all possible rules as candidates. 
In addition, these existing methods are unstable in the sense that the selected rules can change depending on the random seed, which is not preferred from the viewpoint of interpretability. 
To illuminate these points, we computed the similarity scores between two trained models. 
First, to confirm their suboptimality, we computed the scores between the existing methods and SRF. 
Next, to confirm their instability, we computed the scores for each of the existing methods with different random seeds. 
%
\par
Given two sets of rules
$R = \{r_1, r_2, \dots, r_N\}$ and
$R^\prime = \{r^\prime_1, r^\prime_2, \dots, r^\prime_{N^\prime}\}$,
the similarity between them (not symmetric) is defined as
\begin{align}
\mathit{Sim}(R, R^\prime) := \frac{1}{N} \sum_{i\in[N]} \max_{j\in[N^\prime]} \mathit{Jac}(V_{r_i}, V_{r^\prime_j}), \label{eq:rule-simiarlity}
\end{align}
where $\mathit{Jac}(A, B) := |A\cap B|/|A\cup B|$ is the Jaccard index for two sets, and $V_{r_i}$ is the set of ``boxes'' bordered by $\bm{\omega}$ (dotted lines in Fig. \ref{fig:segment}). See Appendix \ref{app:rule-similarity} for details.
%
This implies that $\mathit{Sim}(R, R^\prime) = 1$ if $R$ and $R^\prime$ are identical, whereas $\mathit{Sim}(R, R^\prime) = 0$ if $R$ and $R^\prime$ has no overlap.
%
\par
For each of the methods, we retrieved approximately 100 rules:
For SRF, we selected the model with the number of rules nearest to 100 in the regularization path.
For random forests, we composed 10 trees with 10 rules (leaves in a decision tree).
For RuleFit, we first retrieved approximately 2000 rules and then took a regularization path to find the model with the number of rules nearest to 100. 
Moreover, to examine randomness, random forests and RuleFit are run with 100 different random seeds.
See Appendix \ref{chap:implementation-details} for implementation details of existing methods,
and Appendix \ref{chap:hyperparameters} for the detailed configuration of the regularization path.
In this experiment we set the number of discretizations as $M = 3$. So that the setups were common between methods, random forests and RuleFit are also conducted with the same discretization.
\par
We show the results in Table \ref{tab:result2}.
We see that, from the columns ``RForest-SRF'' and ``RuleFit-SRF'', the rules retrieved by the existing methods diverge from the optimal result by SRF.
In addition, from the columns ``RForest-RForest'' and ``RuleFit-RuleFit'', rules retrieved by these methods are unstable except for the cases of similarity 0.8 or higher with the random forests for regression datasets.
The results indicate that it is difficult to properly interpret the rules selected by these existing methods since they are suboptimal and unstable. 
In contrast, the rules selected by SRF are straightforward to interpret since they are optimal (in a certain sense) and do not change.
%
\begin{table}[tp]
\begin{center}
\caption{Average of similarity scores between methods. The value 1 means that two sets of rules between two methods are identical while 0 that two sets have no overlap. Averages are taken from $100$ trials for ``RForest-SRF'' and ``RuleFit-SRF'' (100 randomized results versus a non-randomized result), while ${}_{100} P_2 = 9900$ cases for ``RForest-RForest'' and ``RuleFit-RuleFit'' (since $\mathit{Sim}$ is not symmetric).}
\label{tab:result2}
{\tabcolsep=1mm
\begin{tabular}{ccccc}
\hline
        & {\small RForest-} & {\small RForest-} & {\small RuleFit-} & {\small RuleFit-} \\
Dataset & {\small SRF}      & {\small RForest}  & {\small SRF}      & {\small RuleFit} \\
\hline
{\small {\tt CONCRETE}}      & $0.362$ & $0.874$ & $0.455$ & $0.614$ \\
{\small {\tt ABALONE}}       & $0.227$ & $0.916$ & $0.273$ & $0.488$ \\
{\small {\tt WHITEWINE}}     & $0.276$ & $0.855$ & $0.447$ & $0.584$ \\
{\small {\tt CASP}}          & $0.315$ & $0.961$ & $0.320$ & $0.525$ \\
\hline
{\small {\tt CONTRACEPTIVE}} & $0.211$ & $0.316$ & $0.252$ & $0.436$ \\
{\small {\tt REDWINE}}       & $0.243$ & $0.397$ & $0.356$ & $0.527$ \\
{\small {\tt PAGEBLOCKS}}    & $0.240$ & $0.379$ & $0.331$ & $0.490$ \\
{\small {\tt MAGIC}}         & $0.272$ & $0.447$ & $0.472$ & $0.653$ \\
\hline
\end{tabular}
} 
\end{center}
\end{table}

\subsection*{Exp.3: Comparison with REPR}

In this experiment, we compare the computational cost of SRF with REPR \cite{eckstein2017rule}.
First of all, both SRF and REPR employ the same formulation of the learning \eqref{eq:primal}
(although REPR is applicable only to regression),
that is, the final result of the learned predictor is the same between SRF and REPR.
We note that, since the computational cost for REPR is measured for the retrieval of first 100 rules in the paper of REPR, we compare the cost with that by SRF when it provides 100 rules.

We compared the number of accessed nodes (which mainly affects the computational cost; for SRF it is the number of enumerated rules in Algorithm \ref{alg:enumerate_all_closed}) and the actual computation time.
Unfortunately, the code for REPR is not available in public (authors of this paper contacted the authors of REPR).
Thus we compared the computational cost of SRF with the numbers stated in the paper of REPR by considering the same configurations as much as possible: discretization, number of instances in the training set, etc. (details in Appendix \ref{app:setups-REPR}).
Since the paper of REPR shows the computational cost of retrieving 100 rules,
we trained SRF with a regularization path and stopped it when 100 or more rules were retrieved.
We conducted this for 10 random training sets and took the average.
Note that, according to the paper of REPR, REPR used some heuristics for stopping the algorithm before being completely optimized ({\S}5 in \cite{eckstein2017rule}), and employed parallel computations for some costly computations.
In the current experiment, SRF did not employ such techniques, i.e., we thus conjecture that SRF was run under a severer setup than REPR.

Table \ref{tab:result3} shows the results. Even under a severer setup, SRF ran much quicker than REPR, although the number of accessed nodes is larger than that of REPR for some datasets.
The results indicate that SRF is computationally more efficient than REPR because the former can retrieve multiple rules simultaneously\footnote{
As stated in \S\ref{sec:path_alg}, after retrieving the candidates of active rules $\tilde{{\cal R}}^{(\lambda_t)}$ (Algorithm \ref{alg:path_alg}), we run a coordinate descent with $|\tilde{{\cal R}}^{(\lambda_t)}|$ variables at once to retrieve rules. Here, ``\#nodes'' in Table 5 for SRF is the total number of rules of multiple $\tilde{{\cal R}}^{(\lambda_t)}$'s.
},
whereas the latter can only retrieve rules one by one.

\begin{table}[tp]
\begin{center}
\caption{Computational cost (time in seconds and number of accessed nodes in the tree) of SRF and REPR. {\bf Bold} numbers indicate the better results.}
\label{tab:result3}
{\tabcolsep=1mm
\begin{tabular}{cccccc} \hline
 & & \multicolumn{2}{c}{SRF} & \multicolumn{2}{c}{REPR} \\
Dataset & $|\mathcal{R}|$ & \#nodes & Time & \#nodes & Time \\ \hline
{\tt SERVO} & 9.8e05 & {\bf 8.7e03} & {\bf 1.02} & 3.6e04 & 11 \\
{\tt YACHT} & 2.6e10 & 1.2e05 & {\bf 7.01} & {\bf 5.3e04} & 36 \\
{\tt COOL} & 1.1e10 & {\bf 9.9e04} & {\bf 5.64} & 3.7e05 & 288 \\
{\tt HEAT} & 1.1e10 & {\bf 8.7e04} & {\bf 7.04} & 3.9e05 & 222 \\
{\tt AIRFOIL} & 1.1e11 & 3.2e06 & {\bf 591.4} & {\bf 5.1e05} & 7,984 \\ \hline
\end{tabular}
} 
\end{center}
\end{table}
\subsection*{Exp.4: Scalability of SRF}

\begin{table*}[tp]
\caption{Parameter setups for Exp.4: Scalability of SRF. $d^\prime$ denotes the number of input features in the original dataset.}
\label{tb:num-rules-param}
\begin{center}
\begin{tabular}{c|ccc}
\hline
Setup & \#discretizations $M$ & \#input features $d$   & ``max\_efs'' \\
\hline
(A)   & $2, 3, \dots, 8$      & $d^\prime$             & 3 \\
(B)   & $3$                   & $3, 4, \dots d^\prime$ & $d^\prime$ \\
(C)   & $3$                   & $d^\prime$             & $2, 3, \dots, d^\prime$ \\
\hline
\end{tabular}
\end{center}
\end{table*}

\begin{figure*}[tp]
\begin{center}
\includegraphics[width=0.8\hsize]{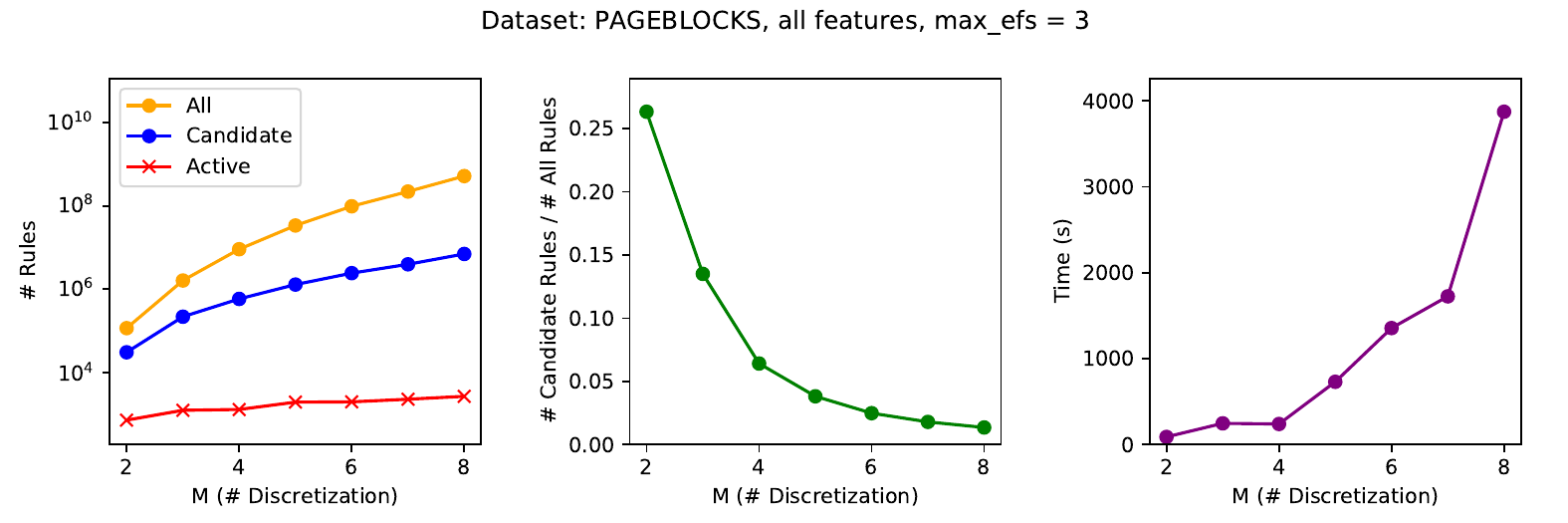}\\
(A) Changing the number of discretizations $M$

~

\includegraphics[width=0.8\hsize]{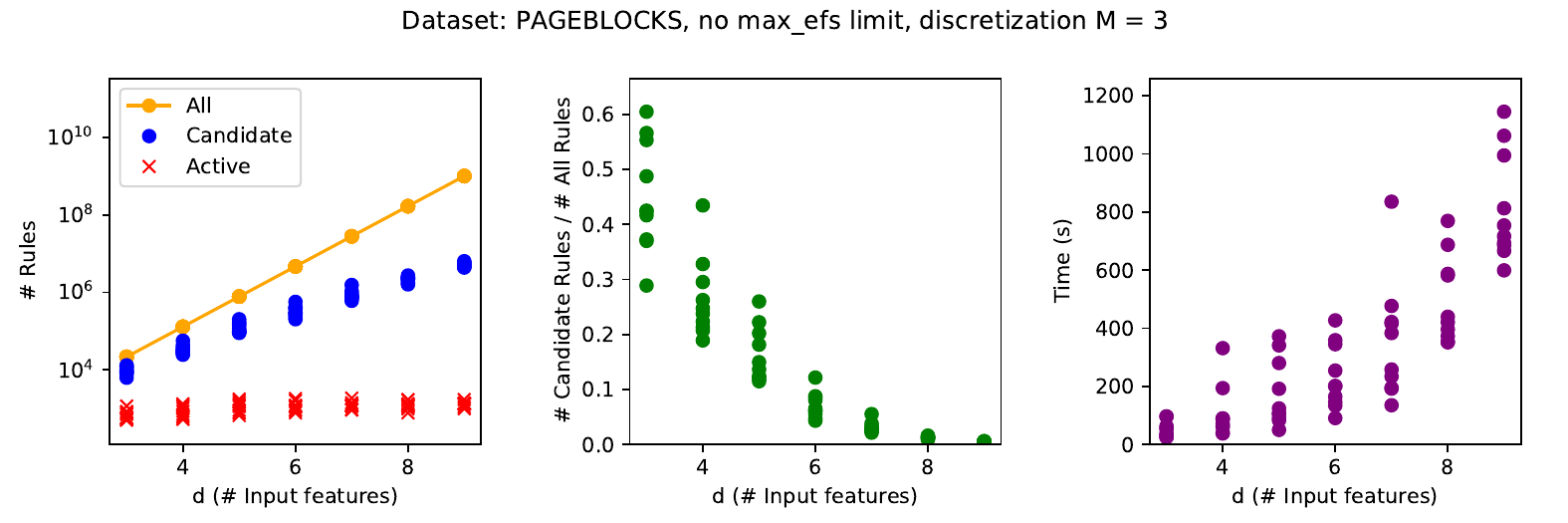}\\
(B) Changing the number of input features $d$

~

\includegraphics[width=0.8\hsize]{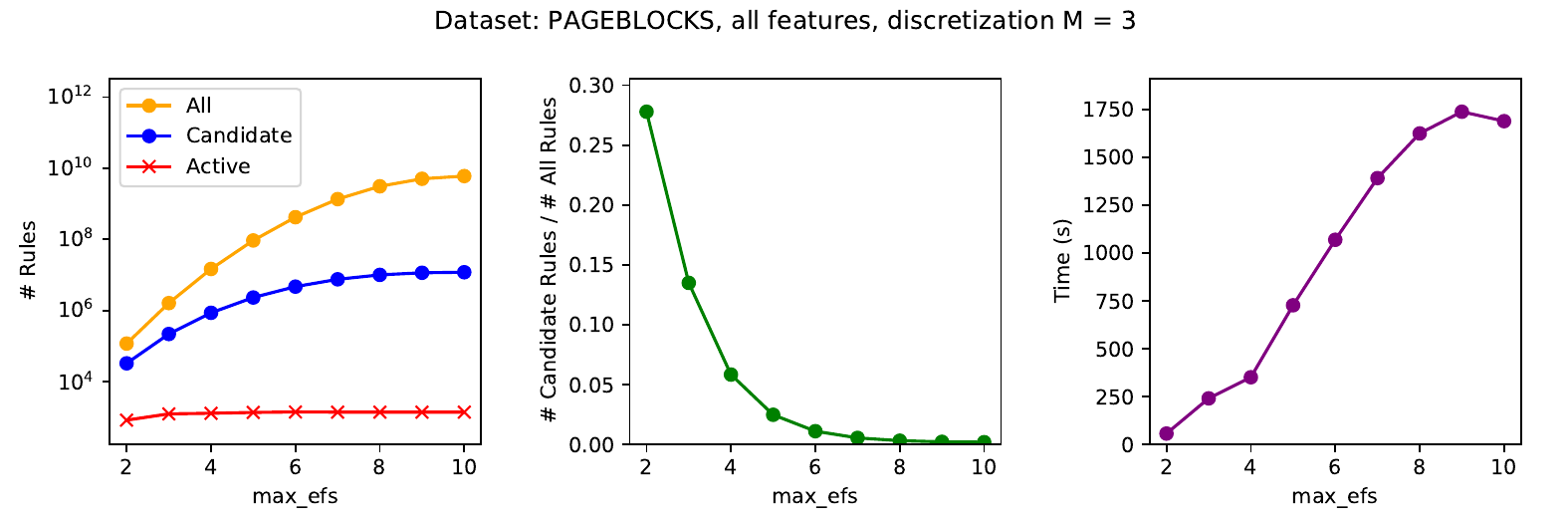}\\
(C) Changing ``max\_efs''
\end{center}
\caption{Number of enumerated rules and computation times by SRF for {\tt PAGEBLOCKS} dataset. The left is the number of all possible rules, of candidate active rules identified by SRF, and of rules truly active at last. The center is the ratios of the number of former two rules. The right is the computation times. Since (B) is tried for several random choices of input features (Appendix \ref{app:random-subset-of-features}), there are multiple points (except for ``all possible rules'' in yellow) for each $M$, $d$ and ``max\_efs''.}
\label{fig:exp4-redwine}
\end{figure*}

In this experiment, to evaluate the scalability of SRF, we compute the number of rules enumerated in SRF.
Specifically, we change the three parameters to evaluate this:
(A) the number of discretizations $M$,
(B) the number of input features $d$, and
(C) ``max\_efs'' in \S\ref{sec:pruning-in-enumeration}.
As discussed in \eqref{eq:num-all-rules}, $|\bm{\omega}^{(j)}| = M+1$ and $d$ largely affects the number of input features.
In addition, as is the case for many pattern mining algorithms, ``max\_efs'' also largely affects.
Precisely, in case ``max\_efs'' is specified, the number of all possible rules can be computed as follows\footnote{
Different from the expression $|\bm{\omega}^{(j)}|(|\bm{\omega}^{(j)}|-1)/2$ in $\prod_{j\in[d]}$ in \eqref{eq:num-all-rules}, that in $\prod_{j\in E}$ in \eqref{eq:num-rules-efs} is subtracted by 1. This is because the former counts all possible intervals for the $j$th input feature including $[-\infty, +\infty]$, which EFS should not count (see \eqref{eq:EFS}).}, which can be increased greatly when $\text{max\_efs}$ is increased:
\begin{align}
\sum_{E\subseteq[d],~0\leq|E|\leq\text{max\_efs}}\Biggl[ \prod_{j\in E}\frac{|\bm{\omega}^{(j)}| (|\bm{\omega}^{(j)}|-1)}{2} - 1 \Biggr] \label{eq:num-rules-efs}
\end{align}

For each of these settings, 
we compared the numbers of (i) all rules \eqref{eq:num-rules-efs},
of (ii) candidates of active rules $\tilde{\cR}$ by Algorithm \ref{alg:enumerate_all_closed},
and of (iii) the rules finally active (nonzero elements in $\chbw^*$ after solving \eqref{eq:primal}).
Here, the most important point is how much (ii) is reduced compared to (i), since (ii) mainly reflects the computational cost.
Moreover, we also measured the running time.
The parameter setups for the settings (A) to (C) are presented in Table \ref{tb:num-rules-param}.
Here, for (C), we tried several times of random choices of input features, and did not run for {\tt CONTRACEPTIVE} dataset. See Appendix \ref{app:random-subset-of-features} for details.
For these setups, we calculate SRF for the regularization path of setting $\lambda$ from $\lammax$ to $0.01\lammax$, divided into 100 $\lambda$'s in logarithmic scale. As a result, we see the results of (ii) and (iii) by the total of 100 cases, while (i) by just multiplying \eqref{eq:num-rules-efs} by 100 times.

The results are presented in Figures \ref{fig:exp4-redwine} for one of the datasets {\tt PAGEBLOCKS}, and all results are in Appendix \ref{app:exp4-results-all}.
For almost all datasets we can see that, although the number of enumerated rules by SRF ((ii), blue plots) are increasing for the increase of $M$, $d$ and ``max\_efs'', the ratio to the all possible rules ((ii)/(i), green plots) are as small as 2\% or less.
In addition, for all of the setups (A) to (C), we can see that the ratio is decreasing against the parameters ($M$, $d$ and ``max\_efs'', respectively). This means that SRF can suppress also the growth of enumerated rules compared to the growth of all possible rules.
We can see that the truly active rules ((iii), red plots) are fewer than those enumerated by SRF ((ii), blue plots).

\subsection*{Exp.5: Comparison of SRF with and without linear terms}

\begin{figure}[tp]
\begin{center}
\includegraphics[width=\hsize]{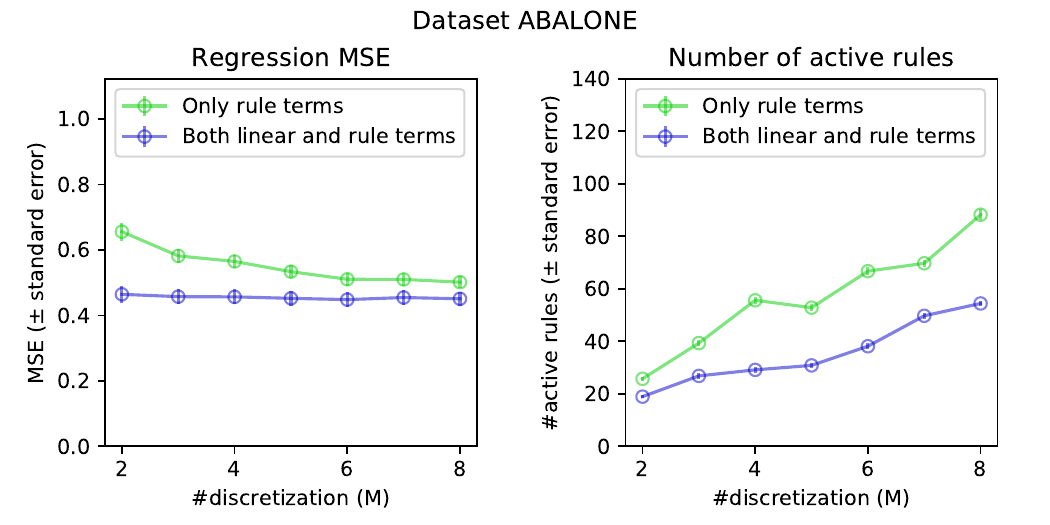}\\
\includegraphics[width=\hsize]{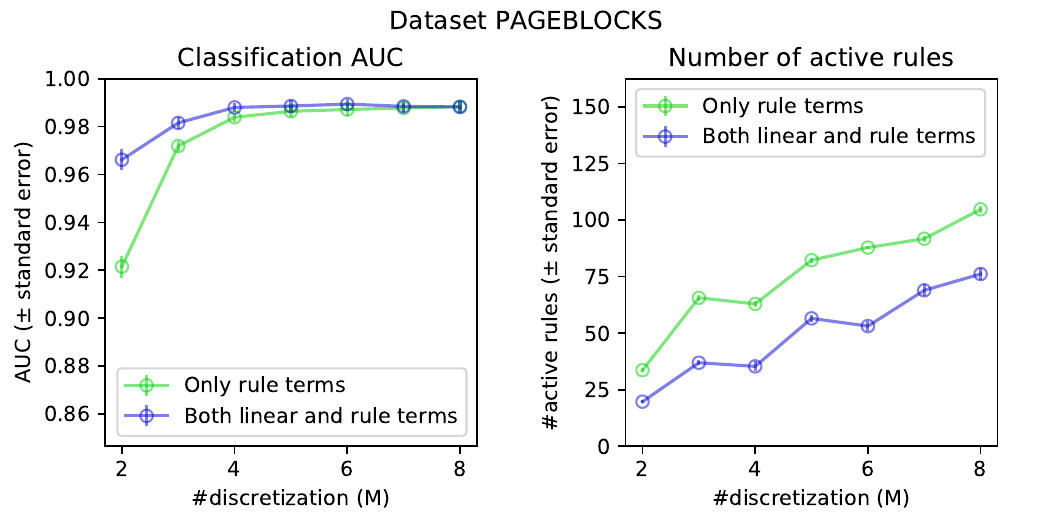}
\end{center}
\caption{Comparisons in prediction accuracy and number of rules between the models with and without linear terms for {\tt ABALONE} (regression) and {\tt PAGEBLOCKS} (classification) datasets. The left is the prediction accuracy (MSE for regressions, AUC for binary classifications), and the right is the number of active rules (rule $k$ such that $\chw_k\neq 0$). The standard errors (error bars) are taken for ten randomized trials.}
\label{fig:exp5-abalone-redwine}
\end{figure}

In this experiment we compare SRF with and without linear terms (see \S\ref{sec:sparse_learning}) to examine how the prediction accuracy changes.
We compare prediction accuracy of these two models under the same setup as Exp.1 except that
\begin{itemize}
\item the number of discretization $M$ is changed from $2$ to $8$, and
\item ten randomized datasets are prepared by the manner of ten-fold cross-validation in order to see the stability of the results.
\end{itemize}

The results are presented in Figures \ref{fig:exp5-abalone-redwine} for two of the datasets {\tt ABALONE} (regression) and {\tt PAGEBLOCKS} (binary classification), and all results are in Appendix \ref{app:exp5-results-all}.
We can find that SRF with linear terms provided higher accuracy (lower MSE or higher AUC) for most of the cases.
Although predictions with only linear terms is not so accurate compared to SRF (see ``LASSO'' and ``logistic regression'' in Exp.1),
we can confirm that including linear terms to rule terms makes predictions better.
In addition, to examine the effect of the linear terms on the model,
we plotted the number of active rules (rule $k$ such that $\chw_k\neq 0$; \S\ref{sec:safe_screening}).
We can see that, for most of the cases, the model without linear terms has more active rules by tens than that with linear terms.
This implies that we need to include many active rules to replace linear terms, and therefore linear terms are helpful in the viewpoint of interpretability.

\subsection*{Exp.6: Comparison of SRF and Group SRF (GSRF)} 

In this experiment we compare SRF and its extension GSRF to examine how the number of distinct EFSs (abbreviated as \#dEFS) is reduced for interpretability (see \S \ref{sec:extension}).

For each dataset we took one-third of the instances as the test set and the rest as the training set. Then we trained SRF and GSRF with a regularization path and stopped the algorithm when we retrieved a model with 100 or more rules (Appendix \ref{chap:hyperparameters}). Finally, we computed \#dEFS in the approximately 100 rules and the prediction accuracy (same as Exp.1). For simplicity of comparisons and computational costs, we set the number of discretization as $M = 3$ for all datasets, and retrieved only closed rules (\S\ref{sec:tree}). We fixed the GSRF-specific parameters as $\htau=\chtau=0.8$. We conducted the above with 10 random training/test sets and took the average.

Table \ref{tab:result4} shows the results.
In contrast to SRF that often retrieved a set of rules with almost completely distinct EFS's (\#dEFS = 100),
GSRF could reduce \#dEFS into 40 or less except for \#dEFS$\approx$67 for ``{\tt CONTRACEPTIVE}'' dataset.
Moreover, we can confirm that GSRF does not negatively impact on prediction accuracy so much.
The results indicate that a sparse-group penalty could be useful for results which are easier to interpret without sacrificing prediction accuracy.

\begin{table}[tp]
\begin{center}
\caption{Averages of \#dEFS and accuracy measures with the proposed SRF and GSRF when (approximately) 100 rules are retrieved. {\bf Bold} numbers indicate the better results.}
\label{tab:result4}
{\tabcolsep=1mm
\begin{tabular}{ccccc}
\hline
Regression & \multicolumn{2}{c}{SRF} & \multicolumn{2}{c}{GSRF} \\
Dataset         & \#dEFS & MSE   & \#dEFS & MSE \\
\hline
{\tt CONCRETE}  & 75.0   & {\bf 0.158} & {\bf 31.9}   & 0.167 \\
{\tt ABALONE}   & 64.6   & {\bf 0.455} & {\bf 18.3}   & 0.464 \\
{\tt WHITEWINE} & 93.2   & {\bf 0.652} & {\bf 35.1}   & 0.676 \\
{\tt CASP}      & 71.8   & {\bf 0.643} & {\bf 13.5}   & 0.699 \\
\hline
\hline
Classification & \multicolumn{2}{c}{SRF} & \multicolumn{2}{c}{GSRF} \\
Dataset & \#dEFS & AUC & \#dEFS & AUC \\
\hline
{\tt CONTRACEPTIVE} & 99.5 & 0.735 & {\bf 67.2} & {\bf 0.739} \\
{\tt REDWINE}       & 95.1 & {\bf 0.821} & {\bf 37.1} & 0.820 \\
{\tt PAGEBLOCKS}    & 77.6 & {\bf 0.984} & {\bf 20.1} & 0.981 \\
{\tt MAGIC}         & 78.8 & {\bf 0.900} & {\bf 25.7} & 0.892 \\
\hline
\end{tabular}
} 
\end{center}
\end{table}

%% file: Sec6.tex
\section{Conclusion} \label{sec:conclusion}
%
%
We proposed a new learning algorithm called SRF for sparse rule models.
The advantage of SRF is that the selected active rules in the model are guaranteed to be optimally chosen from an extremely large number of all possible rules.
The basic idea behind SRF is meta safe screening (mSS) by which multiple features can be safely screened out together.
In a future work we explore further reducing computation time by combining SRF with other non-safe but less conservative methods such as the active set method.

%% file: AppA.tex
\section{Proofs}
\label{app:proof}
\subsection{Proof of Lemma \ref{lem:kkt}} \label{sec:proof_kkt}
\begin{proof}
From the KKT condition (see, e.g., \cite{rockafellar1970convex}) for the primal problem \eqref{eq:primal} and the dual problem \eqref{eq:dual}, the relationship \eqref{eq:kkt-primal2dual} is directly proved. the relationship \eqref{eq:kkt-dual2primal} is proved from the fact that
\begin{align*}
\hbZ_{:j}^{\top}\btheta^*&\in
\begin{cases}
\sign(\hw_j^*) & \text{if}~\hw_j^* \neq 0 \\
[-1,1] & \text{if}~\hw_j^*=0
\end{cases}
\quad\forall j\in[d], \\
\chbZ_{:k}\btheta^*&\in
\begin{cases}
\sign(\chw_k^*) & \text{if}~\chw_k^*\neq0 \\
[-1,1] & \text{if}~\chw_k^*=0
\end{cases}
\quad\forall k\in\cR.
\end{align*}
This suggests that
\begin{align*}
|\hbZ_{:j}^{\top}\btheta^*| & < 1 \Rightarrow \hw_j^* = 0 \quad \forall j \in [d], \\
|\chbZ_{:j}^{\top}\btheta^*| & < 1 \Rightarrow \chw_k^* = 0 \quad \forall k \in \cR.
\end{align*}
\end{proof}
\subsection{Proof of Lemma \ref{lem:val_ub}} \label{sec:proof_ub}
\begin{figure}[t]
\vskip 0.2in
\begin{center}
\centerline{\includegraphics[width=0.9\columnwidth]{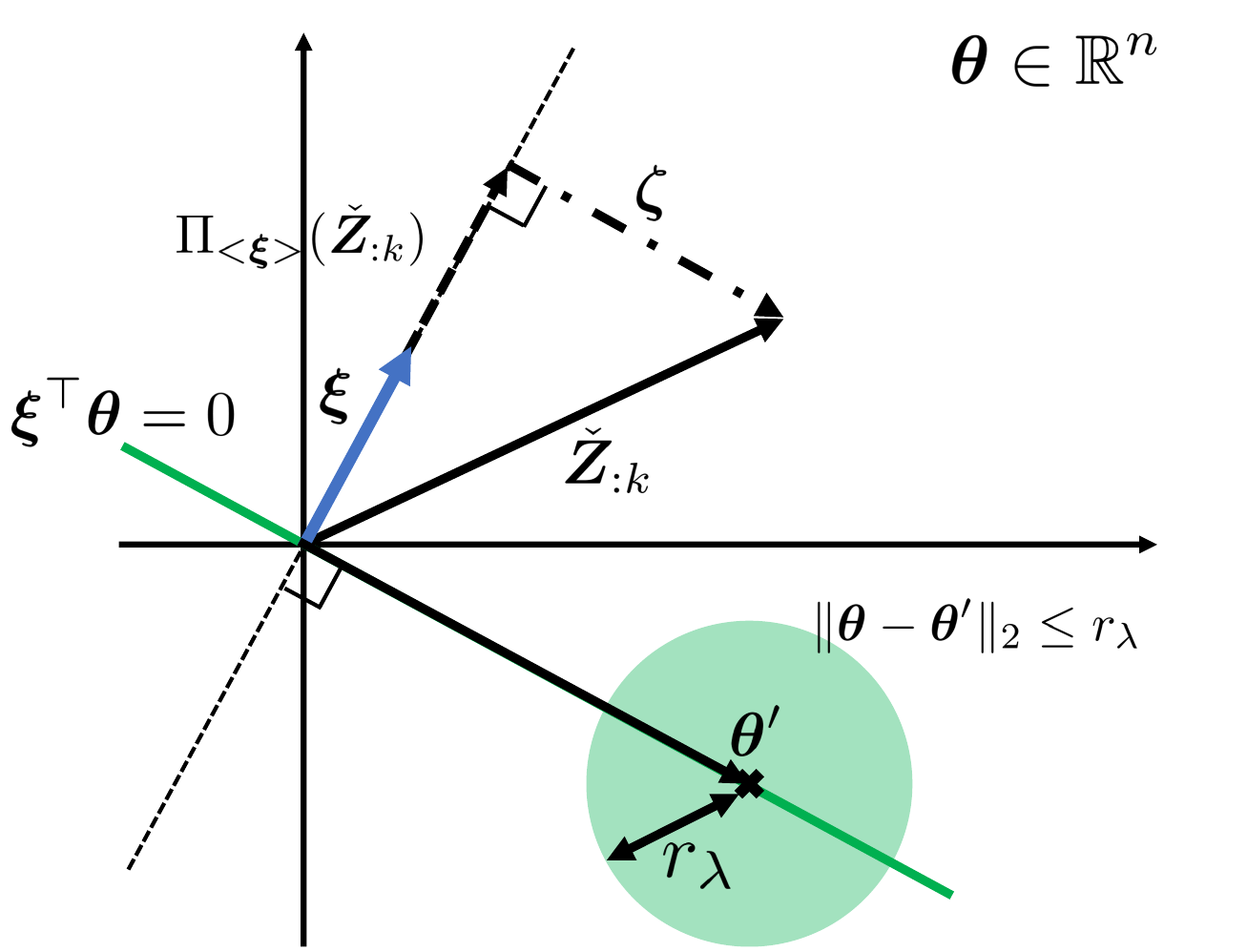}}
\caption{Geometric concept of the proof of Lemma \ref{lem:val_ub}}
\label{fig:geometry_proof}
\end{center}
\vskip -0.2in
\end{figure}

This proof is based on \cite{ndiaye2015gap}, however, it does not consider the case with the bias term $b$ in \eqref{eq:primal}. Fig. \ref{fig:geometry_proof} shows the geometric concept of the proof.
\begin{proof}
We show only the proof of \eqref{eq:val_ub(k)}. (\eqref{eq:val_ub(j)} is similarly proved.)
The optimization problem to solve for \eqref{eq:val_ub(k)} is written as follows:
\begin{align}
\max_{\btheta\in\bbR^n}|\chbZ_{:k}^{\top}\btheta|\quad\text{s.t.}~\|\btheta-\btheta'\|_2\leq r_{\lambda},~\bxi^{\top}\btheta=0.
\label{eq:maximize_score}
\end{align}
For the projected vector $\Pi_{<\bxi>}(\chbZ_{:k}):=\frac{\chbZ_{:k}^{\top}\bxi}{\|\bxi\|_2^2}\bxi\in\bbR^n$ and
$\bm{\zeta}:=\chbZ_{:k}-\Pi_{<\bxi>}(\chbZ_{:k}) \in\bbR^n$,
the optimization problem \eqref{eq:maximize_score} is written as:
\begin{align}
&\max_{\substack{\btheta:\|\btheta-\btheta'\|_2\leq r_{\lambda}, \\ \bxi^{\top}\btheta=0}} |\chbZ_{:k}^{\top}\btheta| \notag \\
&= \max_{\substack{\btheta:\|\btheta-\btheta'\|_2\leq r_{\lambda}, \\ \bxi^{\top}\btheta=0}} |\bm{\zeta}^{\top}\btheta+\Pi_{<\bxi>}(\chbZ_{:k})^{\top}\btheta| \notag \\
&= \max_{\substack{\btheta:\|\btheta-\btheta'\|_2\leq r_{\lambda}, \\ \bxi^{\top}\btheta=0}} |\bm{\zeta}^{\top}\btheta| \label{eq:because_vertical} \\
&\leq \max_{\btheta:\|\btheta-\btheta'\|_2\leq r_{\lambda}} |\bm{\zeta}^{\top}\btheta|, \label{eq:cleary_inequality}
\end{align}
where \eqref{eq:because_vertical} is derived from the constraint $\bxi^{\top}\btheta=0$, and \eqref{eq:cleary_inequality} clearly holds. 
The maximization \eqref{eq:cleary_inequality} can be calculated in the same way as in \cite{ndiaye2015gap} as follows:
\begin{align}
&\max_{\substack{\btheta:\|\btheta-\btheta'\|_2\leq r_{\lambda}, \\ \bxi^{\top}\btheta=0}} |\chbZ_{:k}^{\top}\btheta| \notag \\
&\leq \max_{\btheta:\|\btheta-\btheta'\|_2\leq r_{\lambda}} |\bm{\zeta}^{\top}\btheta| \notag \\
&\leq |\bm{\zeta}^{\top}\btheta'|+r_{\lambda}\|\bm{\zeta}\|_2 \notag \\
&=|(\chbZ_{:k}-\Pi_{<\bxi>}(\chbZ_{:k}))^{\top}\btheta'|+r_{\lambda}\|\chbZ_{:k}-\Pi_{<\bxi>}(\chbZ_{:k})\|_2. \label{eq:pre_equal0}
\end{align}
Due to the precondition of Lemma \ref{lem:sphere_bound}, $\btheta'$ is feasible for the dual problem.
Thus, we have
\begin{align*}
\bxi^{\top}\btheta'=\Pi_{<\bxi>}(\chbZ_{:k})^{\top}\btheta'=0,
\end{align*}
and \eqref{eq:pre_equal0} is calculated as
\begin{align*}
|\chbZ_{:k}^{\top}\btheta'|+r_{\lambda}\|\chbZ_{:k}-\Pi_{<\bxi>}(\chbZ_{:k})\|_2.
\end{align*}
\end{proof}

\subsection{Calculation of Lemma \ref{lem:val_ub} for specific loss functions} \label{app:SS-loss-functions}

Here we present the calculation of the values for the safe screening conditions in Lemma \ref{lem:val_ub}
for two loss functions presented in \S\ref{sec:sparse_learning}.

First suppose the squared loss: $\ell_i(y_i,u_i):=\frac{1}{2}(y_i-u_i)^2$.
As discussed in \S\ref{sec:sparse_learning}, for the squared loss we set
$\bxi = \bm{1}$ and $\ell_i^*(-\lambda\theta_i):=\frac{\lambda^2}{2}\theta_i^2-\lambda y_i\theta_i$.
So we can compute $r_\lambda$ by \eqref{eq:safe_radius} and then Lemma \ref{lem:val_ub} as
\begin{align}
&\hSSUB(j)=|\hbZ_{:j}^{\top}\btheta'|+r_{\lambda}\Bigl\|\hbZ_{:j}-\frac{\hbZ_{:j}^{\top}\bm{1}}{n}\bm{1}\Bigr\|_2, \label{eq:val_ub(j)_squared}\\
&\chSSUB(k)=|\chbZ_{:k}^{\top}\btheta'|+r_{\lambda}\Bigl\|\chbZ_{:k}-\frac{\chbZ_{:k}^{\top}\bm{1}}{n}\bm{1}\Bigr\|_2, \label{eq:val_ub(k)_squared}\\
& \text{where} \nonumber\\
& r_{\lambda}:=\sqrt{2\gamma^{-1}(P_{\lambda}(\hbw',\chbw',b')-D_{\lambda}(\btheta'))}/\lambda, \nonumber\\
& P_{\lambda}(\hbw',\chbw',b') := \frac{1}{2}\sum_{i\in[n]}(y_i-f(\bx_i))^2+\lambda(\|\hbw'\|_1+\|\chbw'\|_1), \nonumber\\
& D_{\lambda}(\btheta') = -\sum_{i\in[n]}\Bigl[ \frac{\lambda^2}{2}{\theta'}_i^2-\lambda y_i{\theta'}_i \Bigr]. \nonumber
\end{align}

Then suppose the logistic loss $\ell_i(y_i,u_i):=\log(1+\exp(-y_iu_i))$.
As discussed in \S\ref{sec:sparse_learning}, for the logistic loss we set
$\bxi = \bm{y}$ and
$\ell_i^*(-\lambda\theta_i):=(1-\lambda\theta_i)\log(1-\lambda\theta_i)+\lambda\theta_i\log(\lambda\theta_i)$ (with $0<\theta_i<\frac{1}{\lambda}$).
So we can compute $r_\lambda$ by \eqref{eq:safe_radius} and then Lemma \ref{lem:val_ub} as
\begin{align}
&\hSSUB(j)=|\hbZ_{:j}^{\top}\btheta'|+r_{\lambda}\Bigl\|\hbZ_{:j}-\frac{\hbZ_{:j}^{\top}\bm{y}}{n}\bm{y}\Bigr\|_2, \label{eq:val_ub(j)_logistic}\\
&\chSSUB(k)=|\chbZ_{:k}^{\top}\btheta'|+r_{\lambda}\Bigl\|\chbZ_{:k}-\frac{\chbZ_{:k}^{\top}\bm{y}}{n}\bm{y}\Bigr\|_2, \label{eq:val_ub(k)_logistic}\\
& \text{where} \nonumber\\
& r_{\lambda}:=\sqrt{2\gamma^{-1}(P_{\lambda}(\hbw',\chbw',b')-D_{\lambda}(\btheta'))}/\lambda, \nonumber\\
& P_{\lambda}(\hbw',\chbw',b') := \nonumber\\
	& \qquad \sum_{i\in[n]}\log(1+\exp(-y_i f(\bx_i)))+\lambda(\|\hbw'\|_1+\|\chbw'\|_1), \nonumber\\
& D_{\lambda}(\btheta') = -\sum_{i\in[n]}\Bigl[ (1-\lambda\theta_i)\log(1-\lambda\theta_i)+\lambda\theta_i\log(\lambda\theta_i) \Bigr]. \nonumber
\end{align}

We add that $\|\bxi\|_2^2 = n$ for both of the loss functions, which is needed to compute $\Pi_{<\bxi>}(\bm{v})$.
In addition, the following properties can be easily confirmed:
\begin{itemize}
\item
	Both $(\hbZ_{:j}^{\top}\bm{1})/n$ in \eqref{eq:val_ub(j)_squared}
	and $(\hbZ_{:j}^{\top}\bm{y})/n$ in \eqref{eq:val_ub(j)_logistic}
	are averages of $X_{:j}$,
	by the definition of $\hbZ$ in \S\ref{sec:sparse_learning}.
\item
	Both $\chbZ_{:k}^{\top}\bm{1}$ in \eqref{eq:val_ub(k)_squared}
	and $\chbZ_{:k}^{\top}\bm{y}$ in \eqref{eq:val_ub(k)_logistic}
	are the supports (\S\ref{sec:pruning-in-enumeration}) of rule $k$ in the dataset $X$,
	by the definition of $\chbZ$ in \S\ref{sec:sparse_learning}.
\end{itemize}

\subsection{Proof of Theorem \ref{thm:pruning}} \label{sec:proof_pruning}
To prove Theorem \ref{thm:pruning}, it is sufficient to show the following lemma.
\begin{lemma}
\label{lem:ub<=Prune}
For any $k'\in\cR_{\text{sub}}(k)$,
\begin{align*}
\chSSUB(k')&=|\chbZ_{:k'}^{\top}\btheta'|+r_{\lambda}\|\chbZ_{:k'}-\Pi_{<\bxi>}(\chbZ_{:k'})\|_2 \\
&\leq \eta_k+r_{\lambda}\|\chbZ_{:k}\|_2=\chmSSUB(k).
\end{align*}
\end{lemma}
To prove Lemma \ref{lem:ub<=Prune}, we use the following lemma. 
\begin{lemma}
\label{lem:rule_bound}
Given an arbitrary $n$-dimensional vector $\bm{c}\in\bbR^n$, for any rule $k\in\cR$, the following relationship holds:
\begin{align*}
&\max\left\{\sum_{i:c_i>0}c_ir_k(\bm{x}_i),-\sum_{i:c_i<0}c_ir_k(\bm{x}_i)\right\} \\
&\geq \left|\sum_{i\in[n]}c_ir_{k'}(\bm{x}_i)\right|,~\forall k'\in\cR_{sub}(k).
\end{align*}
\end{lemma}
%
\begin{proof}[Proof of Lemma \ref{lem:rule_bound}]
First, the following relationship holds:
\begin{align*}
\sum_{i\in[n]}c_{i}r_{k'}(\bm{x}_{i})=\sum_{i:c_{i}>0}c_{i}r_{k'}(\bm{x}_{i})+\sum_{i:c_{i}<0}c_{i}r_{k'}(\bm{x}_{i}).
\end{align*}
Using this equation and \eqref{eq:property-rule}, the following relationship holds:
\begin{align*}
\sum_{i:c_{i}<0}c_{i}r_{k}(\bm{x}_{i})&\leq\sum_{i:c_{i}<0}c_{i}r_{k'}(\bm{x}_{i})\leq\sum_{i\in[n]}c_{i}r_{k'}(\bm{x}_{i}) \\
&\leq\sum_{i:c_{i}>0}c_{i}r_{k'}(\bm{x}_{i})\leq\sum_{i:c_{i}>0}c_{i}r_{k}(\bm{x}_{i}).
\end{align*}
Therefore,
\begin{align*}
&\max\left\{\sum_{i:c_i>0}c_ir_k(\bm{x}_i),-\sum_{i:c_i<0}c_ir_k(\bm{x}_i)\right\}
\geq \left|\sum_{i\in[n]}c_ir_{k'}(\bm{x}_i)\right|.
\end{align*}
\end{proof}
Next, we prove Lemma \ref{lem:ub<=Prune}.
\begin{proof}[Proof of Lemma \ref{lem:ub<=Prune}]
From Lemma \ref{lem:rule_bound},
\begin{align*}
|\chbZ_{:k'}^{\top}\btheta'|\leq\max\left\{\sum_{i:\xi_i\theta_i'>0}\chZ_{ik}\theta_i',-\sum_{i:\xi_i\theta_i'<0}\chZ_{ik}\theta_i'\right\} =:\eta_k.
\end{align*}
Furthermore, by \eqref{eq:property-rule},
\begin{align*}
\|\chbZ_{:k'}-\Pi_{<\bxi>}(\chbZ_{:k'})\|_2\leq\|\chbZ_{:k'}\|_2\leq\|\chbZ_{:k}\|_2.
\end{align*}
From these results,
\begin{align*}
|\chbZ_{:k'}^{\top}\btheta'|+r_{\lambda}\|\chbZ_{:k'}-\Pi_{<\bxi>}(\chbZ_{:k'})\|_2 \leq \eta_k+r_{\lambda}\|\chbZ_{:k}\|_2.
\end{align*}
\end{proof}
\begin{proof}[Proof of Theorem \ref{thm:pruning}]
From Lemmas \ref{lem:sphere_bound}, \ref{lem:val_ub} and \ref{lem:ub<=Prune}, the following relationship holds:
\begin{align}
|\chbZ_{:k'}^{\top}\btheta^*|\leq\chSSUB(k')\leq\chmSSUB(k),~\forall k'\in\cR_{\text{sub}}(k). 
\label{eq:inequality_pruning_proof}
\end{align}
From Lemma \ref{lem:kkt} and eq.\eqref{eq:inequality_pruning_proof}, we obtain Theorem \ref{thm:pruning}.
\end{proof}
\subsection{Proof of Corollary \ref{coro:deeper}} \label{sec:proof_corollary_deeper}
\begin{proof}
For any pair of nodes $k\in\cR$ and $k'\in\cR_{sub}(k)$,
\begin{align}
&\sum_{i:\xi_i\theta_i'>0}\chZ_{ik}\theta_i'=\sum_{i:\xi_i\theta_i'>0}\xi_i\theta_i'r_k(\bm{x}_i) \notag \\
&\quad\quad\geq \sum_{i:\xi_i\theta_i'>0}\xi_i\theta_i'r_{k'}(\bm{x}_i)=\sum_{i:\xi_i\theta_i'>0}\chZ_{ik'}\theta_i', \label{eq:p>p'} \\
&\sum_{i:\xi_i\theta_i'<0}\chZ_{ik}\theta_i'=\sum_{i:\xi_i\theta_i'<0}\xi_i\theta_i'r_k(\bm{x}_i) \notag \\
&\quad\quad\leq \sum_{i:\xi_i\theta_i'<0}\xi_i\theta_i'r_{k'}(\bm{x}_i)=\sum_{i:\xi_i\theta_i'<0}\chZ_{ik'}\theta_i', \label{eq:n<n'}
\end{align}
where both inequalities are proved by \eqref{eq:property-rule}, while all equalities are proved by the definitions of $\chZ_{ik}$ and $\xi_i$. 
Since $\eta_k$ is defined as $\eta_k:=\max\{\sum_{i:\xi_i\theta_i'>0}\chZ_{ik}\theta_i',-\sum_{i:\xi_i\theta_i'<0}\chZ_{ik}\theta_i'\}$, for all four cases we examine which value is larger for $\eta_k$ and $\eta_{k'}$.
\begin{itemize}
\item If $\eta_k=\sum_{i:\xi_i\theta_i'>0}\chZ_{ik}\theta_i'$ and $\eta_{k'}=\sum_{i:\xi_i\theta_i'>0}\chZ_{ik'}\theta_i'$, $\eta_k\geq\eta_{k'}$ holds due to \eqref{eq:p>p'}.
\item If $\eta_k=\sum_{i:\xi_i\theta_i'>0}\chZ_{ik}\theta_i'$ and $\eta_{k'}=-\sum_{i:\xi_i\theta_i'<0}\chZ_{ik'}\theta_i'$, $\eta_k\geq\eta_{k'}$ holds since
\begin{align*}
\eta_k \geq -\sum_{i:\xi_i\theta_i'<0}\chZ_{ik}\theta_i' \geq \eta_{k'},
\end{align*}
where the first inequality is from the definition of $\eta_k$, and the last inequality is from \eqref{eq:n<n'}.
\item If $\eta_k=-\sum_{i:\xi_i\theta_i'<0}\chZ_{ik}\theta_i'$ and $\eta_{k'}=\sum_{i:\xi_i\theta_i'>0}\chZ_{ik'}\theta_i'$, $\eta_k\geq\eta_{k'}$ holds since
\begin{align*}
\eta_k \geq \sum_{i:\xi_i\theta_i'>0}\chZ_{ik}\theta_i' \geq \eta_{k'},
\end{align*}
where the first inequality is from the definition of $\eta_k$, and the last inequality is from \eqref{eq:p>p'} .
\item If $\eta_k=-\sum_{i:\xi_i\theta_i'<0}\chZ_{ik}\theta_i'$ and $\eta_k'=-\sum_{i:\xi_i\theta_i'<0}\chZ_{ik'}\theta_i'$, $\eta_k\geq\eta_{k'}$ holds due to \eqref{eq:n<n'}.
\end{itemize}
Furthermore, by \eqref{eq:property-rule},
\begin{align*}
\|\chbZ_{:k^\prime}\|_2\leq\|\chbZ_{:k}\|_2.
\end{align*}
Since $r_{\lambda}\geq 0$, we obtain
\begin{align*}
\chmSSUB(k)\geq\chmSSUB(k^\prime).
\end{align*}
\end{proof}
\subsection{Proof of \eqref{eq:group_safe} in Theorem \ref{thm:group_screening}} \label{sec:proof_group_screening}

\begin{proof}
To prove this, we modify Theorem 1 in \cite{ndiaye2016gap} for the case of having the bias term $b$ in \eqref{eq:primal_gsrf}. As a result, we have to modify eq.(5) in \cite{ndiaye2016gap} as
\begin{align}
\max_{\substack{\btheta:\|\btheta-\btheta'\|_2\leq r_{\lambda}, \\ \bxi^{\top}\btheta=0}}\|\bm{S}_{\chtau}(\chbZ_g^{\top}\btheta)\|_2.
\label{eq:group_maximize}
\end{align}
The new point here is the equality constraint $\bxi^{\top}\btheta=0$ due to the bias term. Let $\bm{\Xi}_g:=[\bxi,\ldots,\bxi]\in\bbR^{n\times |g|}$, and let $\Pi_{<\bm{\Xi}_g>}(\chbZ_g)\in\bbR^{n\times |g|}$ be
\begin{align*}
\Pi_{<\bm{\Xi}_g>}(\chbZ_g):=\frac{\mathrm{tr}(\bm{\Xi}_g^{\top}\chbZ_g)}{\|\bm{\Xi}_g\|_F^2}\bm{\Xi}_g.
\end{align*}
Moreover, let $\bm{\Gamma}_g:=\chbZ_g-\Pi_{<\bm{\Xi}_g>}(\chbZ_g)\in\bbR^{n\times |g|}$.
Then we have
\begin{align}
&\max_{\substack{\btheta:\|\btheta-\btheta'\|_2\leq r_{\lambda}, \\ \bxi^{\top}\btheta=0}}\|\chbZ_g^{\top}\btheta\|_2 = \max_{\substack{\btheta:\|\btheta-\btheta'\|_2\leq r_{\lambda}, \\ \bm{\Xi}_g^{\top}\btheta=\bm{0}}}\|\chbZ_g^{\top}\btheta\|_2 \label{eq:if_only_if} \\
&=\max_{\substack{\btheta:\|\btheta-\btheta'\|_2\leq r_{\lambda}, \\ \bm{\Xi}_g^{\top}\btheta=\bm{0}}}\|\bm{\Gamma}_g^{\top}\btheta+\Pi_{<\bm{\Xi}>}(\chbZ_g)^{\top}\btheta\|_2 \notag \\
&=\max_{\substack{\btheta:\|\btheta-\btheta'\|_2\leq r_{\lambda}, \\ \bm{\Xi}_g^{\top}\btheta=\bm{0}}}\|\bm{\Gamma}_g^{\top}\btheta\|_2 \label{eq:vertical_group} \\
&\leq \max_{\btheta:\|\btheta-\btheta'\|_2\leq r_{\lambda}}\|\bm{\Gamma}_g^{\top}\btheta\|_2 \label{eq:cleary_inequality_group}
\end{align}
As a result, \eqref{eq:group_maximize} is bounded from above as follows:
\begin{align*}
\max_{\substack{\btheta:\|\btheta-\btheta'\|_2\leq r_{\lambda}, \\ \bxi^{\top}\btheta=0}}\|\bm{S}_{\chtau}(\chbZ_g^{\top}\btheta)\|_2 \leq \max_{\btheta:\|\btheta-\btheta'\|_2\leq r_{\lambda}}\|\bm{S}_{\chtau}(\bm{\Gamma}_g^{\top}\btheta)\|_2.
\end{align*}
Applying eq.(8) in \cite{ndiaye2016gap} to the result above, we have
\begin{align}
&\max_{\substack{\btheta:\|\btheta-\btheta'\|_2\leq r_{\lambda}, \\ \bxi^{\top}\btheta=0}}\|\bm{S}_{\chtau}(\chbZ_g^{\top}\btheta)\|_2 \notag \\
& \leq \max_{\btheta:\|\btheta-\btheta'\|_2\leq r_{\lambda}}\|\bm{S}_{\chtau}(\bm{\Gamma}_g^{\top}\btheta)\|_2 \leq \mathrm{SS_{UB}^G}(g) \label{eq:Tg} \\
&:=
\begin{cases}
\|\bm{S}_{\chtau}(\bm{\Gamma}_g^{\top}\btheta')\|_2+r_{\lambda}\|\bm{\Gamma}_g\|_2 & \text{if}~\|\bm{\Gamma}_g^{\top}\btheta'\|_{\infty}>\chtau \notag \\
(\|\bm{\Gamma}_g^{\top}\btheta'\|_{\infty}+r_{\lambda}\|\bm{\Gamma}_g\|_2-\chtau)_+ & \text{otherwise}\notag 
\end{cases}
.\notag 
\end{align}
Here $\mathrm{SS_{UB}^G}(g)$ is calculated as follows.
Due to the precondition of Lemma \ref{lem:sphere_bound}, $\btheta'$ is feasible for the dual problem.
Thus we have
\begin{align}
\bm{\Xi}_g^{\top}\btheta'=\Pi_{<\bm{\Xi}_g>}(\chbZ_g)^{\top}\btheta'=\bm{0}. \label{eq:equal0_group}
\end{align}
By \cite{ndiaye2016gap} we have
\begin{align}
&\mathrm{SS_{UB}^G}(g)= \\
&\begin{cases}
\|\bm{S}_{\chtau}(\chbZ_g^{\top}\btheta')\|_2+r_{\lambda}\|\bm{\Gamma}_g\|_2 & \text{if}~\|\chbZ_g^{\top}\btheta'\|_{\infty}>\chtau \notag \\
(\|\chbZ_g^{\top}\btheta'\|_{\infty}+r_{\lambda}\|\bm{\Gamma}_g\|_2-\chtau)_+ & \text{otherwise}\notag 
\end{cases}
.
\end{align}
Finally, with Theorem 1 in \cite{ndiaye2016gap}, we have \eqref{eq:group_safe}.
\end{proof}

\subsection{Proof of Theorem \ref{thm:dual_norm_bound}} \label{app:dual_norm_bound}

\begin{proof}
For any vector $\bm{v}\in\bbR^d$, $\epsilon$-norm must satisfy the following inequality (see, for example, \cite{burdakov2001new}: {\S}2, eq.(23)):
\begin{align}
\|\bm{v}\|_{\infty}\leq\|\bm{v}\|_{\epsilon}\leq\frac{\sqrt{d}\|\bm{v}\|_{\infty}}{\sqrt{d}(1-\epsilon)+\epsilon}.
\label{eq:eps_norm_inequality}
\end{align}
Defining $\bm{\phi}\in\bbR^n$ as \eqref{eq:def_phi}, we have
\begin{align*}
\Omega^D(\chbZ^{\top}\bm{\phi})
&= \max_{g\in\cG}\|\chbZ_g^{\top}\bm{\phi}\|_{1-\chtau} \\
&\leq \max_{g\in\cG}\frac{\sqrt{|g|}\|\chbZ_g^{\top}\bm{\phi}\|_{\infty}}{\chtau\sqrt{|g|}+1-\chtau}\quad\because\eqref{eq:eps_norm_inequality} \\
&\leq \max_{g\in\cG}\frac{\sqrt{|g|}\|\chbZ_g^{\top}\bm{\phi}\|_{\infty}}{\chtau\sqrt{|g|}} \quad\because 1-\chtau>0\\
&=\max_{g\in\cG}\frac{\|\chbZ_g^{\top}\bm{\phi}\|_{\infty}}{\chtau}
	=\chtau^{-1}\|\chbZ^{\top}\bm{\phi}\|_{\infty}.
\end{align*}
\end{proof}

%% file: AppB.tex
\section{Experimental Setups and Detailed Results}

\subsection{Preprocesses} \label{sec:preprocesses}

For each categorical input feature of $c$ choices, we composed $c-1$ dummy variables to encode into an ordinary matrix.

For each input feature in each dataset, the values are scaled by linear transformation so that the average and the variance of the values are 0 and 1, respectively, except for RuleFit\footnote{In RuleFit, since a recommended normalization is presented in the paper (outlier removals before normalizing the variance), we used it. See {\S}5 of \cite{friedman2008predictive}.}. For each regression dataset, the same normalization is applied for the label values $\bm{y}$.

Since the following datasets in Table \ref{tab:datasets124} do not have binary labels, we constructed binary labels as follows:
\begin{itemize}
\item Dataset {\tt CONTRACEPTIVE}: This is a dataset for 3-class classification. Since the class labels are ``No use'', ``Long-term use'' and ``Short-term use'' (of contraceptive method), we constructed binary labels as ``No use'' and others.
\item Dataset {\tt REDWINE}: This is a dataset with ordinal labels (0 to 10, quality of red wine). Since 82\% (1,319 out of 1,599) of instances have the label either 5 or 6, we constructed binary labels ``5 or less'' and ``6 or more''.
\item Dataset {\tt PAGEBLOCKS}: This is a dataset for 5-class classification. Since 90\% (4,913 out of 5,473) of instances have the label ``text'', we constructed binary labels ``text'' and ``not text''
\end{itemize}

\subsection{Discretization of data values} \label{sec:discretization}

Here we describe two discretization methods for the segment of rules $\bm{\omega}^{(j)}$ in \S\ref{sec:segment}.

For experiments except for Exp.3, we used discretization by quantiles (Appendix \ref{sec:discretize-quantile}) to limit the number of $\bm{\omega}^{(j)}$. On the other hand, for Exp.3, we used the same discretization method as per REPR (Appendix \ref{sec:discretize-interval}) since we compared the proposed method with REPR in this experiment.
%

\subsubsection{Discretization by quantiles} \label{sec:discretize-quantile}
In this method, we specify a parameter $M \geq 2$, which represents the number of discretizations.
Roughly speaking, for each input feature $j$, we set $\bm{\omega}^{(j)}$ as the
$(q/M)$-quantiles ($q\in[M-1]$) of $\bm{X}_{:j}$.
Since we assume that $\bm{\omega}^{(j)}$ must be a midpoint of two neighboring values
in $\bm{X}_{:j}$, we need some additional operations. See Algorithm \ref{alg:discretization-quantile}
for details.

If there are $M$ or less distinct values in $\bm{X}_{:j}$, we do not conduct the computation above.

\begin{algorithm}[tp]
\caption{Discretization by quantiles}
\label{alg:discretization-quantile}
\begin{algorithmic}
\REQUIRE $M\in\bbN$~($M\geq 2$), $\bm{X}_{:j}\in\bbR^n$~($j\in[d]$)
\STATE $\bm{\omega}^{tmp} \gets \emptyset$
\FOR{$q\in[M-1]$}
	\STATE $i_L \gets \lfloor q/n \rfloor$, $i_U \gets \lceil q/n \rceil$
	\IF{$x_{i_L j} \neq x_{i_U j}$}
		\STATE $\bm{\omega}^{tmp} \gets \bm{\omega}^{tmp} \cup (x_{i_L j} + x_{i_U j})/2$
	\ELSE
		\STATE \COMMENT{Since $\bm{\omega}^{tmp}$ cannot include values in $\bm{X}_{:j}$,}
		\STATE \COMMENT{we take either neighbor of $x_{i_L j}$ ($= x_{i_U j}$):}
		\STATE \COMMENT{smaller neighbor $v_L$ or larger neighbor $v_U$.}
		\STATE $v_L \gets \max\{ x_{ij} : i\in[n], x_{ij} < x_{i_L j} \}$
		\STATE $v_U \gets \min\{ x_{ij} : i\in[n], x_{ij} > x_{i_L j} \}$
		\IF{$v_L = \mathrm{null}$ {\bf or} $v_U - x_{i_L j} > x_{i_L j} - v_L$}
			\STATE $\bm{\omega}^{tmp} \gets \bm{\omega}^{tmp} \cup (v_U + x_{i_L j})/2$
		\ELSE
			\STATE $\bm{\omega}^{tmp} \gets \bm{\omega}^{tmp} \cup (x_{i_L j} + v_L)/2$
		\ENDIF
	\ENDIF
\ENDFOR
\STATE {\bf return} $\bm{\omega}^{(j)} \gets \bm{\omega}^{tmp} \cup \{-\infty, +\infty\}$
\end{algorithmic}
\end{algorithm}

\subsubsection{Discretization by intervals} \label{sec:discretize-interval}

The existing method REPR chose $\bm{\omega}^{(j)}$ with the following procedure.
Given a parameter of interval size $\delta\geq 0$, for each input feature $j$,
we add a midpoint of two neighboring data values to $\bm{\omega}^{(j)}$
if and only if they are apart by $\delta(\max_{i\in[n]} x_{ij} - \min_{i\in[n]} x_{ij})$.
The detailed procedure is described in Algorithm \ref{alg:discretization-interval}.

\begin{algorithm}[tp]
\caption{Discretization by intervals}
\label{alg:discretization-interval}
\begin{algorithmic}
\REQUIRE $\delta \geq 0$, $\bm{X}_{:j}\in\bbR^n$~($j\in[d]$)
\STATE Sort $\bm{X}_{:j}$ and store the list of indices $o$, that is,
	$x_{o_1 j} \leq x_{o_2 j} \leq \dots \leq x_{o_n j}$
	and $o$ is a permutation of $[n]$.
\STATE $\bm{\omega}^{tmp} \gets \emptyset$
\FOR{$i\in[n-1]$}
	\IF{$x_{o_{i+1} j} - x_{o_i j} > \delta(x_{o_n j} - x_{o_1 j})$}
		\STATE $\bm{\omega}^{tmp} \gets \bm{\omega}^{tmp} \cup (x_{o_i j} + x_{o_{i+1} j})/2$
	\ENDIF
\ENDFOR
\STATE {\bf return} $\bm{\omega}^{(j)} \gets \bm{\omega}^{tmp} \cup \{-\infty, +\infty\}$
\end{algorithmic}
\end{algorithm}

\subsection{Implementations of existing methods} \label{chap:implementation-details}

We used the following implementations of existing methods as follows.

In Exp.1, we used the Python library {\em scikit-learn} \cite{scikit-learn} for all baseline methods
({\em sklearn.linear\_model.Lasso} and {\em sklearn.kernel\_ridge.KernelRidge} for regressions, and
{\em sklearn.linear\_model.LogisticRegression} and {\em from sklearn.svm.SVC} for classifications).
We used the RBF kernel for both {\em sklearn.kernel\_ridge.KernelRidge} and {\em from sklearn.svm.SVC}.

For Random Forest in Exp.2, we used {\em scikit-learn} ({\em sklearn.ensemble.RandomForestRegressor} for regressions and {\em sklearn.ensemble.RandomForestClassifier} for classifications). As noted in the subsection of Exp.2, the number of rules for each decision tree (10) and the number of decision trees (10) are specified, and all other hyperparameters are default ones.

For RuleFit in Exp.2, we used the implementation by Christoph Molnar (https://github.com/christophM/rulefit).

\subsection{Hyperparameter setups} \label{chap:hyperparameters}

The hyperparameters in the models are set up as follows.
See \S\ref{sec:regularization_path} and \S\ref{sec:lammax_gsrf} (regularization paths) for the basic concept.

\subsubsection{Exps.1, 4 and 5}
In Exps.1 and 5, we choose the hyperparameters for the best prediction performance with respect to cross-validation performance.
For SRF, LASSO and $L_1$-penalized logistic regression, we set $\lambda$ from $\lammax$ to $0.01\lammax$, divided into 100 $\lambda$'s in logarithmic scale. In the cross-validation we selected the best $\lambda/\lammax$ among 100 candidates, that is, we did this for both the cross-validation and the final model fitting with respective $\lammax$'s. (Note that $\lammax$ differs depending on data values.)
For kernel ridge regression and kernel SVM, since each has two hyperparameters $\lambda$ (strength of $L_2$ penalty) and $\gamma$ (hyperparameter for RBF kernel), we find the best one with a grid search for $\lambda, \gamma\in\{2^{10}, 2^9, \dots, 2^{-10}\}$.

In Exp.4, since we would like to examine just the computational costs, we set $\lambda$ as above without cross-validations, and just counted the number of rules for the 100 $\lambda$'s.

\subsubsection{Exps.2, 3 and 6}
In these experiments, we compared the models when each of the methods retrieved 100 rules.

In Exp.2, for SRF we set $\lambda$ from $\lammax$ to $10^{-3}\lammax$, divided into 10,000 $\lambda$'s in logarithmic scale.
For RuleFit, since it employs $L_1$-penalized empirical risk minimization after candidates of rules are separately determined, we set $\lambda$ from $\lammax$ to $10^{-4}\lammax$, divided into 10,000 $\lambda$'s in logarithmic scale.

In Exp.3, for SRF we set $\lambda$ from $\lammax$ to $0.5$, divided into 1,000 $\lambda$'s in logarithmic scale.
Since $\lambda = 0.5$ in REPR implies that it retrieves at least 100 rules, we set it as the smallest $\lambda$ in the regularization path.

In Exp.6, for SRF we set $\lambda$ from $\lammax$ to $10^{-3}\lammax$, divided into 1,000 $\lambda$'s in logarithmic scale. For GSRF we used $\tilde{\lambda}_{\max}$ instead (see Appendix \ref{sec:lammax_gsrf}).

\subsection{Supplement for Exp.2: Similarity between rules} \label{app:rule-similarity}

\begin{figure}[t]
\begin{center}
\includegraphics[width=0.5\hsize]{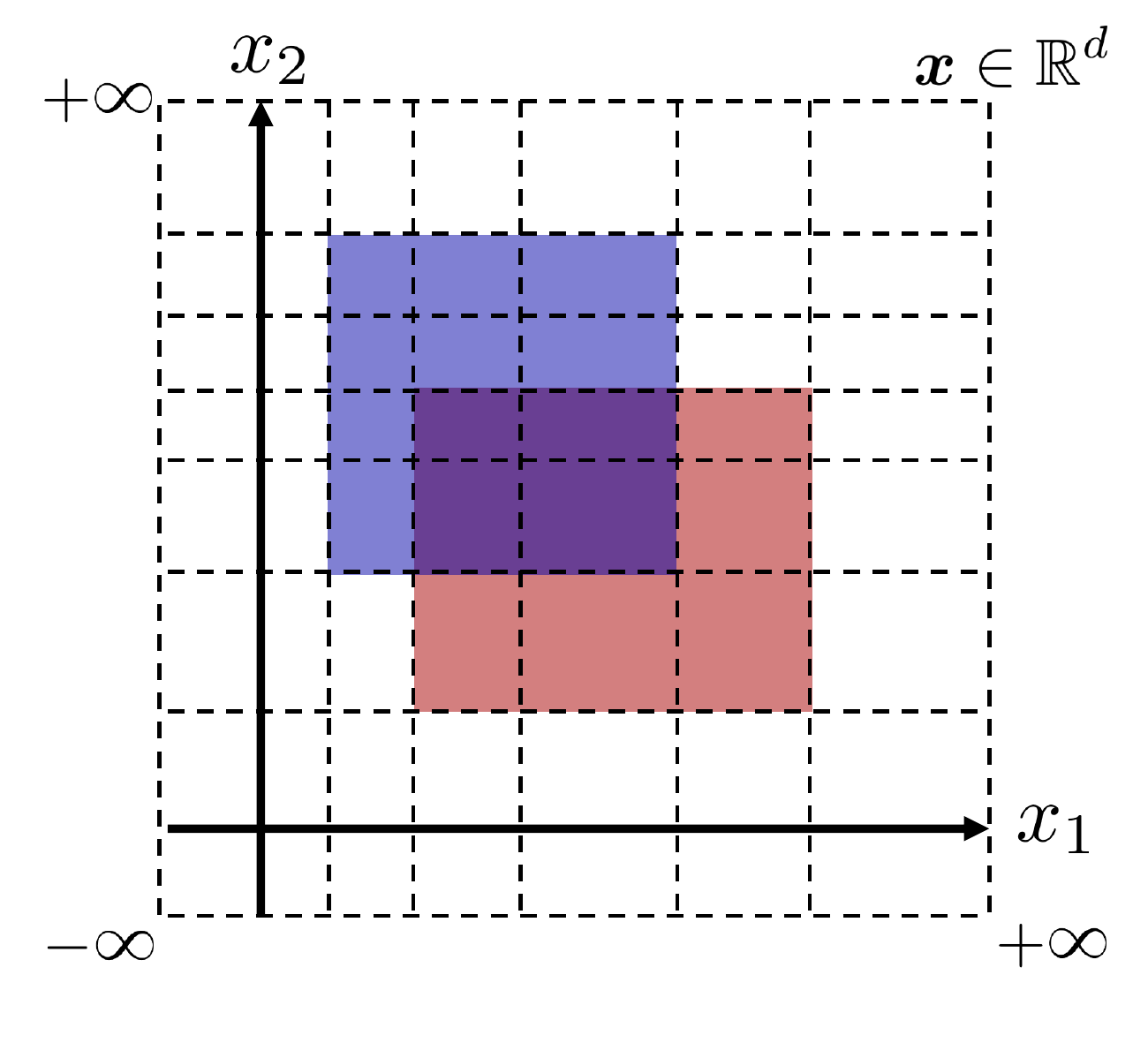}
\end{center}
\vspace{-1.5em}
\caption{Example of the similarity of two rules. In this example, rules of blue and red rectangles include 12 and 9 boxes, respectively. Moreover, 17 boxes are included by either of the rules, and 4 of them are included by both of the rules. Thus, the similarity is $4/17\approx 0.24$ by the Jaccard index.}
\label{fig:rule_similarity}
\end{figure}

In this section we present the formal definition of similarity between rules used in Exp.2.
Let a data matrix $X\in\bbR^{n\times d}$ and the candidates of hyper-rectangle edges $\{\bm{\omega}^{(j)}\}_{j\in[d]}$ be fixed.
Then, intuitively, the similarity between two rules $k$, $k^\prime$ is defined by the Jaccard index for the sets of ``boxes'' defined by two rule hyper-rectangles $[\bm{\ell}^{(k)}, \bm{u}^{(k)}]$ and $[\bm{\ell}^{(k^\prime)}, \bm{u}^{(k^\prime)}]$. See Fig. \ref{fig:rule_similarity} for example.

We first define the set of ``boxes'' for a rule.
A hyper-rectangle $[\bm{\ell}, \bm{u}]$
(i.e., $\ell_j < u_j$ and $\ell_j, u_j\in\bm{\omega}^{(j)}$ for all $j\in[d]$)
is called a {\em box} if, for any $j\in[d]$,
there exists no $v_j\in\bm{\omega}^{(j)}$ such that $\ell_j < v_j < u_j$.
Then we define $V_k$ for a rule $k$ as the set of boxes included in $[\bm{\ell}^{(k)}, \bm{u}^{(k)}]$. Finally, for two rules $k$ and $k^\prime$, their similarity is defined as \eqref{eq:rule-simiarlity} in Exp.2.

\subsection{Supplement for Exp.3: Running SRF under the same settings as REPR as much as possible} \label{app:setups-REPR}
In this section we show the details of setups in Exp.3: running SRF under the same settings as REPR as much as possible.
\begin{description}
\item[Discretization]~\\We used the same discretization as that in REPR. See Appendix \ref{sec:discretize-interval}.
\item[Number of training instances]~\\We took one-fifth of instances for each dataset as the training set at random. In the paper of REPR, a dataset is divided into training and testing sets by 1:4. Since in this experiment we did not examine prediction performance, we took one-fifth of instances as the training set.
\item[Hyperparameter settings]~\\In the paper of REPR, $\lambda = 0.5$ (in the formulation in this paper) was used and retrieved 100 rules. This indicates that the number of active rules with $\lambda = 0.5$ is at least 100. Thus, for SRF, we set $\lambda$ from $\lammax$ to $0.5$ for finding $\lambda$ retrieving nearly 100 active rules (details in Appendix \ref{chap:hyperparameters}).
\end{description}

\subsection{Supplement for Exp.4}

\subsubsection{Setup of experiment (B): choosing the subset of input features at random} \label{app:random-subset-of-features}

As stated in Exp.4 in \S\ref{sec:experiment}, the experiment (B) examines the increase of the number of generated rules by SRF against the increase of the input features. To do this, we consider examining the number by randomly picking up a part of input features.
Since the results are expected to largely depend on the chosen input features, we examined pick-ups of input features at random in the following manner (example in Figure \ref{fig:shuffling}):
\begin{itemize}
\item Let $d^\prime$ be the number of all input features in the dataset, and $d$ be the number of input features to be picked up.
\item First, if the dataset has $v$ binary input features (input features whose values are only 0 or 1), then we pick up all of them (i.e., we do not consider the additions or removals of binary features). This is because, even if we set $M=3$, such features are discretized into only two values rather than three (Appendix \ref{sec:discretize-quantile}).
\item Then, we prepared $(d-v)(d^\prime-v)$ cards: $(d-v)$ cards for each of $(d^\prime-v)$ input features. Then we shuffled them into $(d^\prime-v)$ groups of $(d-v)$ cards each, such that no pair of groups are equivalent and no group contains two or more same input feature. Such shuffling can be done by ``latin squares'' algorithm. We consider these $(d^\prime-v)$ groups as the random choices of $d$ input features (input features of $(d-v)$ cards plus $v$ binary input features). We employed such a way in order to use input features for the same numbers of chances.
\end{itemize}
With these setups, for each dataset in Table \ref{tab:datasets124} we created the following numbers of random choices:
\begin{itemize}
\item First, we did not examine for {\tt CONTRACEPTIVE} dataset since $v=15$ out of $d^\prime=17$ features are binary.
\item For the dataset {\tt ABALONE} ($d^\prime = 10$), since there are $v=3$ binary features, for each $d\in\{4, 5, \dots, 9\}$ we created $d^\prime - v = 7$ random choices. For $d=3$ we examined only one case: picking up only $v=3$ binary features.
\item For each of other six datasets, since there are no binary features, for each $d\in\{3, 4, \dots, d^\prime-1\}$ we created $d^\prime$ random choices.
\end{itemize}

\begin{figure}[t]
\begin{screen}
To take $d=3$ input features from a dataset having $d^\prime=5$ input features (no binary features):
\begin{enumerate}
\item Prepare $d=3$ cards for each of input features $[d^\prime]$ (total $dd^\prime=15$): {\tt 1 1 1 2 2 2 3 3 3 4 4 4 5 5 5}.
\item Shuffle it to $d^\prime=5$ groups of $d=3$ cards so that no pair of groups are equivalent and no group contains two or more same input feature: [{\tt 1 3 4}], [{\tt 2 4 5}], [{\tt 1 3 5}], [{\tt 1 2 5}], [{\tt 2 3 4}].
\end{enumerate}
For each of $d^\prime=5$ groups, we create a new dataset having only the specified $d=3$ input features.
\end{screen}
\caption{Overview of choosing the subset of input features at random for Exp.4}
\label{fig:shuffling}
\end{figure}

\subsubsection{Results of the experiments for all datasets} \label{app:exp4-results-all}

\begin{figure*}[!p]
\begin{center}
\noindent
\includegraphics[width=0.75\hsize]{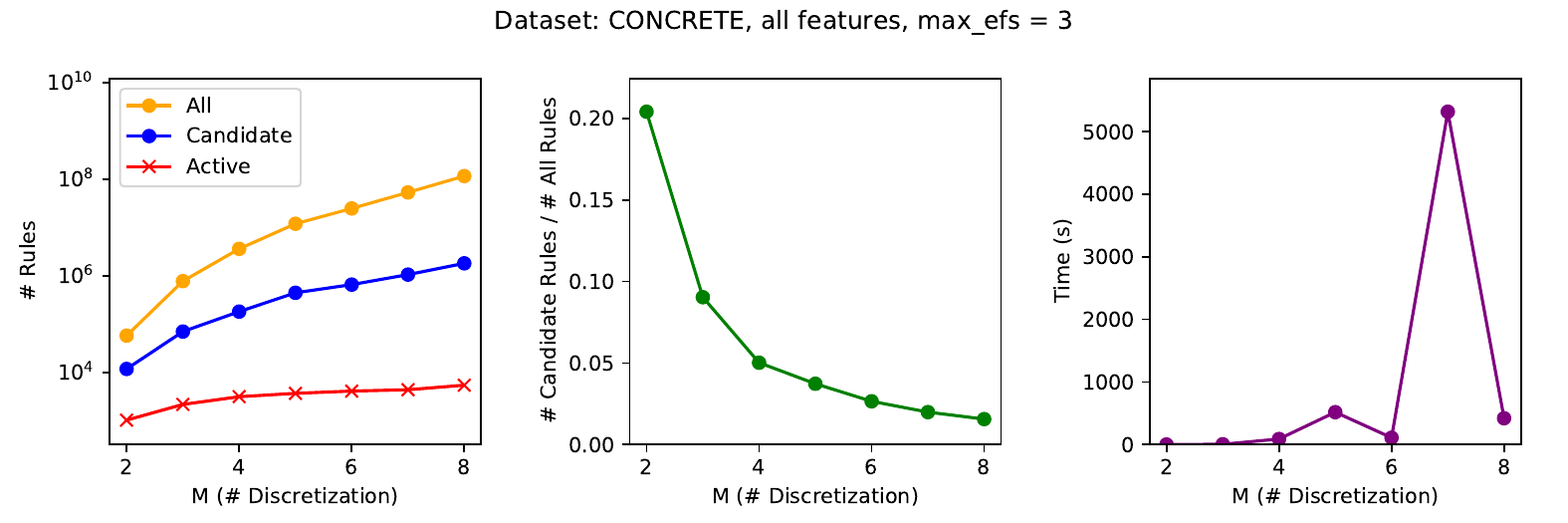}\\
\includegraphics[width=0.75\hsize]{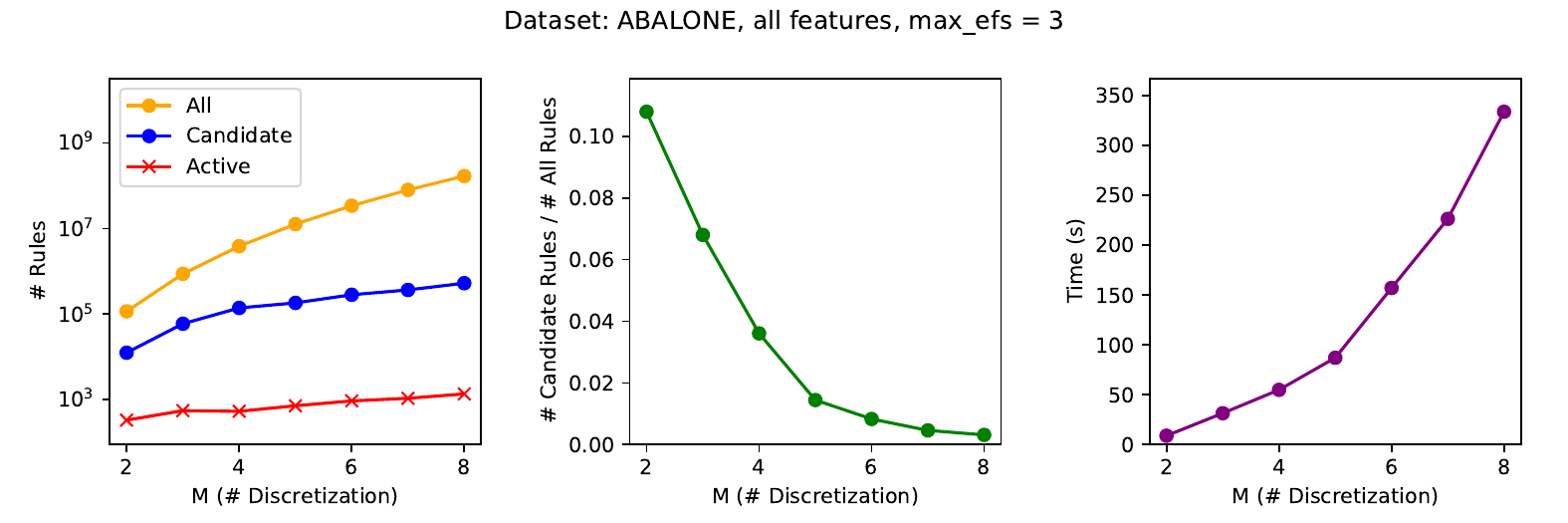}\\
\includegraphics[width=0.75\hsize]{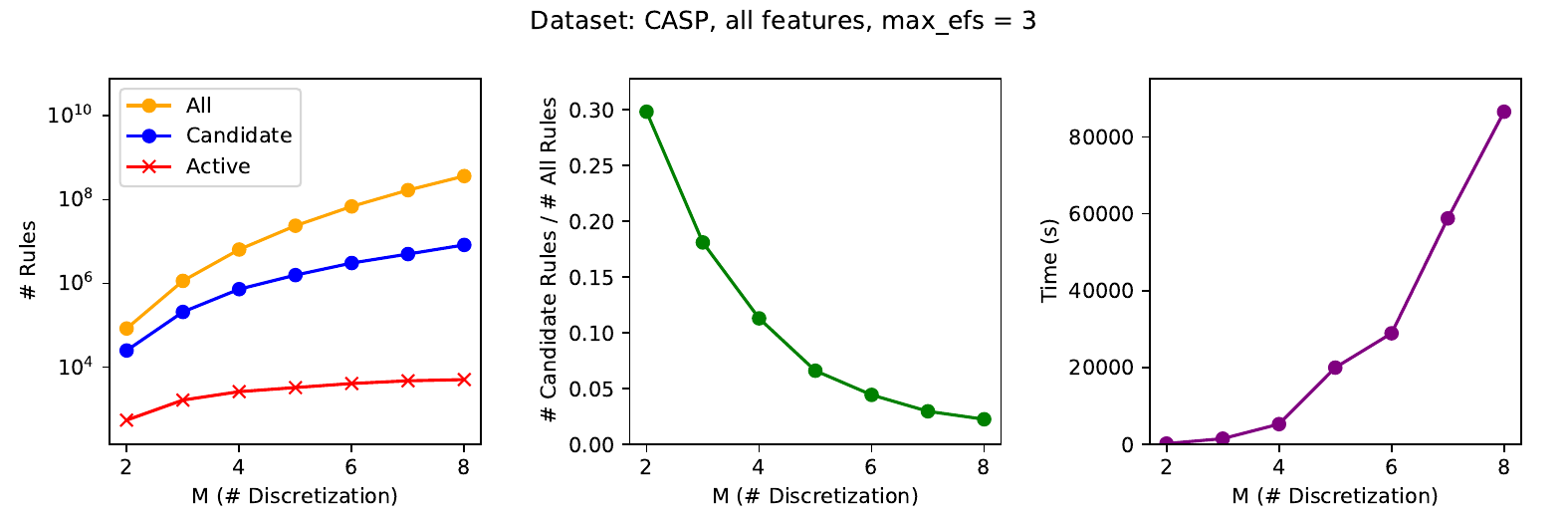}\\ 
\includegraphics[width=0.75\hsize]{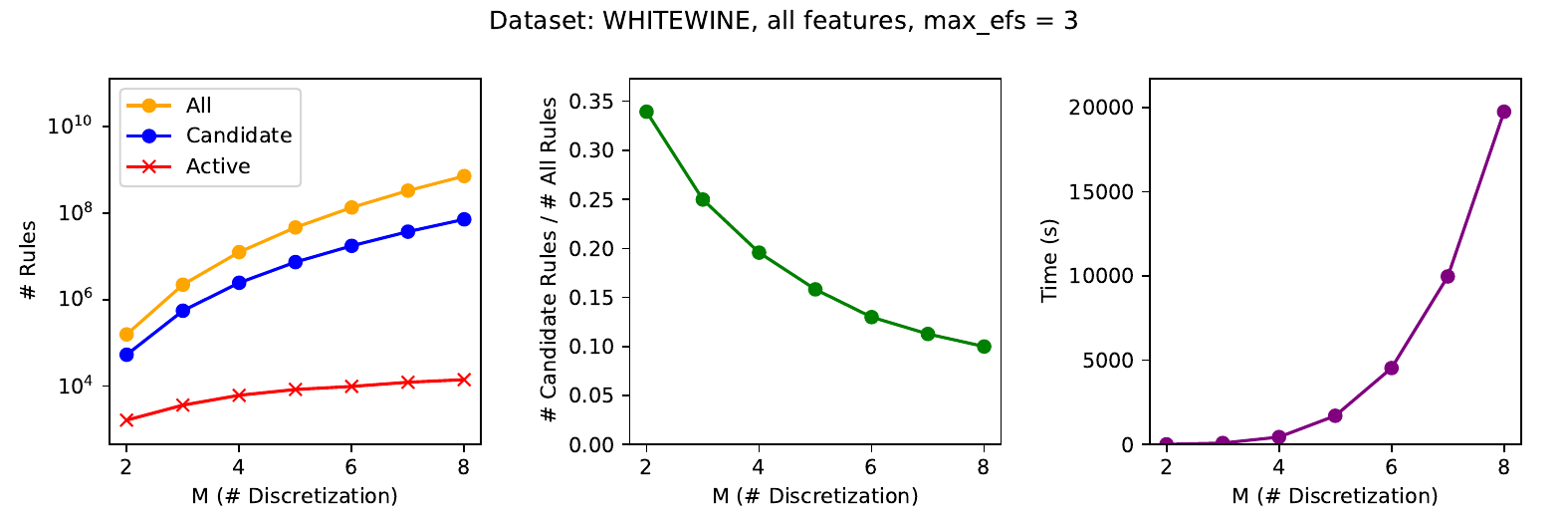}
\end{center}
\caption{Result of Exp.4: Number of enumerated rules by SRF compared to all possible rules, for (A) the changes of the number of discretizations $M$ (regression datasets). The left is the number of rules (all possible rules, enumerated rules by SRF and active rules at last), the center is the the right is the ratios of the number of rules enumerated by SRF to all possible rules, and the right is the computation time.}
\label{fig:exp4-discre-reg}
\end{figure*}

\begin{figure*}[!p]
\begin{center}
\noindent
\includegraphics[width=0.75\hsize]{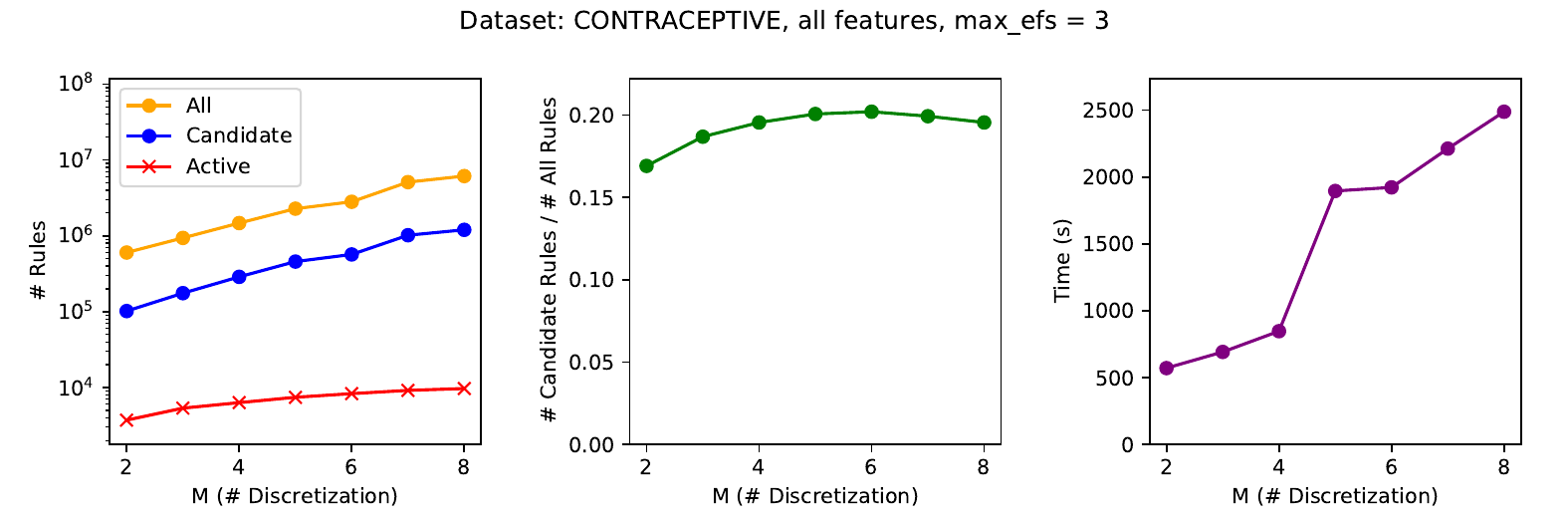}\\
\includegraphics[width=0.75\hsize]{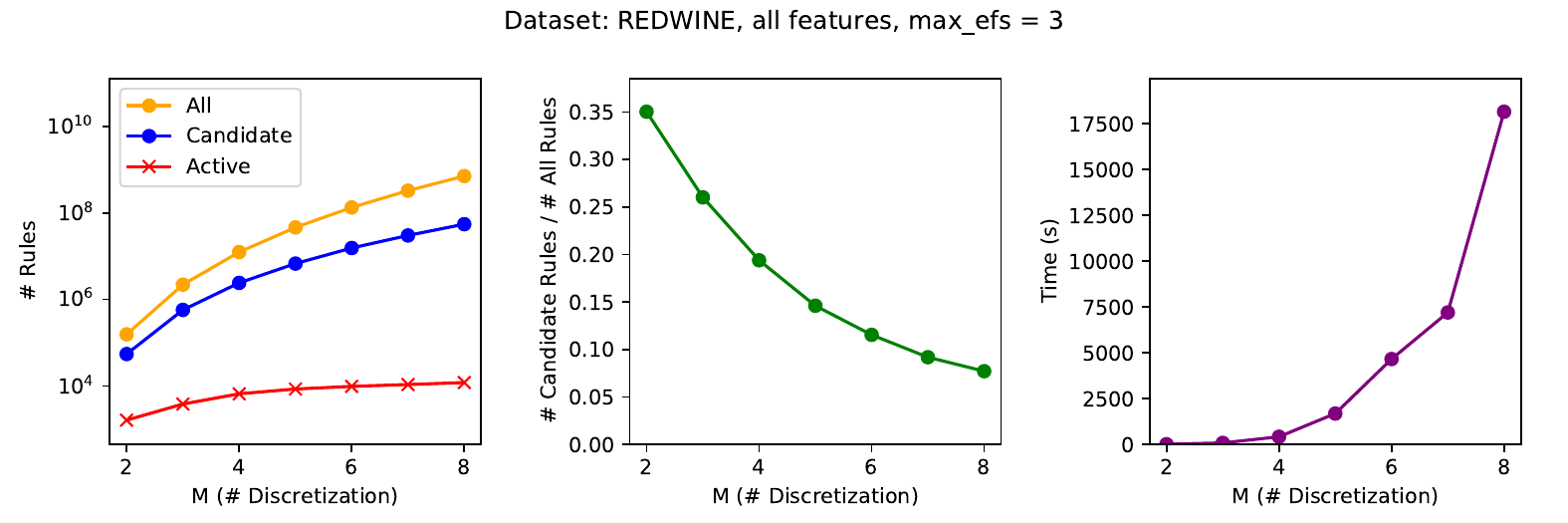}\\
\includegraphics[width=0.75\hsize]{exp-nodes-time/Discre-pageblocks.pdf}\\
\includegraphics[width=0.75\hsize]{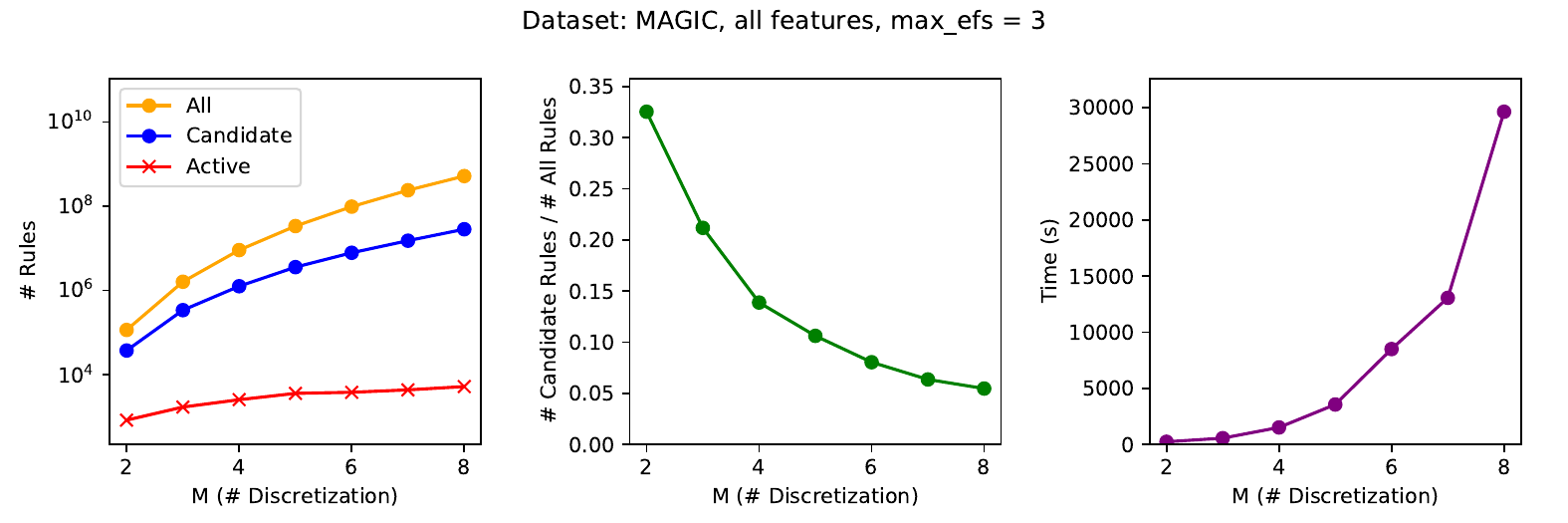}
\end{center}
\caption{Result of Exp.4: Number of enumerated rules by SRF compared to all possible rules, for (A) the changes of the number of discretizations $M$ (classification datasets). The left is the number of rules (all possible rules, enumerated rules by SRF and active rules at last), the center is the the right is the ratios of the number of rules enumerated by SRF to all possible rules, and the right is the computation time.}
\label{fig:exp4-discre-cls}
\end{figure*}

\begin{figure*}[!p]
\begin{center}
\noindent
\includegraphics[width=0.75\hsize]{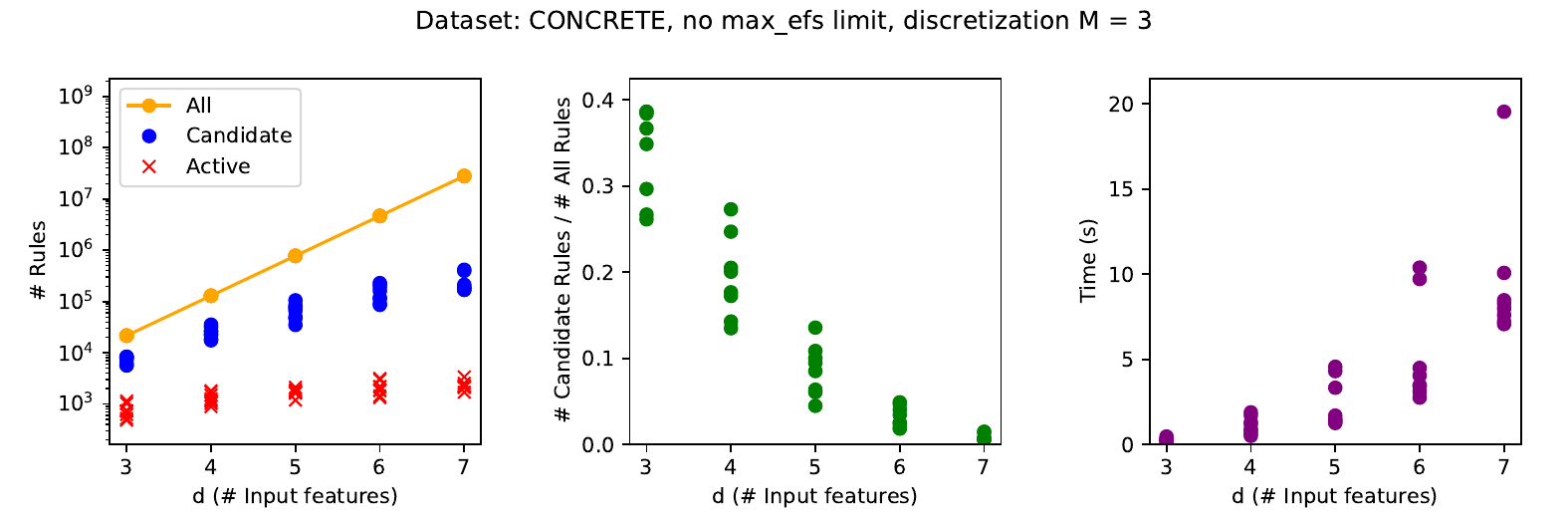} \\
\includegraphics[width=0.75\hsize]{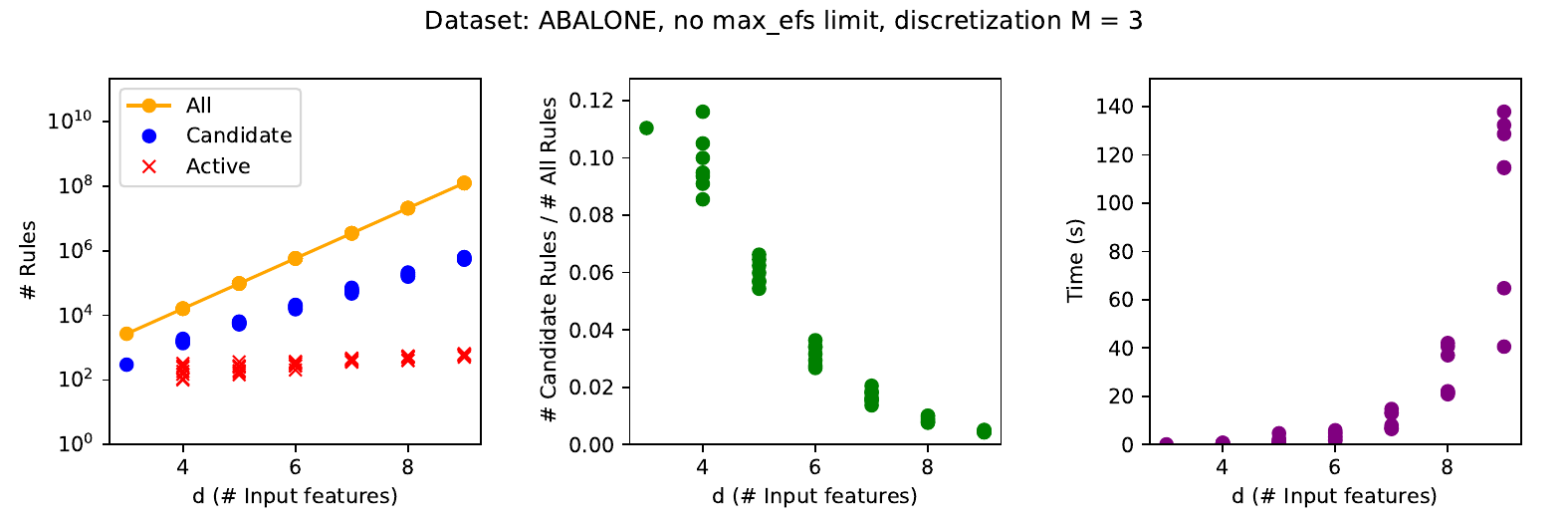}  \\
\includegraphics[width=0.75\hsize]{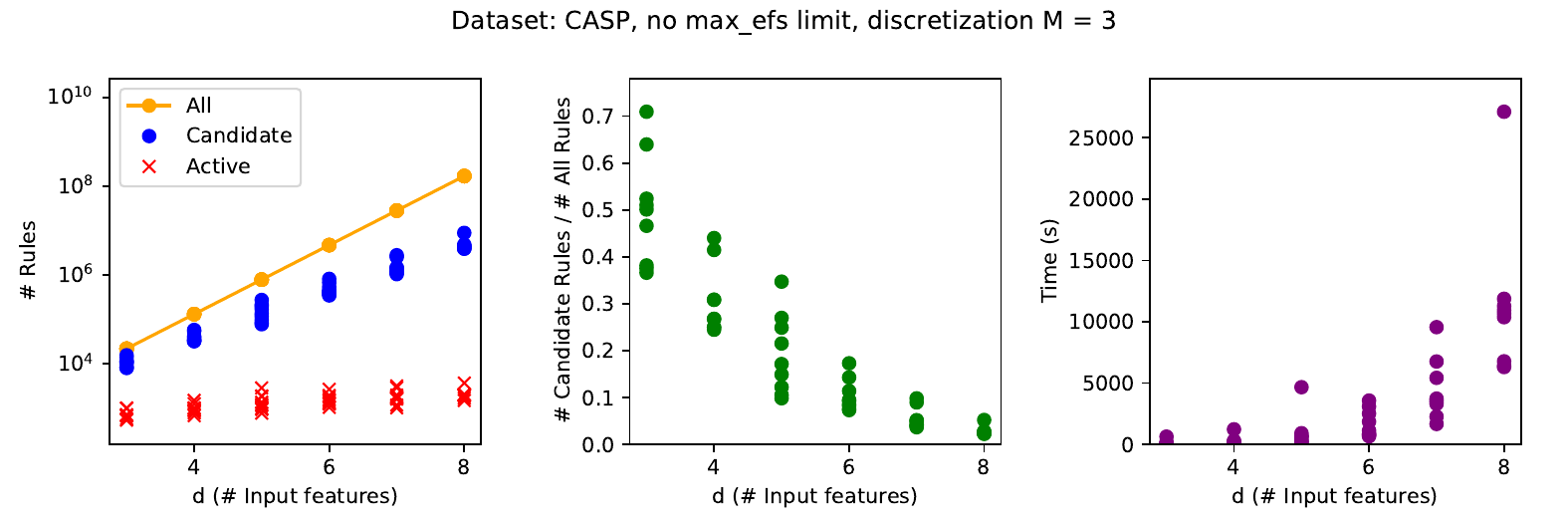}     \\
\includegraphics[width=0.75\hsize]{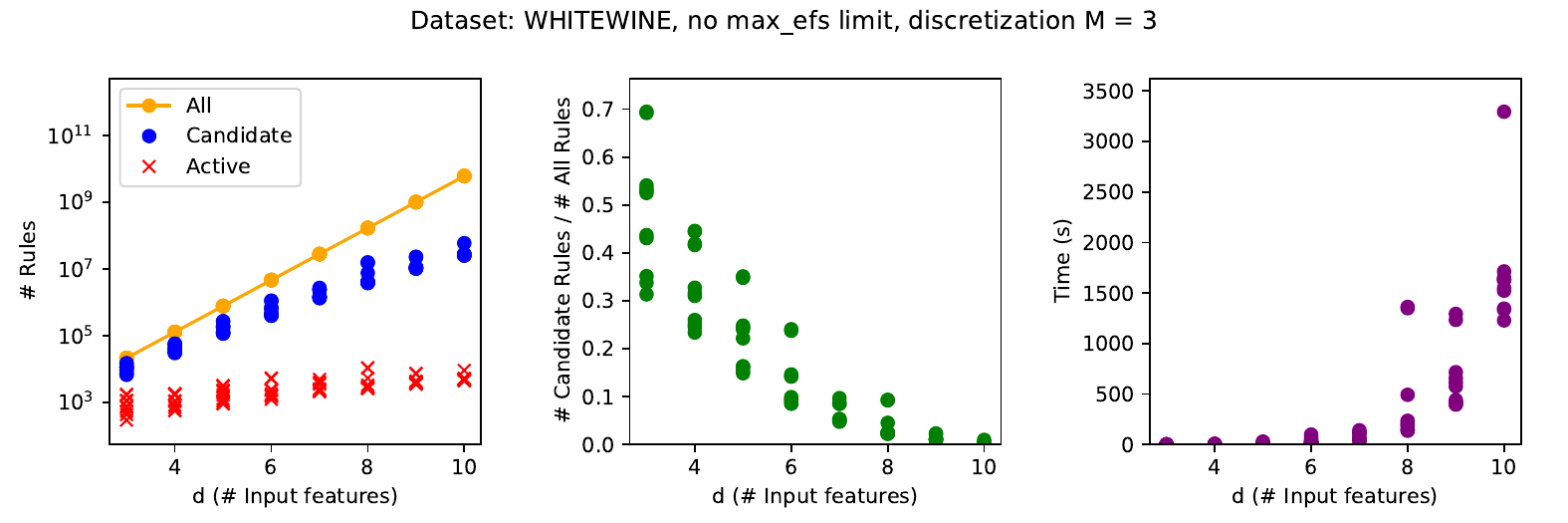}
\end{center}
\caption{Result of Exp.4: Number of enumerated rules by SRF compared to all possible rules, for (B) the changes of the number of input features $d$ (regression datasets). The left is the number of rules (all possible rules, enumerated rules by SRF and active rules at last), the center is the the right is the ratios of the number of rules enumerated by SRF to all possible rules, and the right is the computation time.}
\label{fig:exp4-OrigInputFeat-reg}
\end{figure*}

\begin{figure*}[!p]
\begin{center}
\noindent
\includegraphics[width=0.75\hsize]{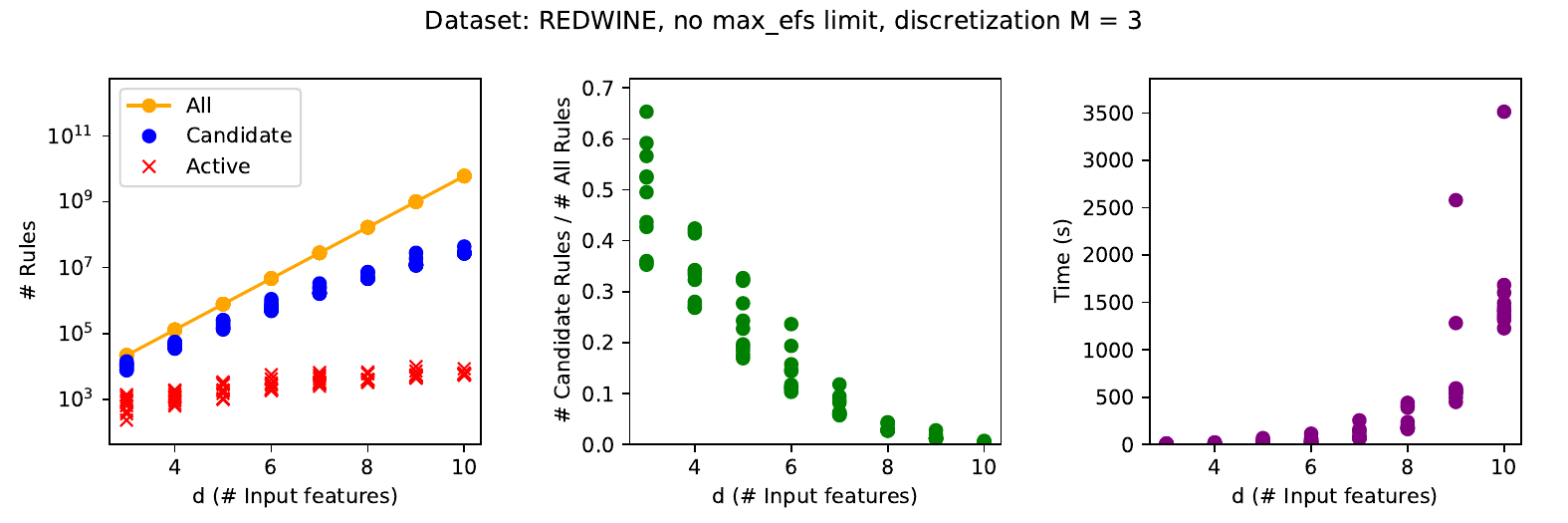}\\
\includegraphics[width=0.75\hsize]{exp-nodes-time/OrigInputFeat-pageblocks.pdf}\\
\includegraphics[width=0.75\hsize]{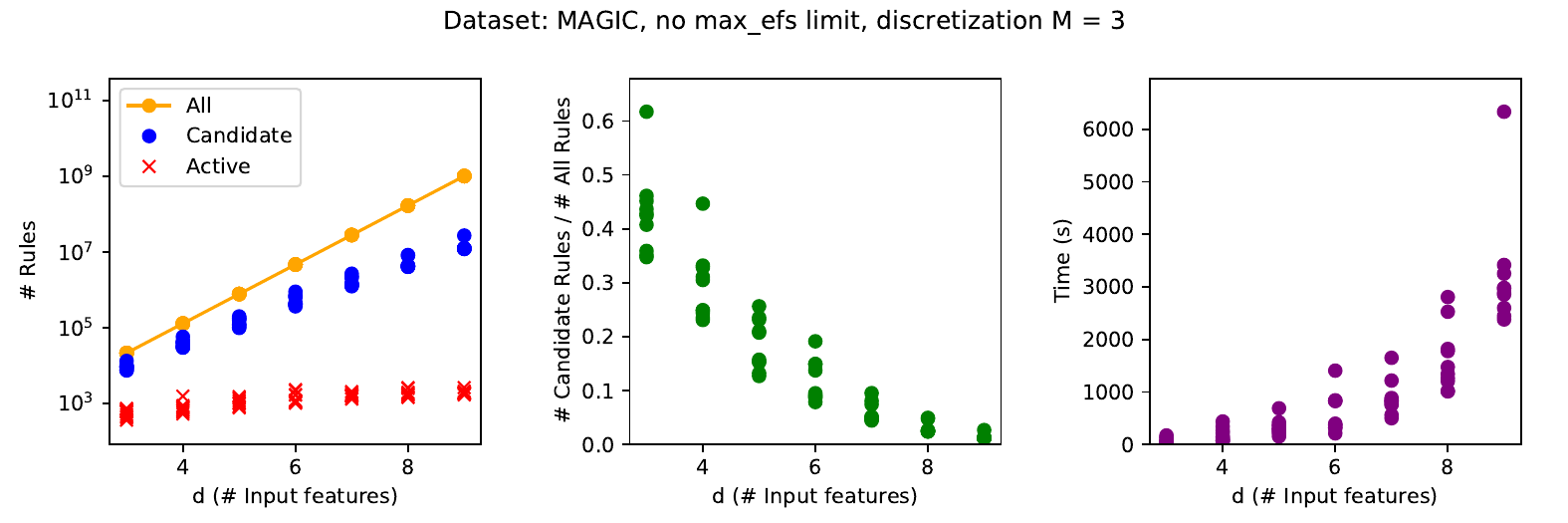}\\
{[No experiment for dataset {\tt CONTRACEPTIVE}; See Appendix \ref{app:random-subset-of-features}]}
\end{center}
\caption{Result of Exp.4: Number of enumerated rules by SRF compared to all possible rules, for (B) the changes of the number of input features $d$ (classification datasets). The left is the number of rules (all possible rules, enumerated rules by SRF and active rules at last), the center is the the right is the ratios of the number of rules enumerated by SRF to all possible rules, and the right is the computation time.}
\label{fig:exp4-OrigInputFeat-cls}
\end{figure*}

\begin{figure*}[!p]
\begin{center}
\noindent
\includegraphics[width=0.75\hsize]{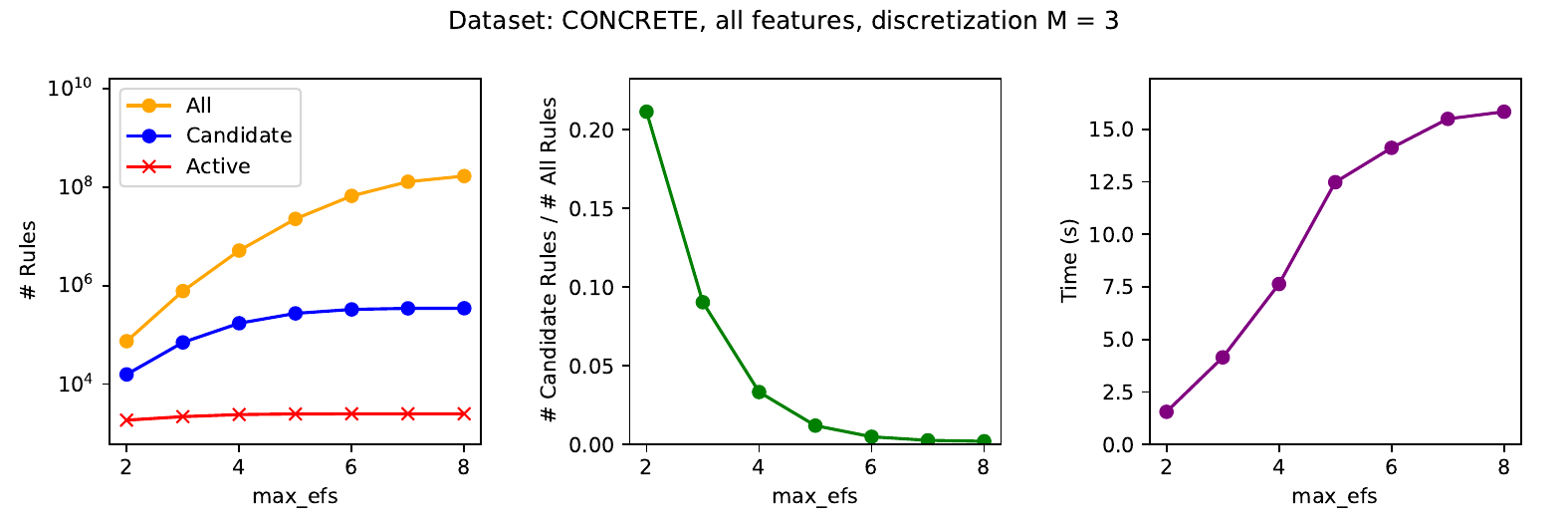} \\
\includegraphics[width=0.75\hsize]{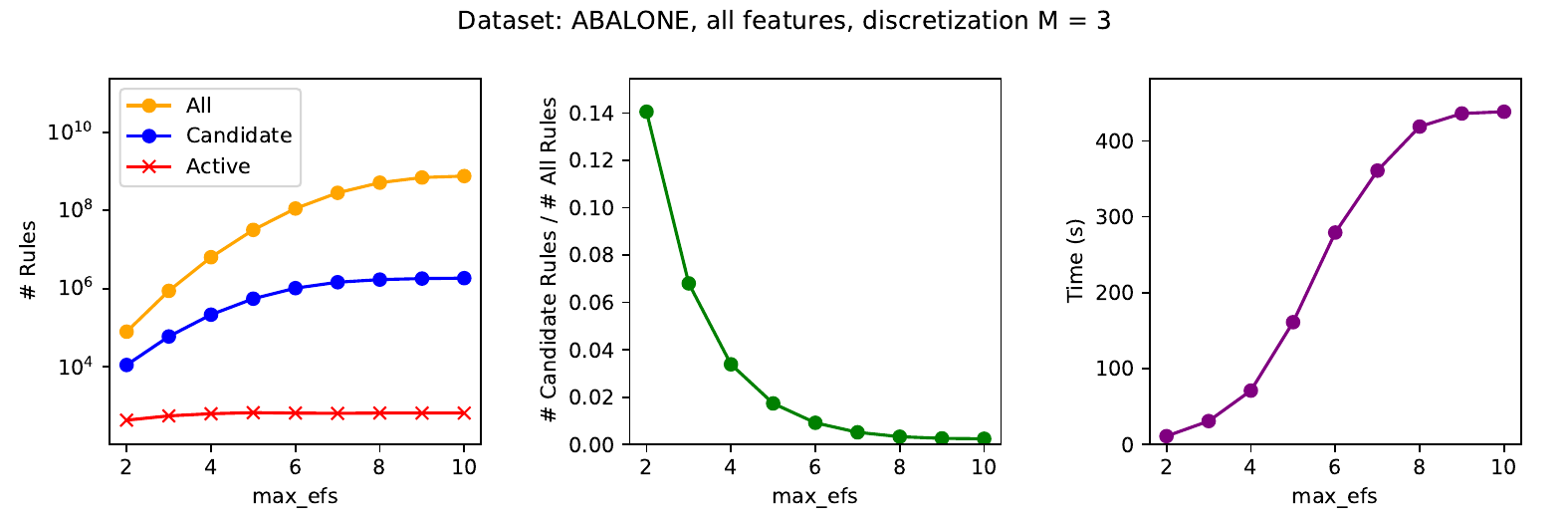}  \\
\includegraphics[width=0.75\hsize]{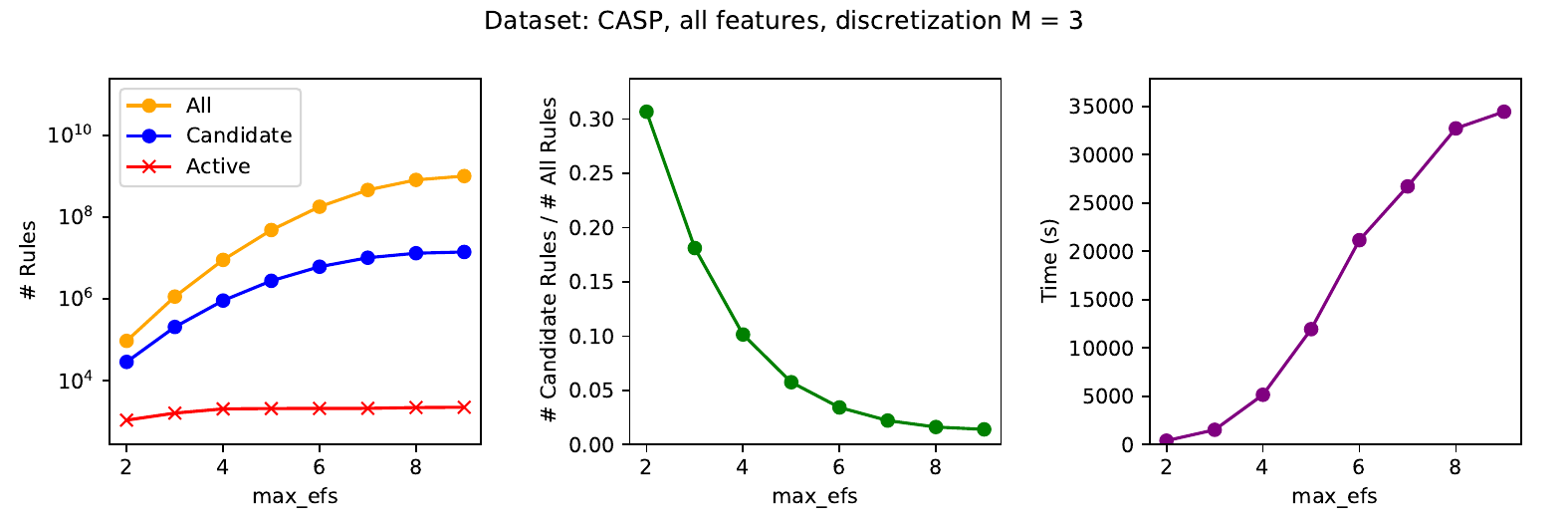}     \\
\includegraphics[width=0.75\hsize]{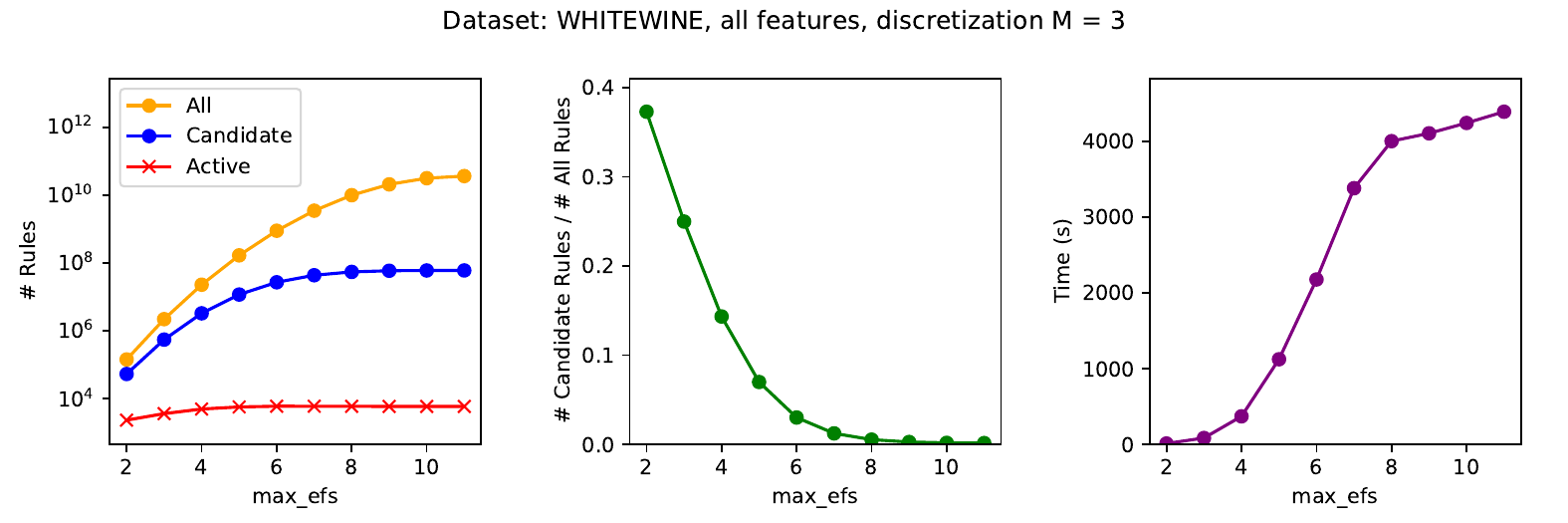}
\end{center}
\caption{Result of Exp.4: Number of enumerated rules by SRF compared to all possible rules, for (C) the changes of ``max\_efs'' (regression datasets). The left is the number of rules (all possible rules, enumerated rules by SRF and active rules at last), the center is the the right is the ratios of the number of rules enumerated by SRF to all possible rules, and the right is the computation time.}
\label{fig:exp4-MaxInputFeat-reg}
\end{figure*}

\begin{figure*}[!p]
\begin{center}
\noindent
\includegraphics[width=0.75\hsize]{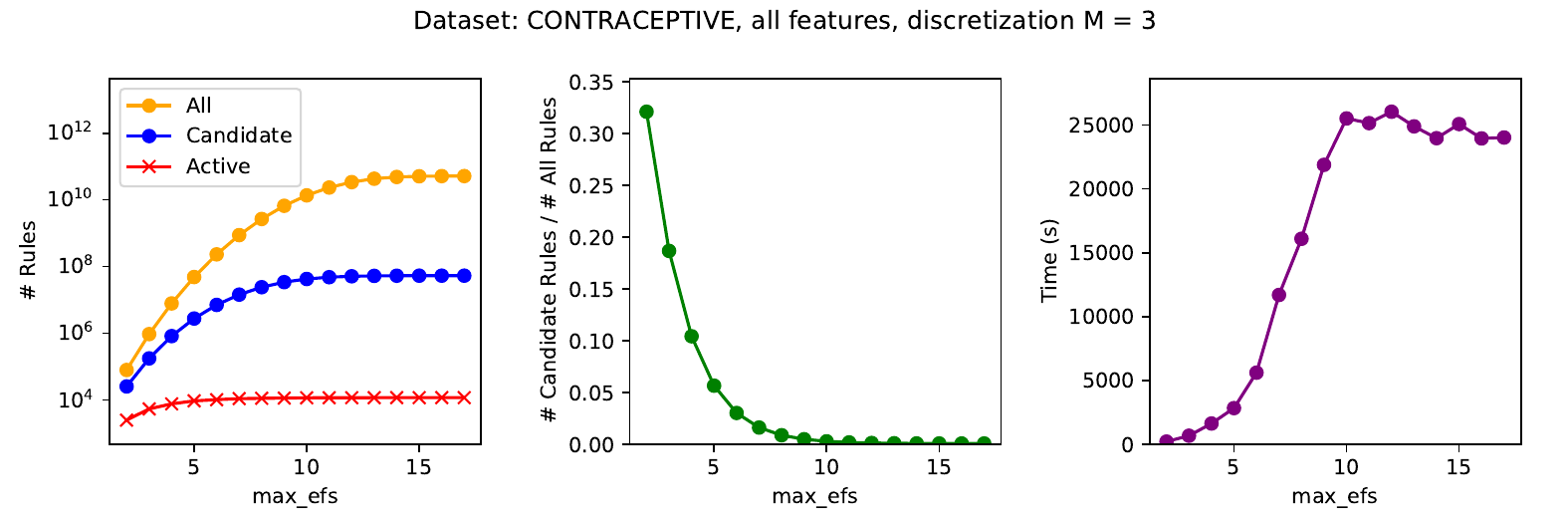}\\
\includegraphics[width=0.75\hsize]{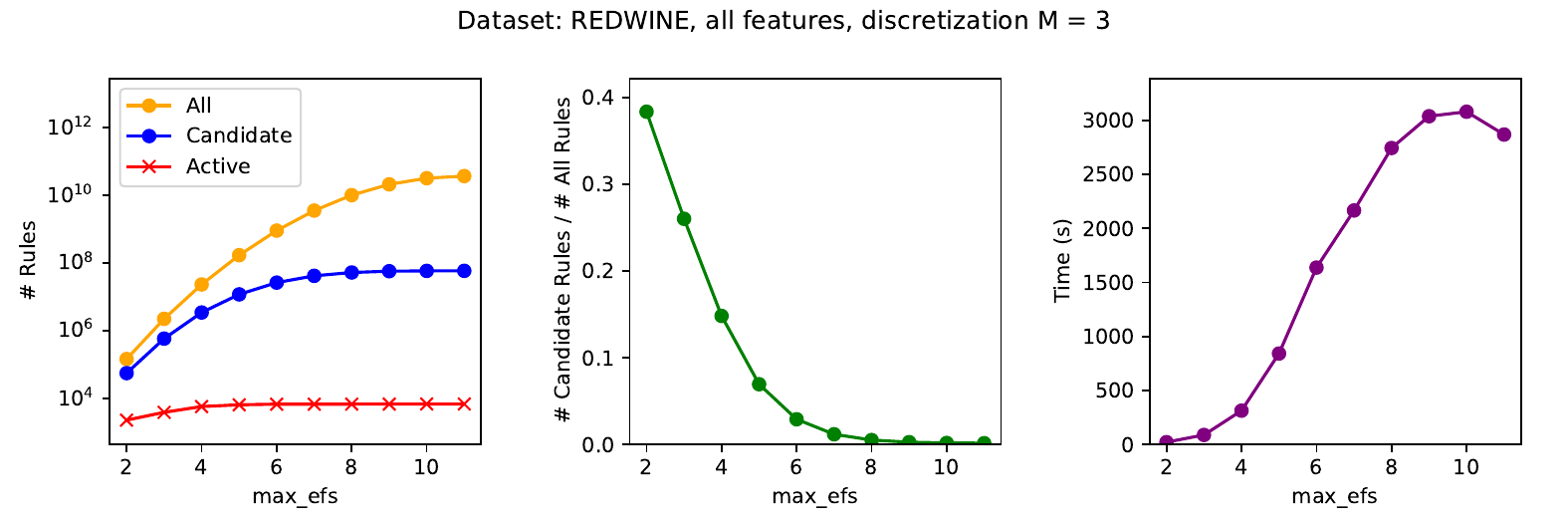}\\
\includegraphics[width=0.75\hsize]{exp-nodes-time/MaxInputFeat-pageblocks.pdf}\\
\includegraphics[width=0.75\hsize]{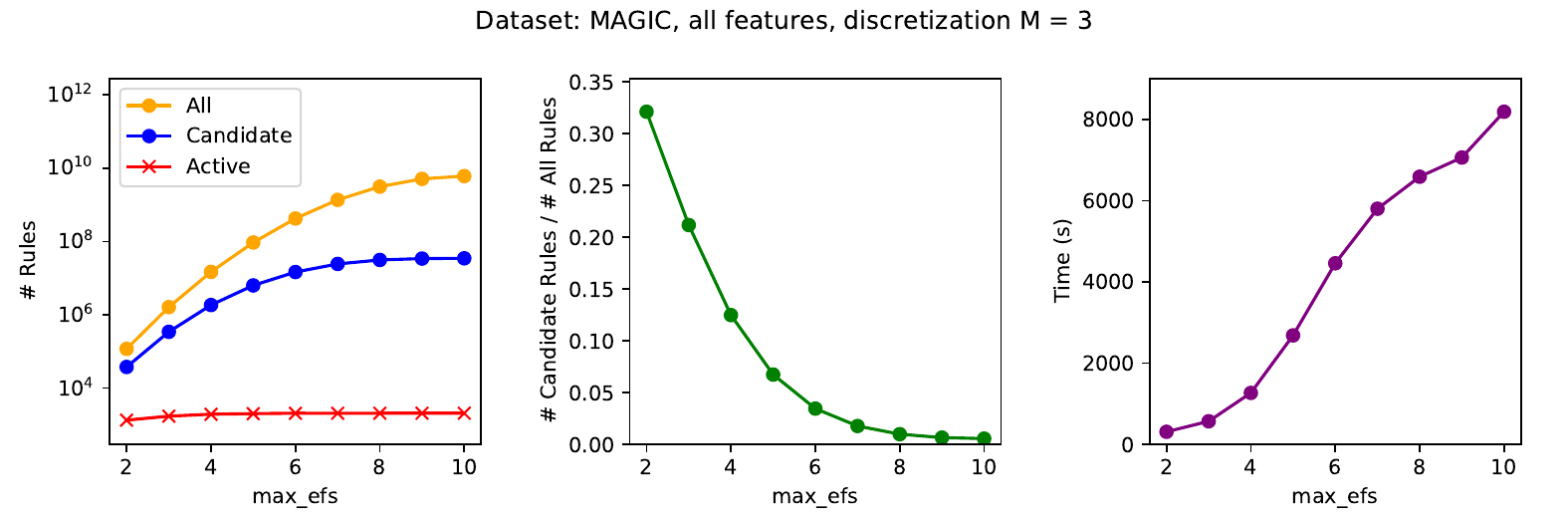}
\end{center}
\caption{Result of Exp.4: Number of enumerated rules by SRF compared to all possible rules, for (C) the changes of ``max\_efs'' (classification datasets). The left is the number of rules (all possible rules, enumerated rules by SRF and active rules at last), the center is the the right is the ratios of the number of rules enumerated by SRF to all possible rules, and the right is the computation time.}
\label{fig:exp4-MaxInputFeat-cls}
\end{figure*}

The results are presented as follows:
(A) the results for the changes of the number of discretizations $M$ are in Figures \ref{fig:exp4-discre-reg} and \ref{fig:exp4-discre-cls},
(B) the results for the changes of the number of input features $d$ are in Figures \ref{fig:exp4-OrigInputFeat-reg} and \ref{fig:exp4-OrigInputFeat-cls}, and
(C) the results for the changes of ``max\_efs'' are in Figures \ref{fig:exp4-MaxInputFeat-reg} and \ref{fig:exp4-MaxInputFeat-cls}.

We can confirm that the rates of enumerated rules are reduced for the increase of the parameters above for almost all results; the only exception is the {\tt CONTRACEPTIVE} dataset in setup (A) (Figure \ref{fig:exp4-discre-cls}), perhaps this is because most of input features in the dataset are binary.

\subsection{Results of Exp.5 for all datasets} \label{app:exp5-results-all}

\begin{figure}[p]
\noindent
\includegraphics[width=\hsize]{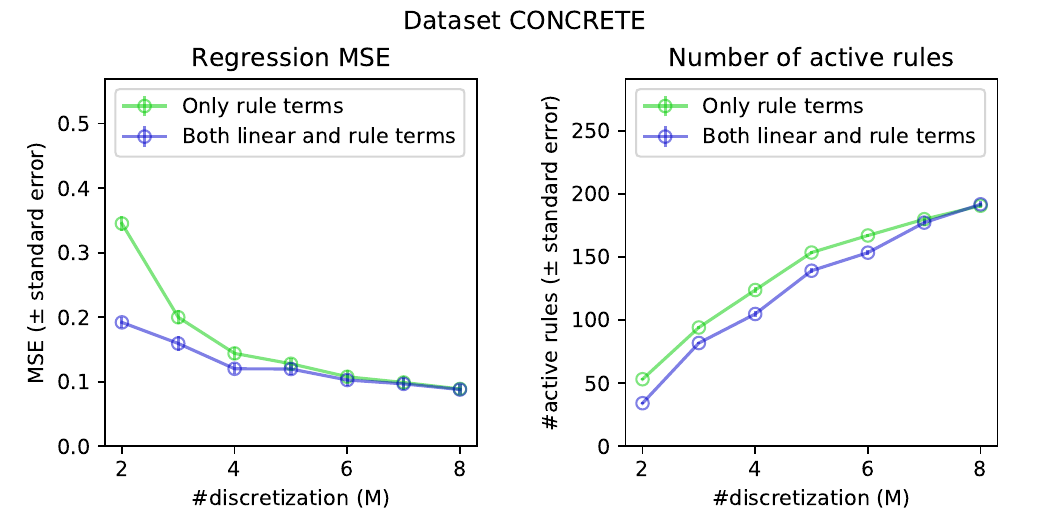}\\
\includegraphics[width=\hsize]{exp-perf-rand/exp5-errorbar-reg-abalone.pdf}\\ 
\includegraphics[width=\hsize]{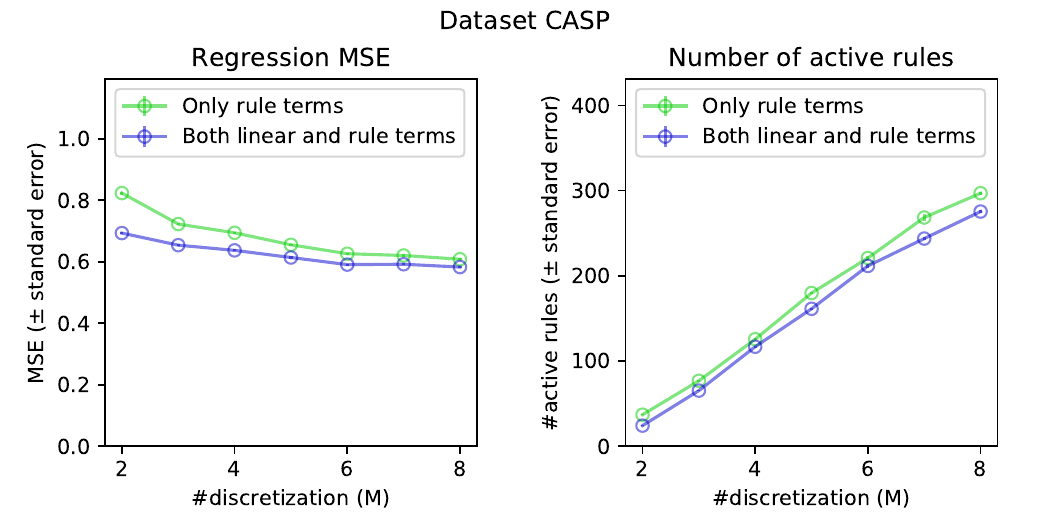}\\ 
\includegraphics[width=\hsize]{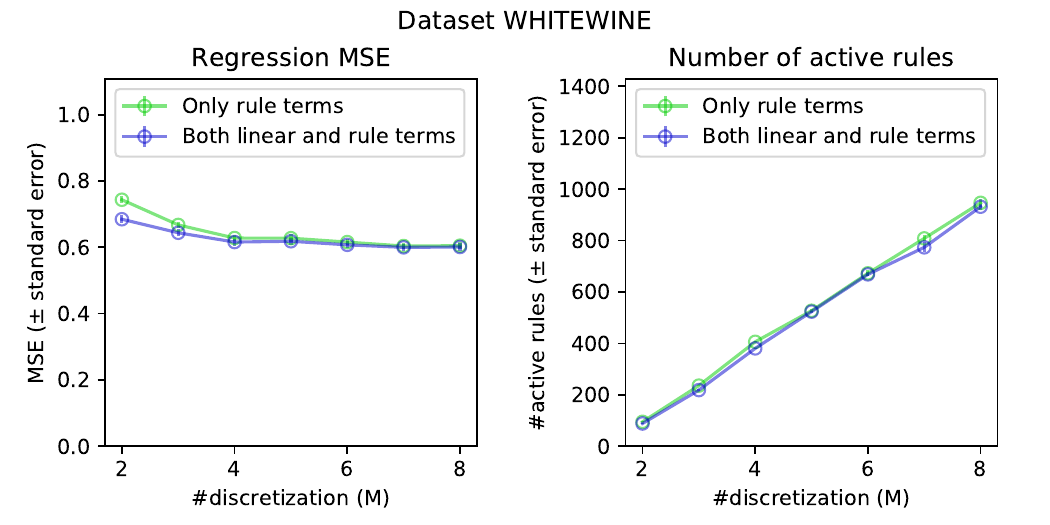}
\caption{Result of Exp.5 for regression datasets. The left is the accuracy by MSE (lower is better), and the right is the number of active rules. The standard errors (error bars) are taken for ten randomized trials.}
\label{fig:exp5-regressions}
\end{figure}

\begin{figure}[p]
\noindent
\includegraphics[width=\hsize]{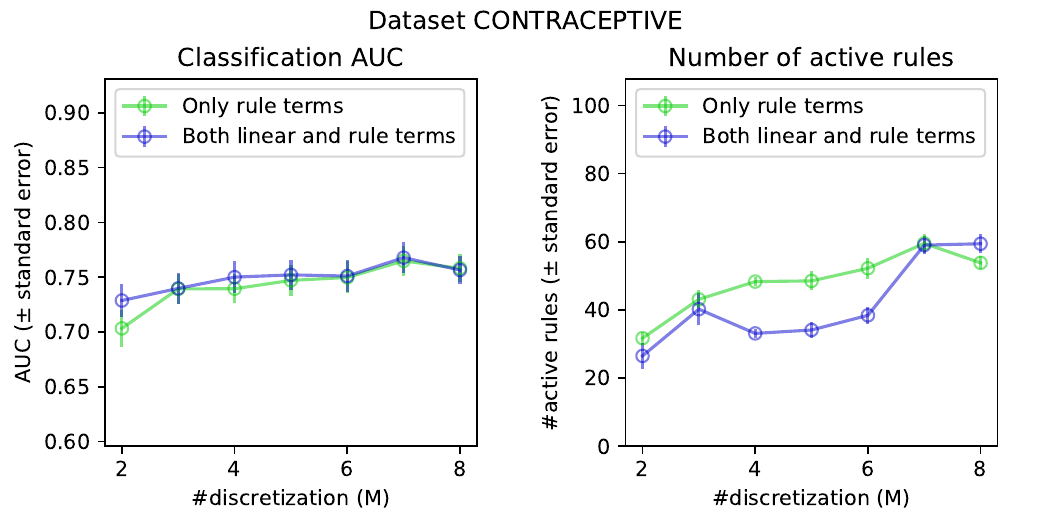}\\
\includegraphics[width=\hsize]{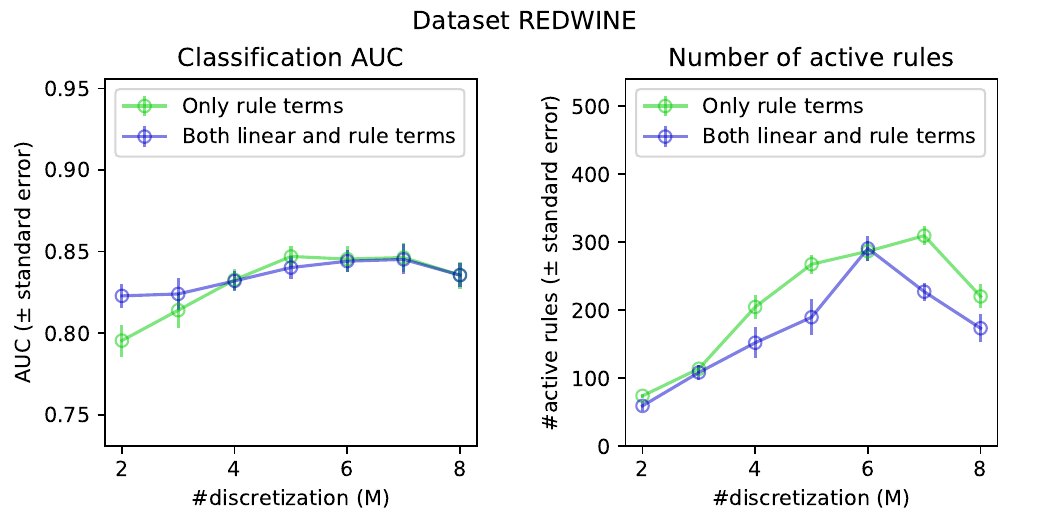}\\
\includegraphics[width=\hsize]{exp-perf-rand/exp5-errorbar-cls-pageblocks.pdf}\\
\includegraphics[width=\hsize]{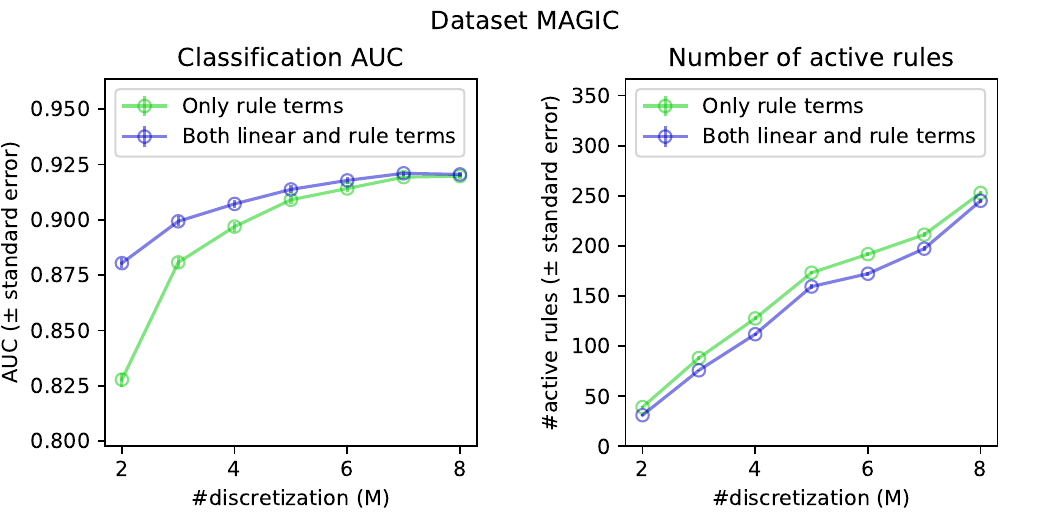}
\caption{Result of Exp.5 for classification datasets. The left is the accuracy by AUC (higher is better), and the right is the number of active rules. The standard errors (error bars) are taken for ten randomized trials.}
\label{fig:exp5-classifications}
\end{figure}

The results are presented in Figure \ref{fig:exp5-regressions} for regression datasets,
and in Figure \ref{fig:exp5-classifications} for classification datasets.

%% file: AppC.tex
\section{Safe RuleFit without linear terms (only rule terms)} \label{app:no-linear-term}

As footnoted in \S\ref{sec:sparse_learning} and experimented in \S\ref{sec:experiment} (Exp.1),
SRF can be applicable for not only the prediction model with both linear and rule terms:
\begin{align}
f(\bx):=b+\bx^{\top}\hbw+\sum_{k\in\cR}r_k(\bm{x})\chw_k,
\tag{\eqref{eq:linear_model} restated}
\end{align}
but also the model with only rule terms:
\begin{align*}
f(\bx):=b+\sum_{k\in\cR}r_k(\bm{x})\chw_k.
\end{align*}

In this appendix we present expressions for SRF for the model with only rule terms.
In principle, to make these expressions, we have only to remove dimensions for the linear terms.
We omit the proofs why just removing dimensions for the linear terms is enough, however, we can easily prove with a similar way to those for both linear and rule terms.

First, the optimization problems of SRF, \eqref{eq:primal} and \eqref{eq:dual}, are respectively rewritten as
\begin{align*}
& \min_{\chbw\in\bbR^{|\cR|}, b\in\bbR} P_{\lambda}(\chbw,b),
	\quad\text{where} \\
& P_{\lambda}(\chbw,b) := \sum_{i\in[n]}\ell_i(y_i,f(\bx_i))+\lambda\|\chbw\|_1. \\
&\max_{\btheta\in\bm{\Delta}} D_{\lambda}(\btheta),
	\quad\text{where}\quad
	D_{\lambda}(\btheta) = -\sum_{i\in[n]}\ell_i^*(-\lambda\theta_i), \\
&\bm{\Delta}:=\{\btheta\in\bbR^n: \|\chbZ^{\top}\btheta\|_{\infty}\leq 1,\bxi^{\top}\btheta=0,\btheta\in\dom(\ell^*)\}.
\end{align*}

Then we show how to modify other expressions for the prediction model with only rule terms.
\begin{itemize}
\item For Lemma \ref{lem:kkt} and Lemma \ref{lem:val_ub} (including expression \eqref{eq:screening}), we have only to use expressions for only rule terms (indexed by the letter $k$) and omit ones for linear terms (indexed by the letter $j$).
\item For Lemma \ref{lem:sphere_bound}, Property \ref{pr:optimize-for-subset} and Theorem \ref{thm:pruning}, we have only to compute expressions as it is even for the model with only rule terms, except that the definitions of $P_\lambda$ and $D_\lambda$ are modified.
\item For expression \eqref{eq:lammax_srf}, we instead compute as $\lammax:=\|\chbZ^{\top}\bm{\phi}\|_{\infty}$.
\end{itemize}

%% file: AppD.tex
\section{Overview of the coordinate descent algorithm} \label{app:coordinate-descent}

As stated in \S\ref{sec:path_alg}, after identifying $\tilde{\cR}$ in Property \ref{pr:optimize-for-subset} by Algorithm \ref{alg:enumerate_all_closed}, we have to solve \eqref{eq:primal} with the help of Property \ref{pr:optimize-for-subset}.
To solve this, we used the coordinate descent \cite{tseng2009coordinate} for SRF and the block coordinate descent \cite{simon2013sparse} for GSRF (\S\ref{sec:extension}).

The overview of the coordinate descent we employed is presented in Algorithm \ref{alg:coordinate} (almost the same applies to the block coordinate descent).
To conduct this, we need to specify the optimality criterion.
In addition, we can conduct {\em dynamic safe screening} stated in \S\ref{sec:path_alg}: we can check SS conditions during conducting the coordinate descent.

Here, there are two parameters that the user must specify.
One is $\varepsilon$: the optimality criterion.
When checking the optimality, we use the duality gap $g$ (it is known that $g=0$ at the optimum),
and we stop the optimization if the ``relative'' duality gap $g/P_{\lambda}(\hbw,\chbw,b)$ fell below $\varepsilon$.
The other is $\text{CheckSteps}$: the frequency of checking the optimality and conducting SS.
Whenever we optimize variables for $\text{CheckSteps}$ cycles, we check the optimality above.
If it is not sufficiently optimized, then we check SS conditions.
Note that the duality gap $g$ calculated for checking optimality is also available for SS, as stated in \eqref{eq:safe_radius}.

In the experiments we used $\varepsilon = 10^{-6}$ and $\text{CheckSteps} = 10$.

\begin{algorithm}[t]
\caption{Overview of the coordinate descent algorithm for SRF.}
\label{alg:coordinate}
\begin{algorithmic}
\STATE {\bfseries Input:} Function $P_{\lambda}(\hbw,\chbw,b)$
	\COMMENT{$\hbw\in\mathbb{R}^d,~\chbw\in\mathbb{R}^{|{\cal R}|},~b\in\mathbb{R}$}
\STATE {\bfseries Input:} $\tilde{\cR}$ \COMMENT{Computed by Algorithm \ref{alg:enumerate_all_closed}}
\STATE {\bfseries Input:} $\varepsilon$, $\text{CheckSteps}$ \COMMENT{Specified by the user}
\STATE {\bfseries Output:} $(\hbw^*,\chbw^*,b^*) = \argmin_{\hbw,\chbw,b} P_{\lambda}(\hbw,\chbw,b)$
\STATE ${\cal D}\gets[d]$
\STATE $\forall k\not\in\tilde{\cR}:~\chw_k \gets 0$
\FOR{$\text{steps} \gets 1, 2, \dots$}
	\STATE \COMMENT{Optimize for variables except for ones that are}
	\STATE \COMMENT{found inactive (Property \ref{pr:optimize-for-subset})}
	\FOR{$j \in {\cal D}$}
		\STATE $\hw_j \gets \argmin_{\hw_j} P_{\lambda}(\hbw,\chbw,b)$
	\ENDFOR
	\FOR{$k \in \tilde{\cR}$}
		\STATE $\chw_k \gets \argmin_{\chw_k} P_{\lambda}(\hbw,\chbw,b)$
	\ENDFOR
	\STATE $b \gets \argmin_b P_{\lambda}(\hbw,\chbw,b)$
	\STATE
	\IF{$\text{steps}\%\text{CheckSteps}=0$}
		\STATE \COMMENT{``$\%$'': remainder of division}
		\STATE
		\STATE \COMMENT{Check optimality}
		\STATE Let $\bm{\theta}$ be the vector $\bm{\theta}^*$ computed by \eqref{eq:kkt-primal2dual} where
			\STATE \hfill $(\hbw^*,\chbw^*,b^*)$ are replaced with $(\hbw,\chbw,b)$.
		\STATE $g\gets P_{\lambda}(\hbw,\chbw,b) - D_{\lambda}(\bm{\theta})$
		\STATE {\bf if}~$g/P_{\lambda}(\hbw,\chbw,b) < \varepsilon$~{\bf then}~{\bf break}
		\STATE
		\STATE \COMMENT{Dynamic safe screening}
		\STATE Compute $r_\lambda$ of \eqref{eq:safe_radius} by
			\STATE \hfill substituting $P_{\lambda}(\hbw',\chbw',b')-D_{\lambda}(\btheta')$ with $g$
		\FOR{$j \gets {\cal D}$}
			\STATE {\bf if}~$\hSSUB(j)<1$~{\bf then}~$\hw_j^*\gets 0$,~${\cal D}\gets{\cal D}\setminus\{j\}$
		\ENDFOR
		\FOR{$k \in \tilde{\cR}$}
			\STATE {\bf if}~$\chSSUB(k)<1$~{\bf then}~$\chw_j^*\gets 0$,~$\tilde{\cR}\gets\tilde{\cR}\setminus\{k\}$
		\ENDFOR
	\ENDIF
\ENDFOR
\STATE {\bf return}~$(\hbw,\chbw,b)$
\end{algorithmic}
\end{algorithm}